\documentclass{article}

\PassOptionsToPackage{numbers, compress}{natbib}

    \usepackage[preprint]{neurips_2023}

\usepackage[utf8]{inputenc} 
\usepackage[T1]{fontenc}    
\usepackage{hyperref}       
\usepackage{url}            
\usepackage{booktabs}       
\usepackage{amsfonts}       
\usepackage{nicefrac}       
\usepackage{microtype}      
\usepackage{xcolor}         
\usepackage{makecell}       
\usepackage{times}
\usepackage{epsfig}
\usepackage{graphicx}
\usepackage{subfigure}
\usepackage{amsmath}
\usepackage{amssymb}
\usepackage{amsfonts}
\usepackage{booktabs}
\usepackage{siunitx}
\usepackage{multirow}
\usepackage[T1]{fontenc}
\usepackage{colortbl}
\usepackage{authblk}
\usepackage{lineno} 

\title{CeRF: Convolutional Neural Radiance Fields for New View Synthesis with Derivatives of Ray Modeling}

\author{
  Xiaoyan Yang\thanks{Equal contribution.}\quad
  Dingbo Lu$^{*}$\quad
  Yang Li\thanks{Yang Li and Changbo Wang are the corresponding authors.}\quad
  Chenhui Li\quad
  Changbo Wang$^{\dagger}$\quad \\ 
  \small School of Computer Science and Technolgy, East China Normal University \\
  \small \texttt{\{51215901035, 51215901103\}@stu.ecnu.edu.cn} \\
  \small \texttt{\{yli, chli, cbwang\}@cs.ecnu.edu.cn} 
}

\begin{document}

\maketitle

\begin{abstract}
 In recent years, novel view synthesis has gained popularity in generating high-fidelity images. While demonstrating superior performance in the task of synthesizing novel views, 
 the majority of these methods are still based on the conventional multi-layer perceptron for scene embedding. 
 Furthermore, light field models suffer from geometric blurring during pixel rendering, while radiance field-based volume rendering methods have multiple solutions for a certain target of density distribution integration. 
 To address these issues, we introduce the Convolutional Neural Radiance Fields to model the derivatives of radiance along rays. 
 Based on 1D convolutional operations, our proposed method effectively extracts potential ray representations through a structured neural network architecture.
 Besides, with the proposed ray modeling, a proposed recurrent module is employed to solve geometric ambiguity in the fully neural rendering process. 
 Extensive experiments demonstrate the promising results of our proposed model compared with existing state-of-the-art methods.
\end{abstract}

\section{Introduction}

Recently, novel view synthesis with neural implicit representations has rapidly advanced due to its surprisingly high-quality generated image with different camera poses.
Based on the original neural radiance field~(NeRF) method~\cite{Mildenhall2020NeRFRS}, various research directions have been explored to further improve rendering quality, rendering speed, and other aspects~\cite{barron2021mipnerf, Chen2022ECCV,  wang2023f2nerf, yu2022monosdf}. 
In addition, as NeRF bridges the radiance and geometrics information, many other applications exploit NeRF as their building block to achieve other graphic goals, such as extracting geometrical, semantic, and material information from the scene\cite{rakotosaona2023nerfmeshing, wong2023factored, zhu20232}, and extending static setting into dynamic scenes~\cite{attal2023hyperreel}. 

\begin{figure}[!t]
    \centering
    \begin{minipage}{0.51\textwidth}
        \centering 
        \includegraphics[clip, trim=9.1cm 5cm 7.4cm 3.0cm, width=\textwidth]{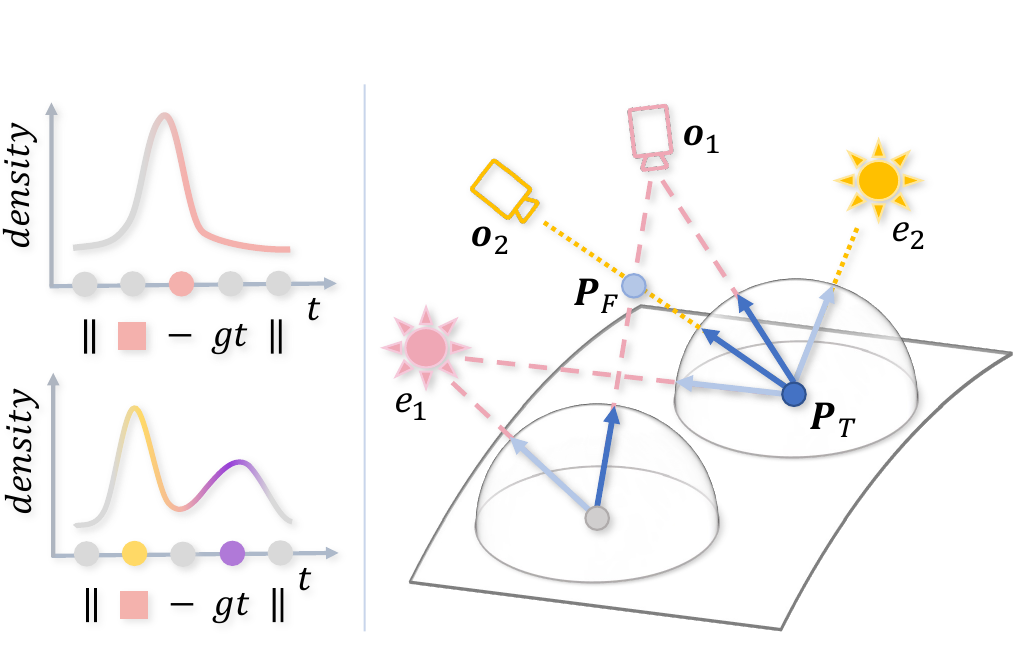} 
        \caption{Ambiguity in light field and radiance field}
        \label{fig:multi-sol}
    \end{minipage}
    \hfill
    \begin{minipage}{0.48\textwidth}
        \centering
        \includegraphics[clip, trim=8.5cm 3.6cm 8.6cm 4.2cm, width=\linewidth]{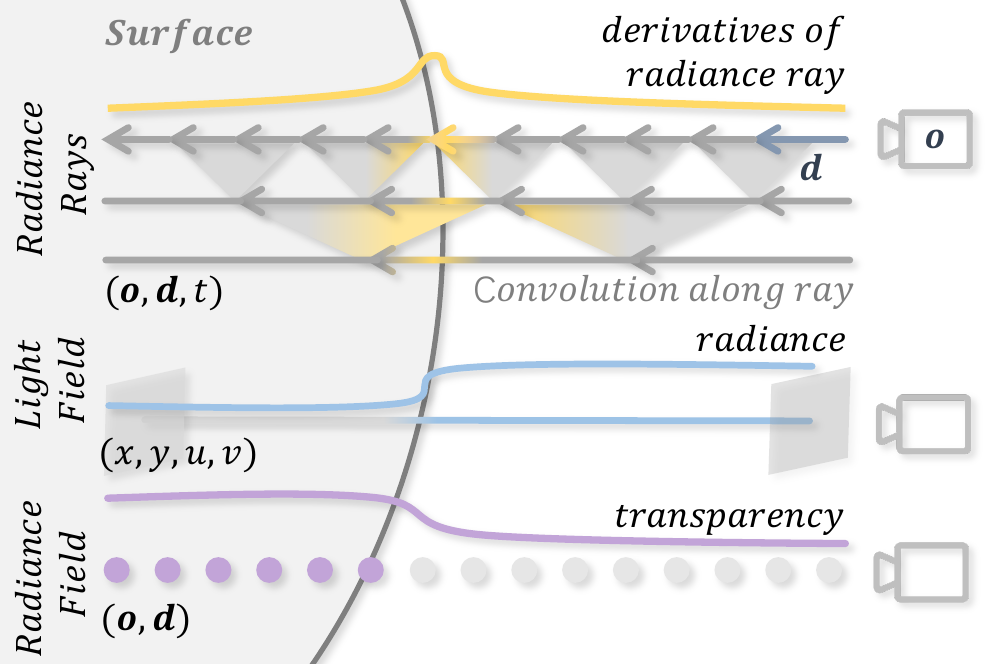}
        \caption{Visualization for different fields}
        \label{fig:conv-op}
    \end{minipage}
\end{figure}

As most of these methods are based on the NeRF framework, the volume rendering strategy remains in the processing pipeline yet has multi-solution ambiguity.
As shown in the left of Figure~\ref{fig:multi-sol}, the same integral color is obtained under completely different density distributions. 
Similarly, for light field-based methods~\cite{Sitzmann2021LightFN}, the geometric configuration is ambiguous since both the point on the surface $\mathbf{P}_T$ and the point in space $\mathbf{P}_F$ have the same colors viewing from two pose, as shown in the right of Figure~\ref{fig:multi-sol}.
Specifically, the different points $\mathbf{P}_T$ and $\mathbf{P}_F$ receive same radiance $L(\mathbf{P}_T, \mathbf{o}_i)=L(\mathbf{P}_F,\mathbf{o}_i)$ for $i=\{1,2\}$ from two camera observation $\mathbf{o_1}$ and $\mathbf{o_2}$.
This indicates that there are potentially multiple position solutions to satisfy the same view-dependent radiance, which is hard to optimize for neural networks. NeuS~\cite{wang2021neus} introduces neural implicit surfaces with surface constraints. However, surface-based approaches have complicated conversion from SDF to volume density.



Although SDF-based methods have complicated data structures, the network structure is very simple as they mainly employ Multi-Layer Perceptron~(MLP) for implicit representation.
In the field of deep learning, MLP is initially the simplest design, but many architectures have been developed since then, including Convolutional Neural networks (CNN)~\cite{gu2018recent} and Transformers~\cite{vaswani2017attention}. 
Furthermore, CNNs have been shown to be successful in image classification tasks, such as ResNet~\cite{he2016deep}. Additionally, recurrent neural networks (RNNs) have proved to be efficient in handling sequential data.

In this paper, we model the novel view synthesis task as an implicit neural representation of the derivatives of radiance in the scene. 
As depicted in Figure~\ref{fig:conv-op}, the radiance remains constant until the rays hit the surface, and the radiance remains zero internally for homogeneous materials. 
Compared to modeling in NeRF and light field methods, our proposed method only requires the network to store information on the surface, rather than the entirety of the empty space from origin to intersection.
Therefore, modeling the radiance rays as derivatives yields sparse solutions along individual rays and the optimization problem changes from a regression task to a classification-like task.


To further leverage the advantages of the modeling, we propose the Convolutional Neural Radiance Fields~(CeRF) to extract local features along the ray with a fully neural rendering scheme.
Our key ideas are to incorporate a sophisticated CNN structure to achieve a more streamlined and uninterrupted ray representation and formulate a gated recurrent unit~(GRU) in the rendering network.
By approximating the render equation with a fully neural network process, the theoretical upper limit of our proposed approach exceeds the expression capability of the volume-based rendering scheme in conventional NeRF.
Extensive experiments demonstrate that, by modeling radiance derivatives, our proposed CeRF exhibits a more elegant and effective encoding of scenes, and achieves comparable results with state-of-the-art methods. Our proposed structure can be easily incorporated into all NeRF-based methods designed for specific scenes, leading to improved rendering results.
Our contributions can be summarized as follows:

(1) A theoretical analysis of the rendering equation with novel radiance model. To the best of our knowledge, we are the first to model the derivatives of radiance along a ray in novel view synthesis.

(2) A novel CeRF framework to encode ray features and employs a neural ray rendering based on our proposed rendering scheme.

(3) Experiments demonstrate that CeRF attains comparable results to the state-of-the-art models and accurately fit complex geometric details. We will release our source code for further research.



\section{Related work}

\subsection{Neural light fields}

Neural light fields have been well-received in synthesizing new views due to their good results on scenes with complex optical phenomena and fast rendering speed~\cite{suhail2022light}. 
NeX~\cite{wizadwongsa2021nex} utilizes a linear combination of multiple neural network-based basis functions to obtain pixel color values. 
RSE~\cite{attal2022learning} maps the input rays to a high-dimensional space, and the rendering of each pixel requires only one visit to the network, resulting in faster view synthesis.
For dense input images, light field-based rendering methods~\cite{Gortler1996TheL,Levoy1996LightFR} can be used to synthesize new views. However, for sparse input images, light field-based interpolation alone fails to generate images with scene coherence due to the large spacing between viewpoints. 
Wang \textit{et al.}~\cite{wang2022progressively} propose a progressively connected neural network for light field view synthesis, improving both efficiency and quality. 
LFN~\cite{Sitzmann2021LightFN} proposes a method for learning a neural representation of the scene that can be rendered by a single forward pass to a new view. 
LFNR~\cite{suhail2022light} further improves upon this by introducing constraints on the geometry of multiple viewpoints in the light field to learn the representation of a scene in a sparse set of inputs. 
R2L~\cite{wang2022r2l} distills neural radiance fields into neural light fields for efficient view synthesis, converting the former into the latter for more efficient view synthesis. 
Compared with light field-based approaches, our proposed CeRF is capable of dealing with geometric information and has a very different light ray modeling scheme.

\subsection{Neural radiance fields}

Mildenhall \textit{et al.}~\cite{Mildenhall2020NeRFRS} proposed NeRF, which represents the scene as a continuous function of 3D spatial coordinates, allowing for high-quality rendering of new views. 
Subsequently, many improvements to NeRF have been proposed.
Some methods aim to improve rendering quality~\cite{barron2021mipnerf,verbin2022ref,huang2023local,wu2023alpha,Peng2022CageNeRFCN,Tang2022NeuralSD,Yin2022CoordinatesAN}. Others focus on optimizing inference speed or model size~\cite{muller2022instant,lindell2022bacon,barron2023zip,Chen2022ECCV,Tang2022CompressiblecomposableNV,Liu2022DeVRFFD}. 
Among these, Mip-NeRF\cite{barron2021mipnerf} extends the rays-based representation to frustums, significantly improving the rendering effectiveness of NeRF. Plenoxels~\cite{Yu2021PlenoxelsRF} uses a sparse voxel grid to explicitly represent a scene, speeding up convergence and rendering. DVGO~\cite{Sun2022ImprovedDV} proposes the post-activation interpolation on voxel density and incorporates a priori to solve suboptimal geometry solutions, achieving fast convergence of the network. In addition, some methods not only generate rendered images from implicit representations but also model more about {geometry}~\cite{yang2022neumesh}, {semantic}~\cite{wang2022dm}, or {material} information~\cite{verbin2022ref, zhu20232, bao2023sine}.  NeuRay~\cite{Liu2021NeuralRF} predicts the visibility of 3D points in input views, allowing the network to focus on visible image features when constructing the radiance field. Methods that simultaneously estimate SDF can alleviate the problem of multiple solutions, like PhySG~\cite{Zhang2021PhySGIR}, an SDF-based hybrid representation of specular BRDFs and environmental illumination, to enable physics-based editing of material and lighting radiance. Ref-NeRF\cite{Verbin2021RefNeRFSV} parameterizes the scene as a view-dependent outgoing radiance, improving the realism of specular reflections. However, all these methods employ MLP as the main neural network. In contrast, our proposed CeRF focuses on ray derivatives modeling and its corresponding structure of neural networks.
Mukund \textit{et al.}~\cite{varma2022gnt} proposed to use a transformer instead of a volume rendering function, which is closest to our method. However, our architecture utilizes a simpler convolutional network, while achieving better performance.



\section{Approach}

\begin{figure*}[!t]
  \centering
     \includegraphics[clip, trim = 2cm 4.cm 2.3cm 4cm, width=0.98\linewidth]{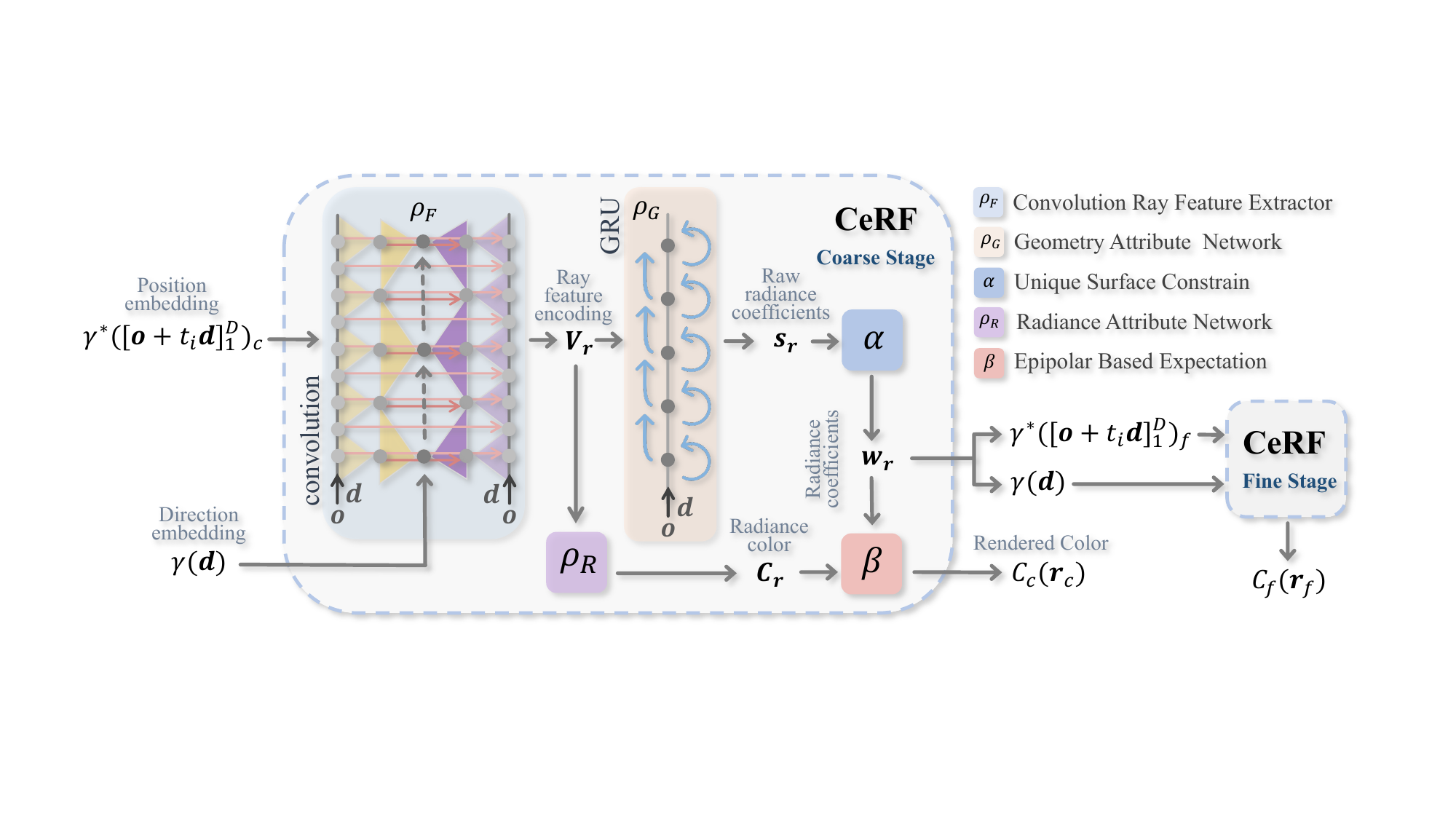}
     \caption{$\mathbf{Overview}$. Based on the modeling of the derivatives of radiance ray, we proposed a fully neural rendering framework. Two key ideas guide our design: 1) encoding the light using the convolutional operation on sampled points along a ray, and 2) using a neural network to fit the rendering process, which can render new views.}
   \label{fig:overview}
\end{figure*}

Given a set of images with known camera poses and intrinsic parameters, our target is rendering new views for unseen poses. To this end, we introduce CeRF, as shown in Figure \ref{fig:overview}. Specifically, for each pixel in an image, CeRF emits a ray $\mathbf{r}$ from a viewing origin $\mathbf{o}$ with a normalized view direction $\mathbf{d}$ and a set of distances ${t_i}, i \in[1,D] $ along rays, and predicts the pixel color $C(\mathbf{r})$.

Viewing points along a ray as a whole, the position embedding 
and direction embedding 
are inputted into Convolutional Ray Feature Extractor to extract ray feature encoding $\mathbf{V_r}$. A GRU-based Geometry Attribute Network with Unique Surface Constraint $\alpha$ to estimate geometry attribute coefficients $\mathbf{w_r}=[w_1,...,w_D]$. Meanwhile, Radiance Attribute Network predicts radiance color $\mathbf{C_r}=[c_1,...,c_D]$. The final pixel color $C(r)$ is obtained by calculating the expectation and introducing epipolar points.
The coarse-to-fine structure is used as the same as other NeRF-based methods~\cite{barron2021mipnerf}.

In the following sections, we first derivate the radiance ray from the render equation, followed by an overview of the Convolutional Ray Feature Extractor. Then, a detailed description of our proposed neural radiance ray rendering. Finally, the loss function for parameters optimization is introduced.

\subsection{Derivatives of radiance ray}

We begin with the render equations in computer graphics to build our model. To render a synthesized image, we need to calculate the radiance $L(\mathbf{p}, \mathbf{\omega}_o)$ leaving point $\mathbf{p}$ in direction $\mathbf{\omega}_o$ along the view ray.
For new view synthesis settings, the pose, and intensity of the light sources in the scene do not change with time or wavelength. Therefore, the formula of the render equation at one surface point $\mathbf{p}$ can be simplified to scattering equation~\cite{foley1996computer} as follows:
\begin{equation}
L(\mathbf{p}, \mathbf{\omega}_o) = L^{\text{out}}(\mathbf{p}, \mathbf{\omega}_o)+
\int_{\mathbf{\omega}_i \in \mathbf{S^2(p)}} 
L^{\text{in}}(\mathbf{p}, -\mathbf{\omega}_i) 
f_s(\mathbf{p}, \mathbf{\omega}_i, \mathbf{\omega}_o) 
|\mathbf{\omega}_i \cdot \mathbf{n}_p |
 d\mathbf{\omega}_i,
 \label{eq:scatter}
\end{equation}
where $L^{\text{out}}(\mathbf{p}, \mathbf{\omega}_o)$ symbolized the light emitted by point $\mathbf{p}$ in direction $\mathbf{\omega}_o$, and $L^{\text{in}}(\mathbf{p}, -\mathbf{\omega}_i)$ represents the light coming to the surface.  $f_s(\mathbf{p}, \mathbf{\omega}_i, \mathbf{\omega}_o)$ is the bidirectional scattering distribution functions (BSDFs) that characterizes the scattering behavior of light, for given incoming and outcoming directions $\mathbf{\omega}_i$ and $\mathbf{\omega}_o$, respectively. 
The integrable region $\mathbf{S}^2$ is the direction of all incident light coming to a point $\mathbf{p}$. 

Given a ray emitted from the camera origin $\mathbf{o}$ in the direction $\mathbf{d}$, its intersection with a surface can be represented as $\mathbf{p}=\mathbf{o}+t\mathbf{d}$, where $t$ is the distance from the intersection to the origin of the view ray. The direction of the outgoing radiance $\mathbf{\omega}_o$ is opposite to the view direction $\mathbf{d}$, i.e., $\mathbf{d}=-\mathbf{\omega}_o$. 
As we only consider the radiance flux along a single ray, the radiance can be rewritten as $L(\mathbf{o},\mathbf{d},t)$, which only depends on three variables $\mathbf{o}, \mathbf{d}, t$. Based on the universal function fitting capability of neural networks, we believe that a well-designed network $\mathbf{F}$ can approximate the render function for different materials. Hence, we propose an approximation factorization of the scattering equation as
\begin{equation}
\begin{aligned}
 L(\mathbf{o},\mathbf{d},t) 
 =
 \mathbf{F}(\mathbf{o}, \mathbf{d}, t; \Theta),
\label{eq:fit}
\end{aligned}
\end{equation}
where $\Theta$ is network parameters encoding the constant properties for diverse scenes. 
In general, 
we can consider that the radiance of the entire static scene, including both global and local illumination, has been precomputed and baked into a texture map on the object surface~\cite{silvennoinen2019ray}. Consequently, we can use a powerful function to fit the rendering effects.

Light field and radiance field can model the rendering equation, but both of them are multiple possibilities from different perspectives. 
In addition, the neural network needs to learn the surface radiance even at empty space, leading to increased complication in network fitting. In order to obtain a single mapping space, we model the radiance ray as the derivatives of radiance  
$\partial L(\mathbf{o}, \mathbf{d}, t) / {\partial t}$ 
instead of the radiance 
$L(\mathbf{o},\mathbf{d}, t)$
itself. Thus, the network directly predicts the position and color of the surface intersection points. 
The radiance along a ray $r$ from origin $\mathbf{o}$ in direction $\mathbf{d}$ can be expressed as the integral of the derivatives of the radiance ray as
\begin{equation} 
\begin{aligned}
\label{eq:int}
 L(\mathbf{o},\mathbf{d},t) =
 \int_t^{+\infty} \frac{\partial L(\mathbf{o},\mathbf{d},x)}{\partial x} dx.
\end{aligned}
\end{equation}
As the radiance only changes significantly near the surface and remains zeros in empty space or inside objects, we introduce Dirac's delta function $\delta(\mathbf{\mathbf{o},\mathbf{d},t})$ to approximate the derivatives of radiance  
$\partial L(\mathbf{o},\mathbf{d},t) / {\partial t}$ as
\begin{equation} 
\begin{aligned}
\label{eq:dirac}
 \frac{\partial L(\mathbf{o},\mathbf{d},t)}{\partial t} \propto \delta(\mathbf{o},\mathbf{d},t)L(\mathbf{o},\mathbf{d}, t),
\end{aligned}
\end{equation}
where Dirac's delta function only has value on the peak located at the intersection of the light and the surface. 
To implement this radiance ray in practice, we discretize Eq.~\ref{eq:int} and Eq.~\ref{eq:dirac} as 
\begin{equation} 
\begin{aligned}
 L(\mathbf{o},\mathbf{d}, t) 
 \approx \sum\limits_{i=t}^D 
 \frac{\Delta L(\mathbf{o},\mathbf{d}, i)}{\Delta i} 
 = \sum\limits_{i=1}^D\mathbb{I}\left[{\mathbf{o},\mathbf{d},i}\right]
 L(\mathbf{o},\mathbf{d}, i),
\end{aligned}
\label{eq:indicator}
\end{equation}
where the indicator function $\mathbb{I}\left[\mathbf{o},\mathbf{d},i\right]$
is equivalent to the delta function $\delta(\mathbf{o},\mathbf{d},t)$ in the non-continuous situation. As the indicator function $\mathbb{I}\left[\mathbf{o},\mathbf{d},t\right]$ is only non-zero at the sampling points near the surface, the range of summation can be extended from $\left[t, D\right]$ to $\left[1, D\right]$, hence the radiance is not relative with distance $\mathbf{t}$.
Then, the rendering equation can be split into two parts: $\mathbb{I}\left[\mathbf{o},\mathbf{d},i\right]$ that expresses the intersection between light ray from the indicator function and the scene and $L(\mathbf{o},\mathbf{d}, i)$ records the radiance at different locations in the scene.

To approximate Eq.\ref{eq:indicator} with a neural function, we first propose a convolutional Ray Feature Extractor 
$\mathbf{V_r} = \rho_F(\mathbf{o},\mathbf{d},\mathbf{t}; \theta_F), \textup{where} \, \mathbf{t}=[t_1, ...,t_D]
\label{eq:v}$
for ray information embedding. Then we denote the intersection position by the Geometry Attribute Network $\rho_G(\mathbf{V_r};\theta_G)$ with the Unique Surface Constrain $\alpha$ for approximating single-peak indicator function $\mathbb{I}$. Meanwhile, the proposed Radiance Attribute Network $\rho_R(\mathbf{V_r}; \theta_R)$ conveys the intersection radiance.
Finally, by substituting Eq.~\ref{eq:indicator}, we obtain an approximation expression of the scattering equation in radiance ray as
\begin{equation} 
\begin{aligned}
L(\mathbf{o},\mathbf{d}) 
 \propto
 \sum\limits_{i=1}^D\alpha\left[{\rho_G(\mathbf{V_r}; \theta_G)}\right] 
 \rho_R(\mathbf{V_r}; \theta_R).
\end{aligned}
\label{eq:aprox}
\end{equation}

Please note that compared to the traditional NeRF, the Radiance Attribute Network $\rho_R$ of CeRF only needs to have the correct values on the surface, whereas the former needs to accurately predict the density values of all points in the entire space. Due to the introduction of the indicator function $\mathbb{I}\left[\mathbf{o},\mathbf{d},i\right]$, CeRF only selects the strongest signal points. Even if the network predicts noise values at other locations, they will be suppressed by the Unique Surface Constrain. Therefore, the network is more likely to learn corresponding relationships of unimodal distribution.

\subsection{Convolutional Ray Feature Extractor}

After taking the derivatives, only one peak exists in the target distribution of each ray. In order to design a network that parameterizes this model more effectively, we first consider treating the space of each ray as a complete scene representation and utilize 1D convolution to extract features for the one-hot spot of the intersection surface. These sparse interactions enable the network to have a more continuous input representation by parameter sharing and have been proven effective in many previous tasks~\cite{kiranyaz20211d}.
Specifically, inspired by ray parameterize in radiance field~\cite{kuganesan2022unerf}, we define the sequential sampling points on one ray as a whole and model it as our scene representation $\mathbf{r}=\left[\mathbf{o}+t_i\mathbf{d}\right]_1^D$, to achieve more continuous and compact representation $\mathbf{V_r}$ in Eq.~\ref{eq:aprox}.
MLPs only map individual points into high-dimensional feature representations, which has limitations in modeling interdependency between adjacent points. 
CNN encodes the characteristics of neighbor points on the ray through the kernel convolution, allowing adjacent sampled points to be interdependent as shown in Figure \ref{fig:conv-op}. 
We propose the Convolutional Ray Feature Extractor $\rho_F(\mathbf{r}; \theta_F)$ with a U-shaped CNN architecture. 

\begin{figure*}[!t]
    \centering
    \includegraphics[clip, trim = 2.1cm 5.5cm 0.3cm 5.4cm, width=0.99\linewidth]{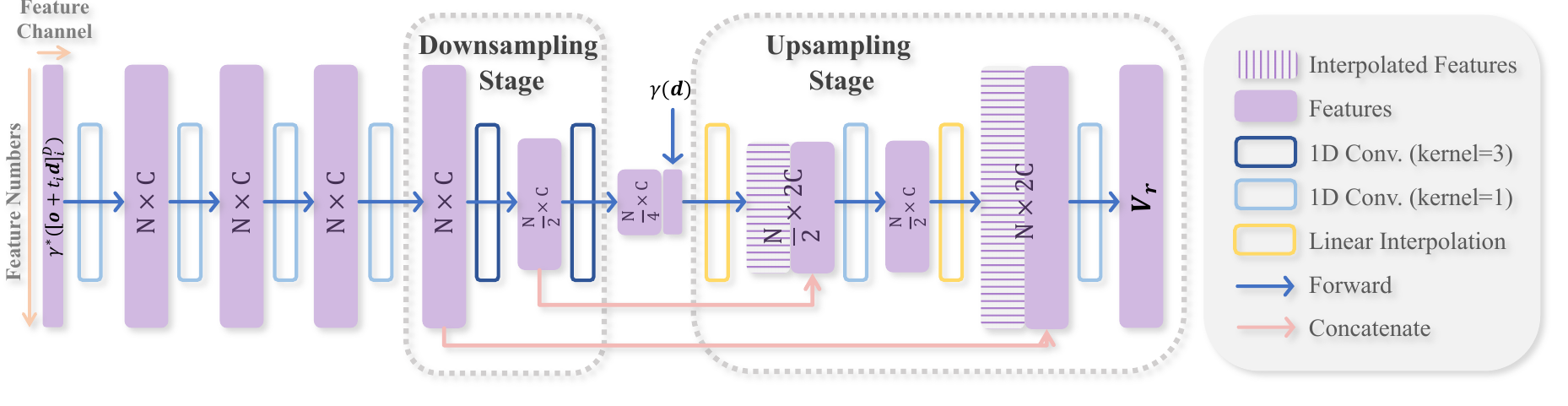}
    \caption{Architecture of Convolutional Ray Feature Extractor}
    \label{fig:pos-net}
\end{figure*}

As shown in Figure ~\ref{fig:pos-net}, when a ray $\mathbf{r}$ is input into the Convolutional Ray Feature Extractor, the spatial position encoding is low-dimension at first. Therefore, the first four layers of convolution with a kernel of 1, as same as MLP, are applied to extract the high-dimensional deep features of individual points. The subsequent convolutional layers learn the local and global information of the light ray in the latent representation.
The next two convolutional downsampling layers, which have a kernel size of 3 and stride of 2, consider the positions and features of neighboring points when producing feature values.
This feature embedding architecture implicitly encodes the observation view direction from sampling positions along a ray.
The convolutional structure compresses the number of sampling points along the light direction, producing bottleneck features. 
Through this layer-by-layer convolution, the receptive field gradually increases, and the bottleneck feature embeds the entire light ray, as shown in Figure \ref{fig:conv-op}.
Finally, the last two upsampling layers utilize linear interpolation for upsampling and concatenate the interpolated feature vectors with relative vectors from the downsampling stage. The output ray feature encoding $\mathbf{V_r}$ contains both position-dependent local features and light-dependent global features.

\subsection{Neural radiance ray rendering}

For the two parts in Eq.\ref{eq:aprox}, we use the Geometry Attribute Network $\rho_G$ and Radiance Attribute Network $\rho_R$ to predict the most possible position of the surface intersection and the radiance of the surface, respectively. The Geometry Attribute Network outputs the raw geometry attribute coefficient $\mathbf{s}_r = [s_1, s_2, ..., s_D]=\rho_{G}(\mathbf{V_r}; \theta_G)$, which ranges from zero to one. Meanwhile, given the ray feature encoding $\mathbf{V_r}$ as input, the Radiance Attribute Network outputs the radiance color $\mathbf{C_r} = [c_1, c_2, ..., c_D]=\rho_R(\mathbf{V_r};\theta_R)$.


Given a ray $\mathbf{r}$, the Geometry Attribute Network primarily learns the intersection position and local incoming radiance. To handle occlusion relationships among points on a light ray, we utilize the GRU~\cite{chung2014empirical}, which has demonstrated strong performance in sequence learning. During each recursive step, all points sampled before the current one are used as priors.  The GRU-based module outputs raw geometry attribute coefficient $\mathbf{s}_r$ through a three-layer fully connected network with $sigmoid$ as the activation function of the last layer.

In order to solve the problem of ambiguity in the fitting problem, we introduce the Unique Surface Constrain $\alpha$ to approximate the indicator function in Eq. \ref{eq:indicator}. 
Because the emissivity is at its maximum only at the first intersection point and is negligible for the rest of the ray, we propose to use $softmax$ which is a differentiable and moderately variable manner.
In order to handle the situation of background in the dataset, we propose an epipolar-based expectation mechanism $\beta$. We hypothesize an epipolar point $\mathbf{p}_e$ at infinity with a raw geometry attribute coefficient $s_e$ and the same color $c_e$ as the background. When the inputs are all zeros, the outputs after $softmax$ calculation are $1/N$. Therefore, we set $s_e=1/N$ to handle the case where the space traversed by the ray is completely empty.
The formula for geometry attribute coefficient of sampled points $\mathbf{w_r}=[w_1, w_2, ..., w_D]$ and geometry attribute coefficient of the epipolar point $w_e$ are shown in the following equation:
\begin{equation}
\begin{matrix}
 \mathbf{w}_{\mathbf{r},e} = \alpha (\mathbf{s}_{\mathbf{r},e}; \theta_\alpha)
 = softmax ( \theta_\alpha * \mathbf{s}_{\mathbf{r},e} ),
\label{eq:softmax}
\end{matrix}
\end{equation}
where $\: \mathbf{w}_{r,e} =  concat(\mathbf{w}_{r}, w_e), \,
\mathbf{s}_{\mathbf{r},e} = concat(\mathbf{s}_{r}, s_e)$. To avoid numerical instability when all inputs are small, we multiply the raw geometry attribute coefficient $\mathbf{s}_{\mathbf{r},e}$ by a rescale parameter $\theta_\alpha$. 
With $sigmoid$ as an activation function, Radiance Attribute Network $\rho_R$ is a two-layer MLP outputting radiance color $\mathbf{C_r}$ from ray feature embedding $\mathbf{V_r}$. Thus, the ray radiance is equivalent to the sum of the radiance variation of all sampling points and epipolar point
\begin{equation} 
\begin{aligned}
 C(\mathbf{r})  = \beta(\mathbf{w}_{\mathbf{r},e}, \mathbf{c}_{\mathbf{r},e})
 =  \sum\limits_{i=1}^N  (w_i * c_i) + w_e * c_e.
\end{aligned}
\label{eq:final}
\end{equation}

Compared to traditional NeRF methods, our approach is a pure neural rendering method that omits the need to compute intermediate opacity and accumulated opacity values, resulting in easy implementation and straightforward usage.

\subsection{Loss with Empty Space Regularization}
For both the coarse and refinement stages, we utilize the square of $\mathcal{L}_2$ loss function as the costs. To handle the situation that a ray does not intersect with the surface of an object, we employ $\mathcal{L}_1$ distance on the geometry attribute coefficient of the sampling points on the ray as an Empty Space Regularization $\mathcal{L}_e$, to make the raw geometry attribute coefficients tend to lie on the epipolar line. Our final loss function for the model can be expressed as
\begin{equation} 
\begin{aligned}
\mathcal{L} = 
\sum_{\mathbf{r} \in \mathbf{r}_c, \mathbf{r}_f}
 \lambda \left\|
C(\mathbf{r}) - \hat{C}(\mathbf{r})
\right\|^2_2
+
\lambda_w
\cdot
||\mathbf{w}_{\mathbf{r}}||_1, 
\label{eq:c_dev}
\end{aligned}
\end{equation}
where $\mathbf{w}_{\mathbf{r}} = \alpha[\rho_G(\rho_F(\mathbf{r}))]$.
$C(\mathbf{r})$ is the prediction and $\hat{C}(\mathbf{r})$ is the ground-truth. $\mathbf{r}_c$ corresponds to the sampled rays in the coarse stage, while $\mathbf{r}_f$ represents to the sampled rays in the fine stage. 

\section{Experiments}

\subsection{Experiments setting}

\paragraph{Datasets and metrics}

Our proposed method was evaluated using both the Blender dataset~\cite{Mildenhall2020NeRFRS} and the Shiny Blender dataset~\cite{verbin2022ref}. Both datasets provided 100 training images, 200 testing images, and 100 validation images with a resolution of $800 \times 800$ for each scene.
To assess the performance of our method, we employed three commonly used metrics in the field of novel view synthesis: Peak Signal-to-Noise Ratio (PSNR), Structural Similarity Index (SSIM)~\cite{Wang2004ImageQA}, and Learned Perceptual Image Patch Similarity (LPIPS)~\cite{Zhang2018TheUE}.

\begin{figure*}[t]
  \centering
  \begin{tabular}{ccccccccc}
    \includegraphics[width=0.11\linewidth]{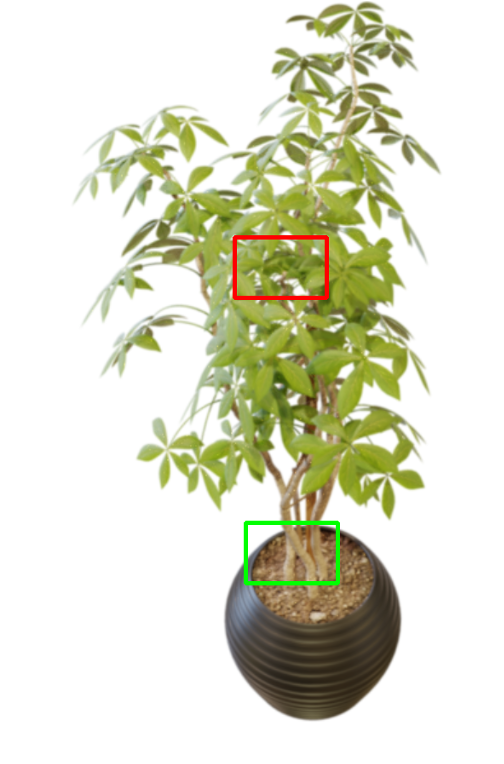} \hspace{-0.5cm} &
    \includegraphics[width=0.119\linewidth]{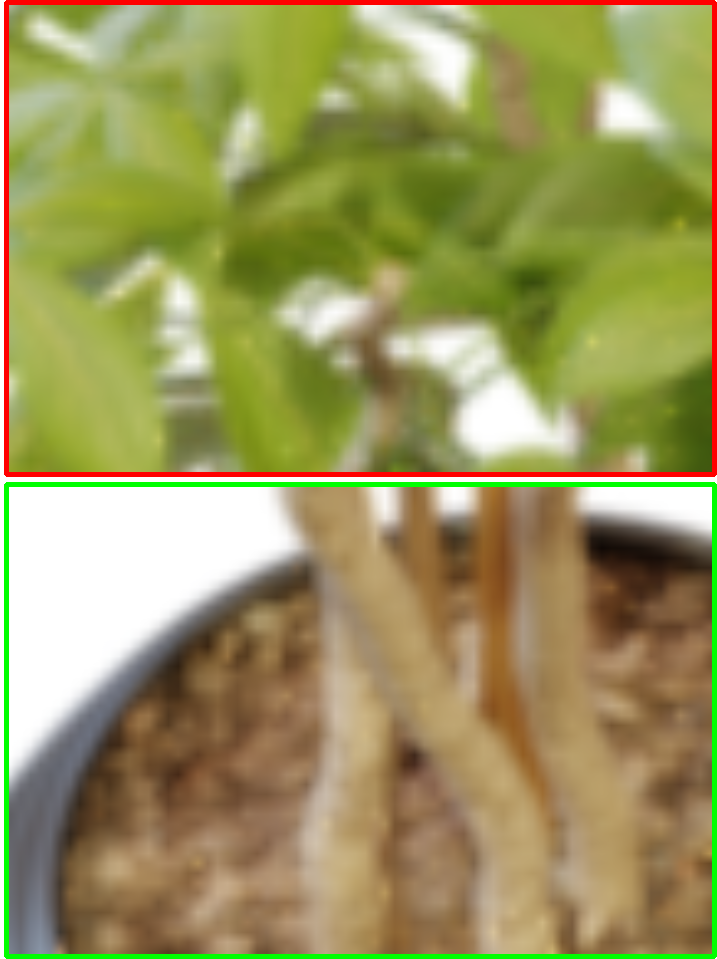} \hspace{-0.5cm} &
    \includegraphics[width=0.119\linewidth]{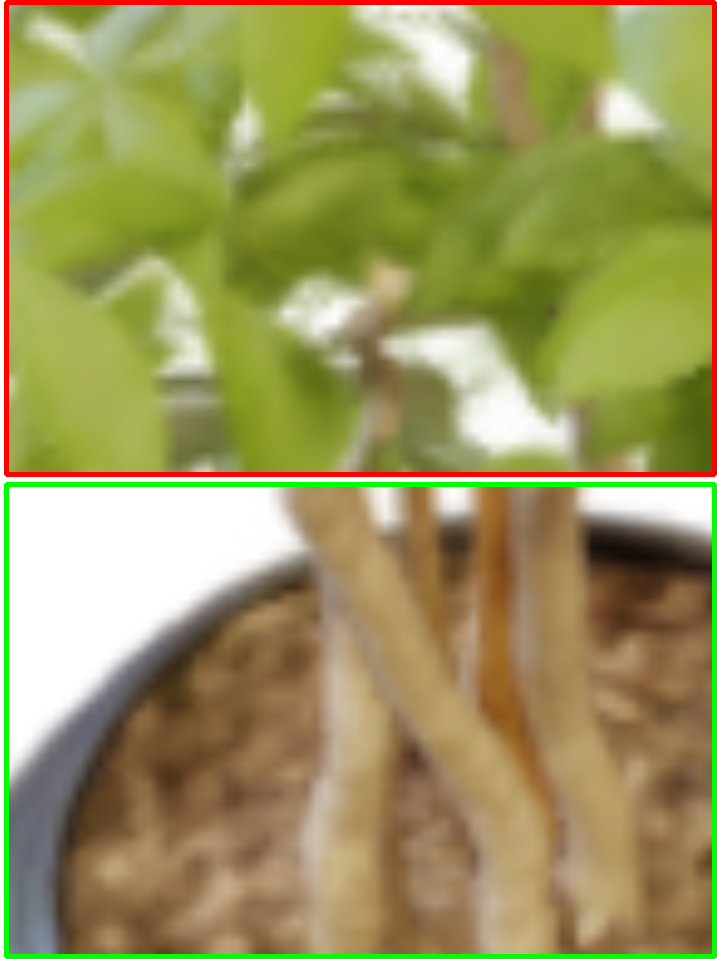} \hspace{-0.5cm} &
    \includegraphics[width=0.119\linewidth]{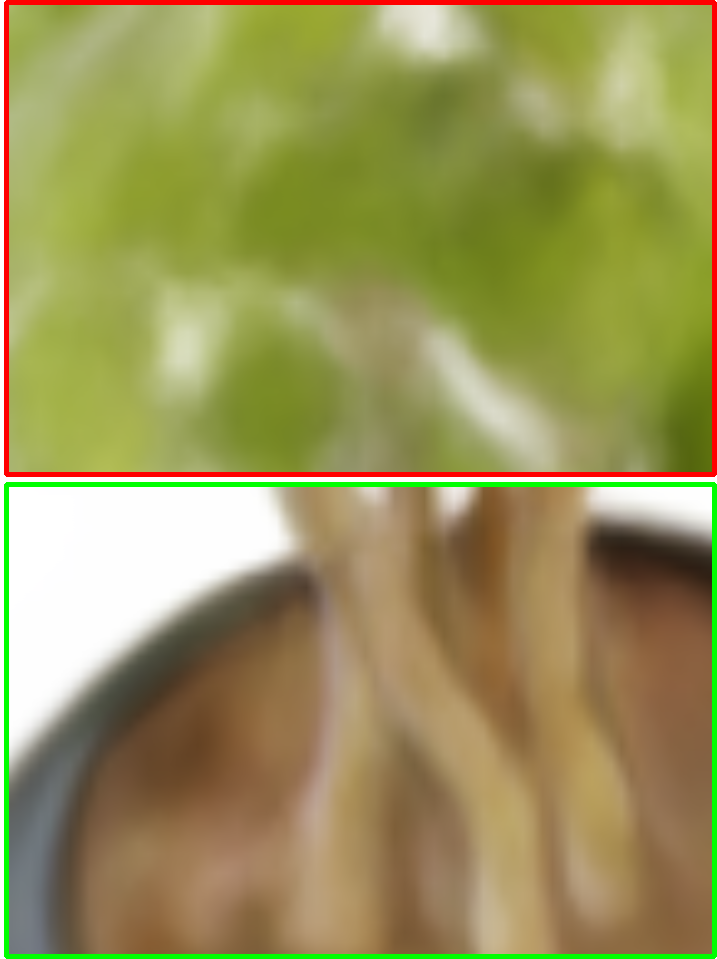} \hspace{-0.5cm} &
    \includegraphics[width=0.119\linewidth]{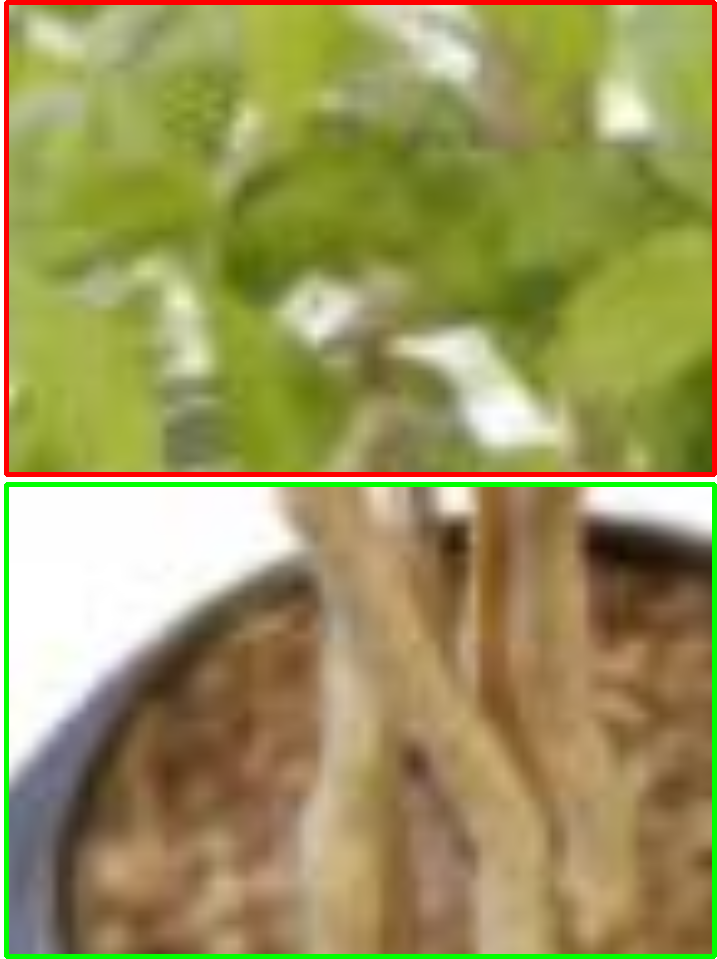} \hspace{-0.5cm} &
    \includegraphics[width=0.119\linewidth]{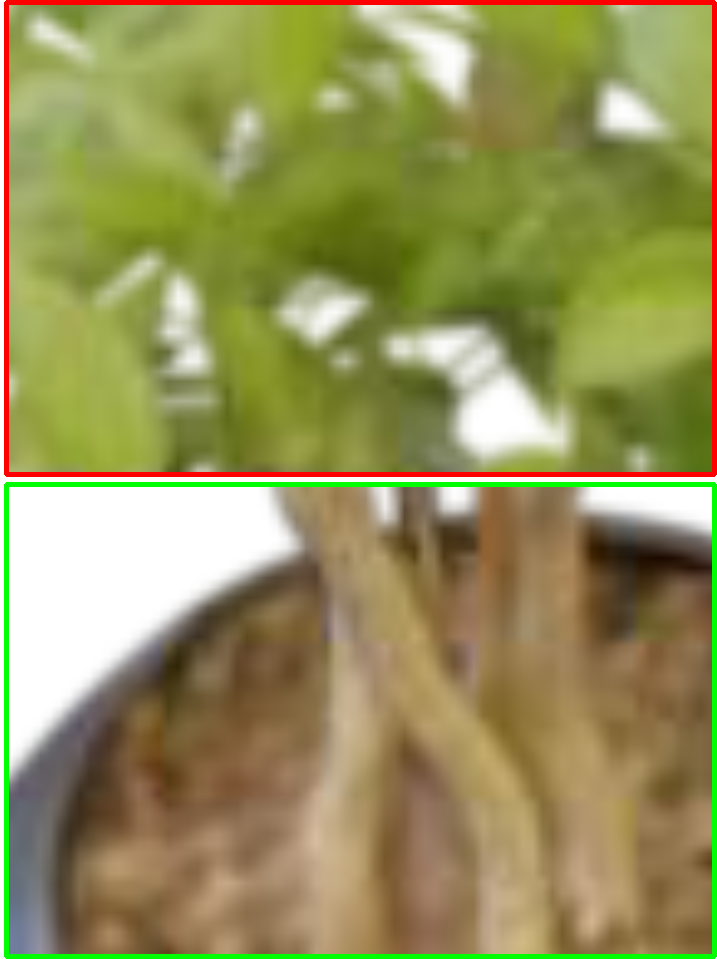} \hspace{-0.5cm} & 
    \includegraphics[width=0.119\linewidth]{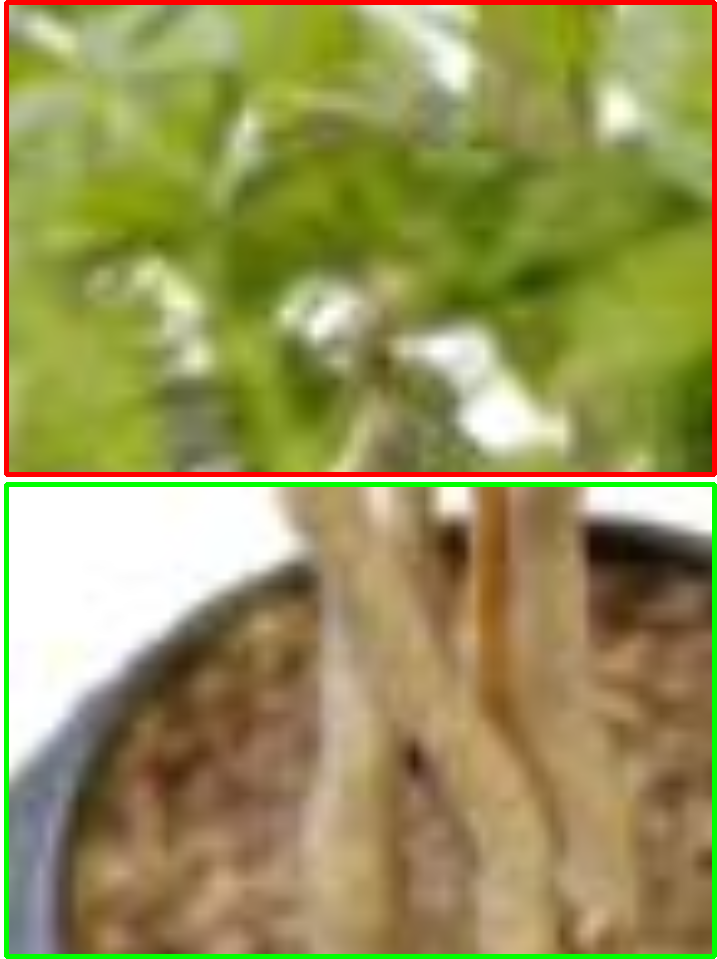} \hspace{-0.5cm} &
    \includegraphics[width=0.119\linewidth]{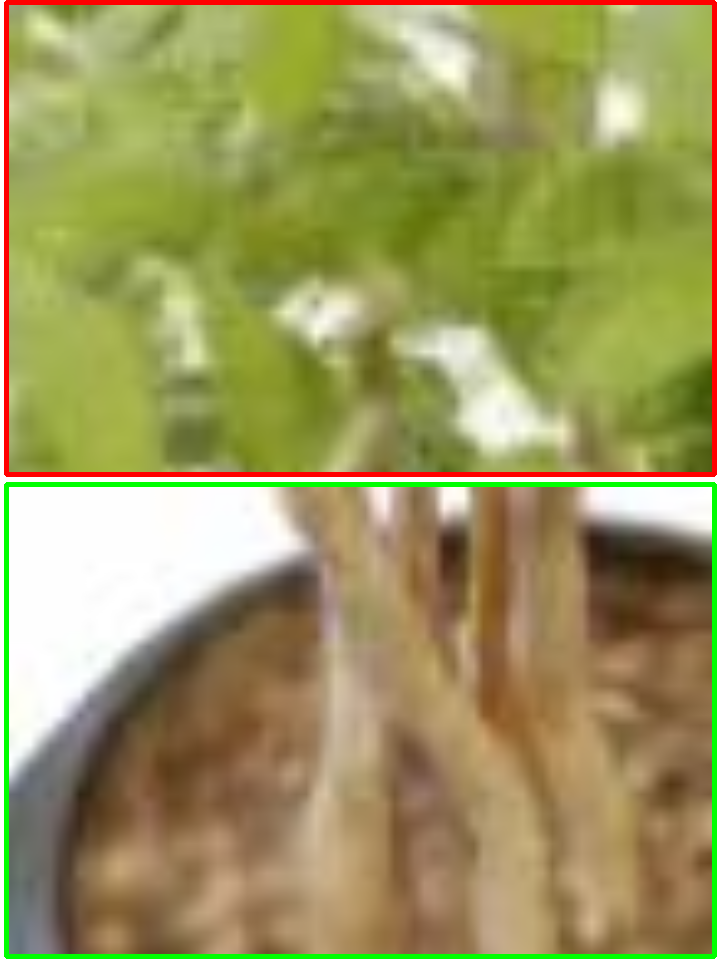} \hspace{-0.5cm} & 
    \vspace{-0.1cm}
    \\ 
    \includegraphics[width=0.12\linewidth]{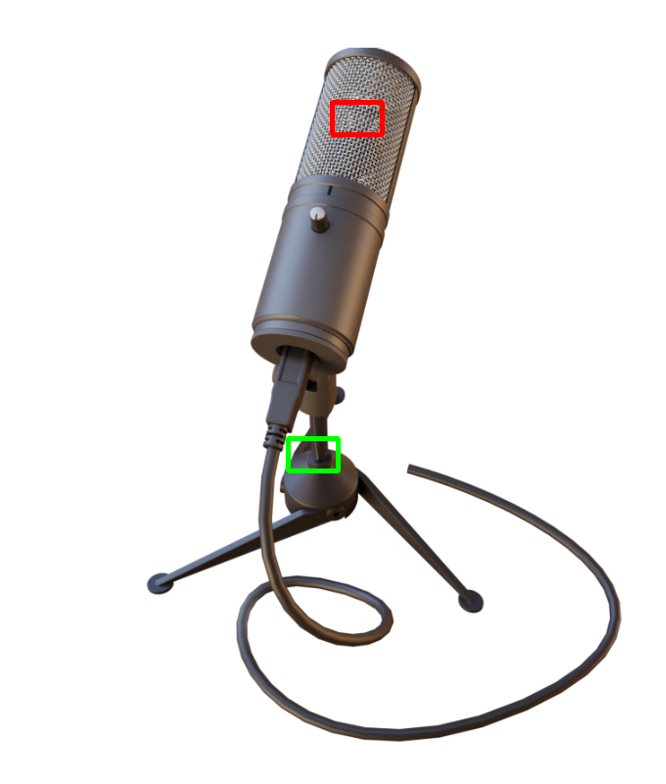} \hspace{-0.5cm} &
    \includegraphics[width=0.119\linewidth]{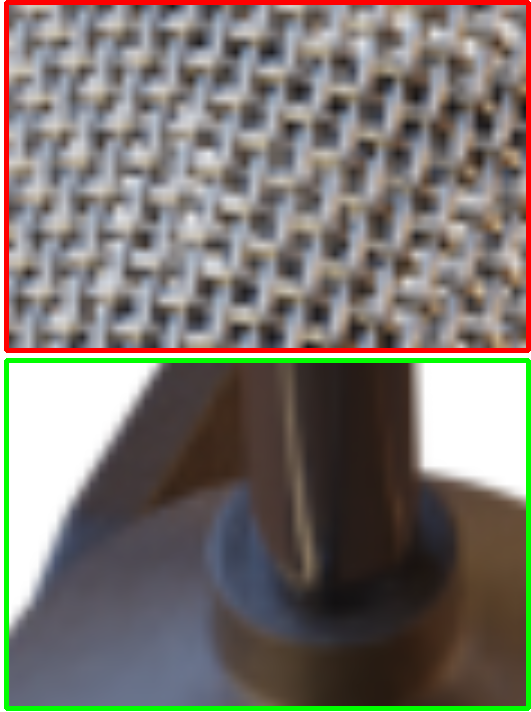} \hspace{-0.5cm} &
    \includegraphics[width=0.119\linewidth]{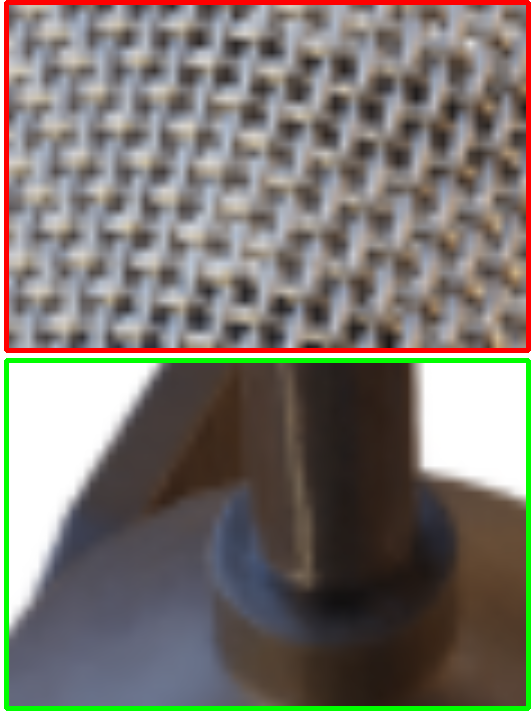} \hspace{-0.5cm} &
    \includegraphics[width=0.119\linewidth]{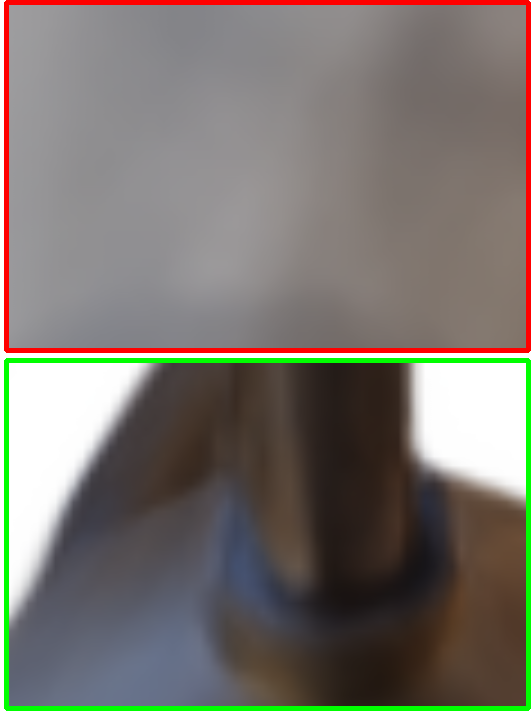} \hspace{-0.5cm} &
    \includegraphics[width=0.119\linewidth]{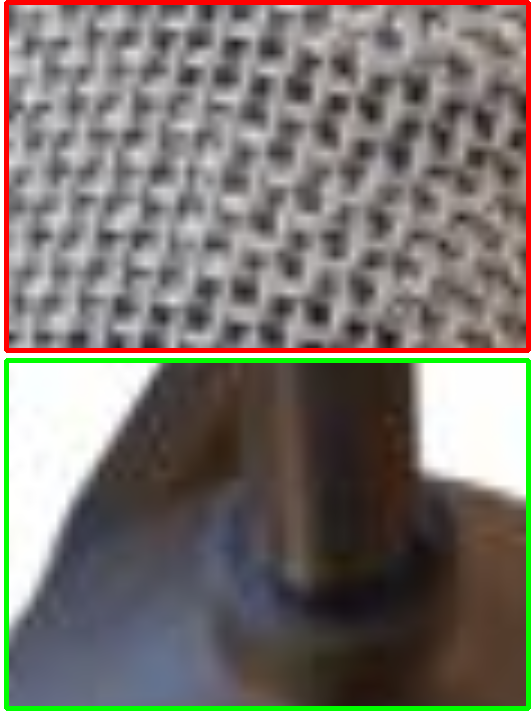} \hspace{-0.5cm} &
    \includegraphics[width=0.119\linewidth]{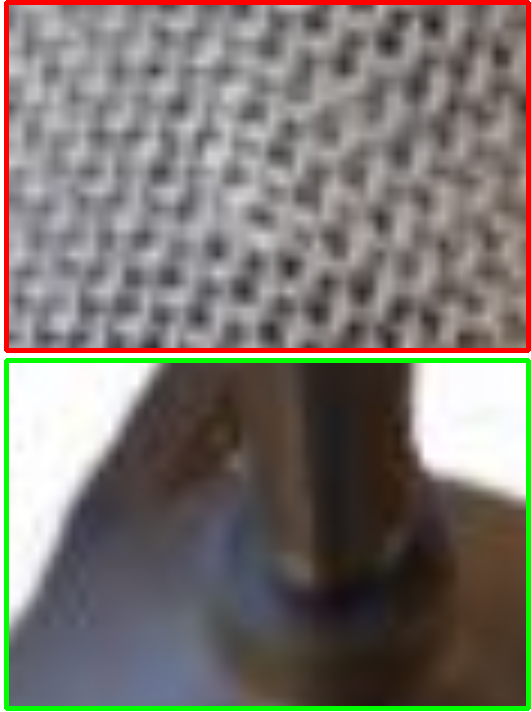} \hspace{-0.5cm} & 
    \includegraphics[width=0.119\linewidth]{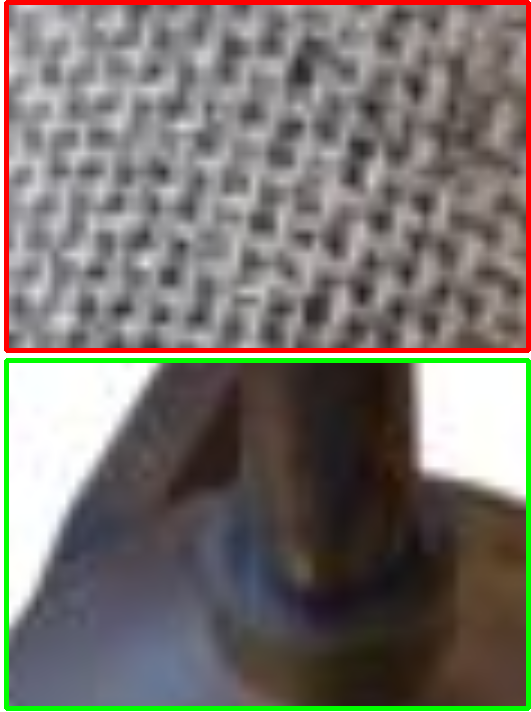} \hspace{-0.5cm} &
    \includegraphics[width=0.119\linewidth]{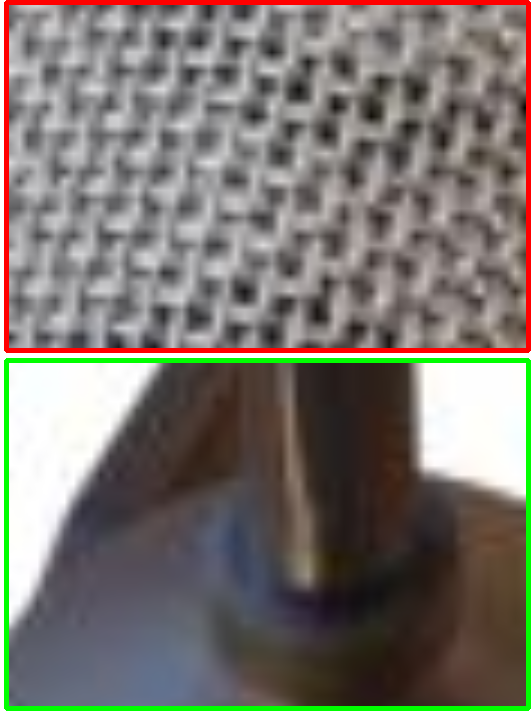} \hspace{-0.5cm} & 
    \vspace{-0.1cm}
    \\ 
    \includegraphics[width=0.12\linewidth]{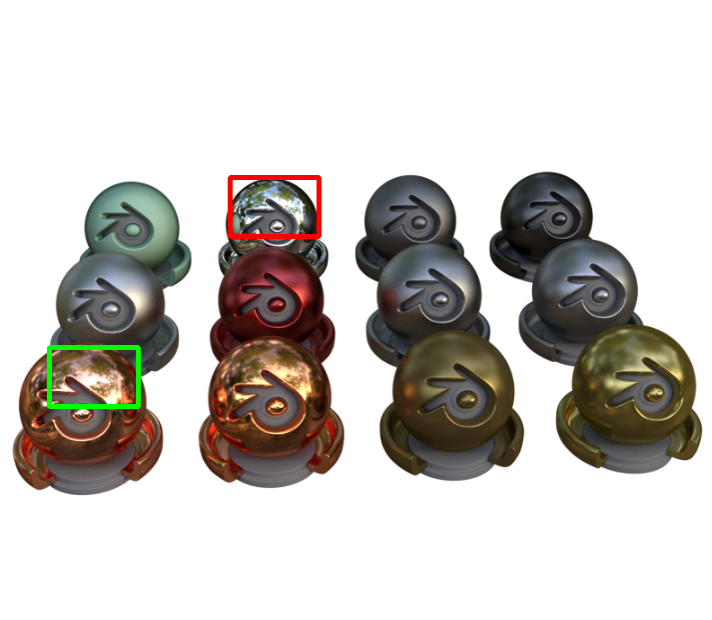} \hspace{-0.5cm} &
    \includegraphics[width=0.119\linewidth]{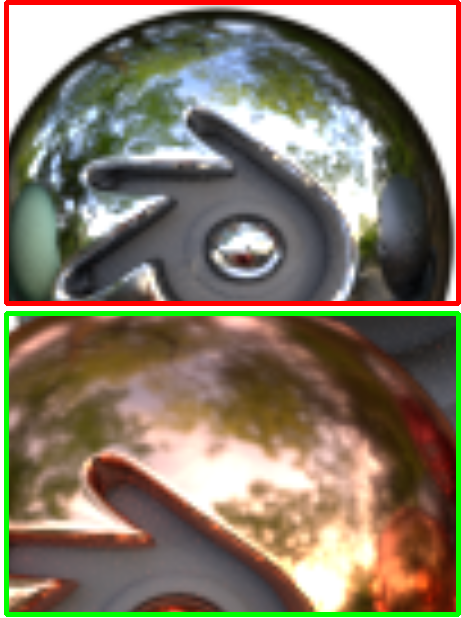} \hspace{-0.5cm} &
    \includegraphics[width=0.119\linewidth]{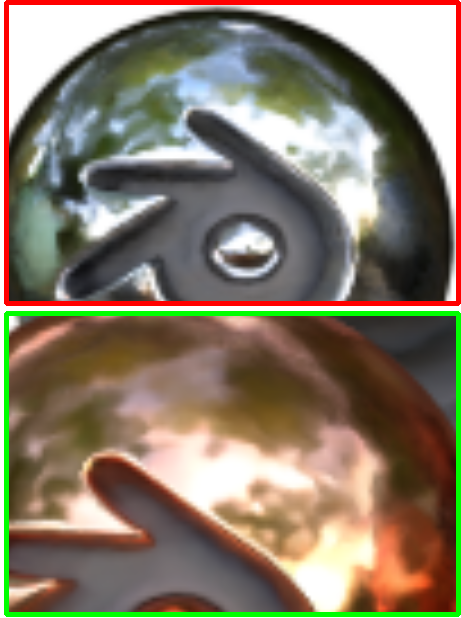} \hspace{-0.5cm} &
    \includegraphics[width=0.119\linewidth]{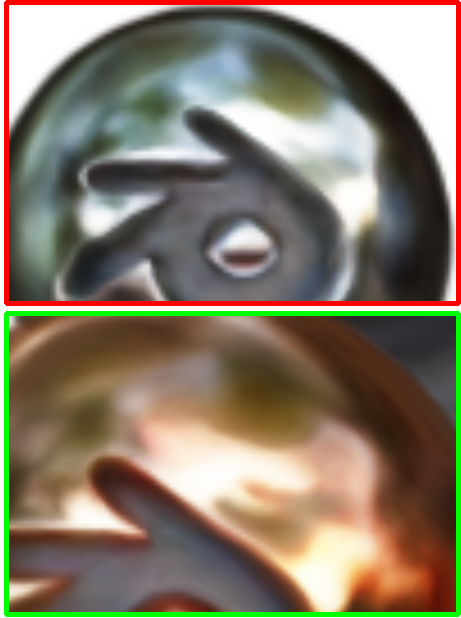} \hspace{-0.5cm} &
    \includegraphics[width=0.119\linewidth]{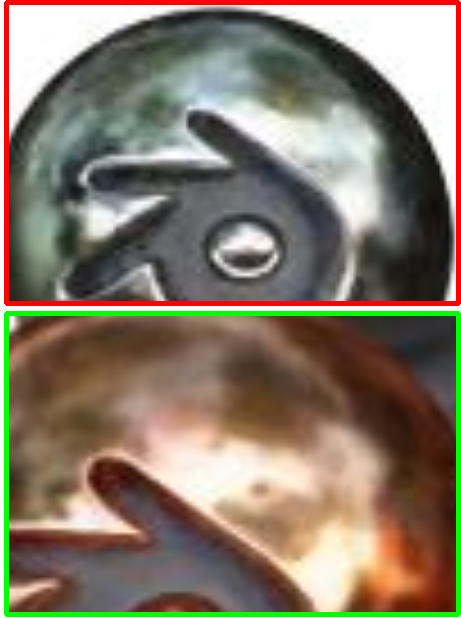} \hspace{-0.5cm} &
    \includegraphics[width=0.119\linewidth]{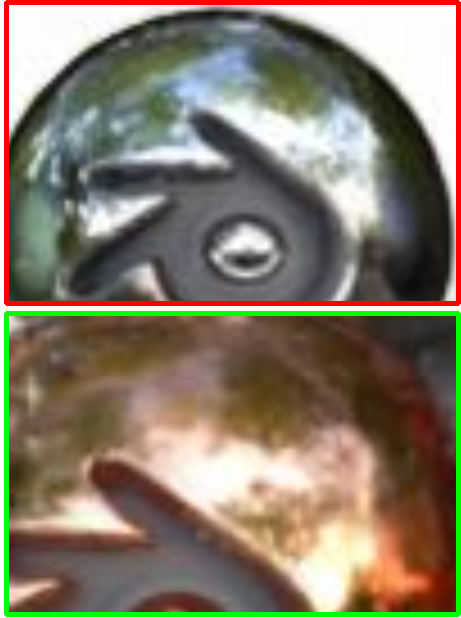} \hspace{-0.5cm} & 
    \includegraphics[width=0.119\linewidth]{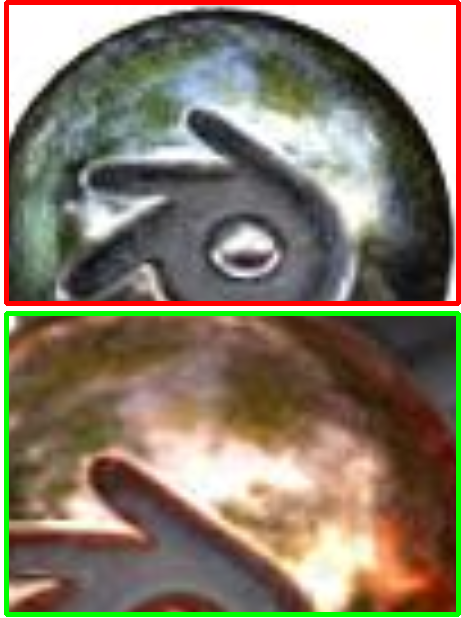} \hspace{-0.5cm} &
    \includegraphics[width=0.119\linewidth]{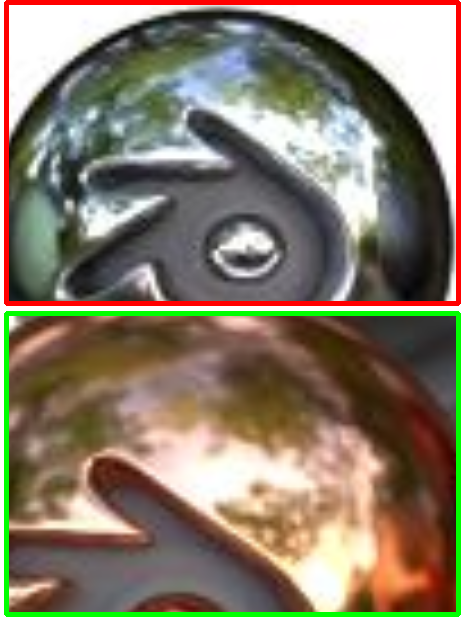} \hspace{-0.5cm} & 
    \vspace{-0.1cm}
    \\ 
    \scriptsize   & 
    \scriptsize  GT  & 
    \scriptsize  Ours & 
    \scriptsize  R2L & 
    \scriptsize  Mip-NeRF & 
    \scriptsize  NeuRay & 
    \scriptsize  Plenoxel & 
    \scriptsize  Ref-NeRF 
    \end{tabular}
    \caption{Qualitative results on the Blender dataset. CeRF excels at capturing complex geometric structures and fine details. In the ficus scene, our CeRF has the ability to accurately render the subsurface scattering of leaves and displays clear separation of layered and occluded branches. Despite Ref-NeRF performing well in scenes with strong specular reflections, it falls in accurately modeling light refraction and subsurface scattering. In the mic scene, CeRF demonstrates an aptitude for small mesh surfaces and accurately reconstructs subtle blue reflections on the base and small white highlights on the rods. In the metal ball scene, our specular reflections showcase rich and intricate details comparable to the results of Ref-NeRF without decoupled materials.}
    \label{fig:blender}
\end{figure*}

\begin{figure}[!t]
\centering
\begin{tabular}{ccccccccc}
\hspace{-0.6cm}
\includegraphics[width=0.2175\linewidth]{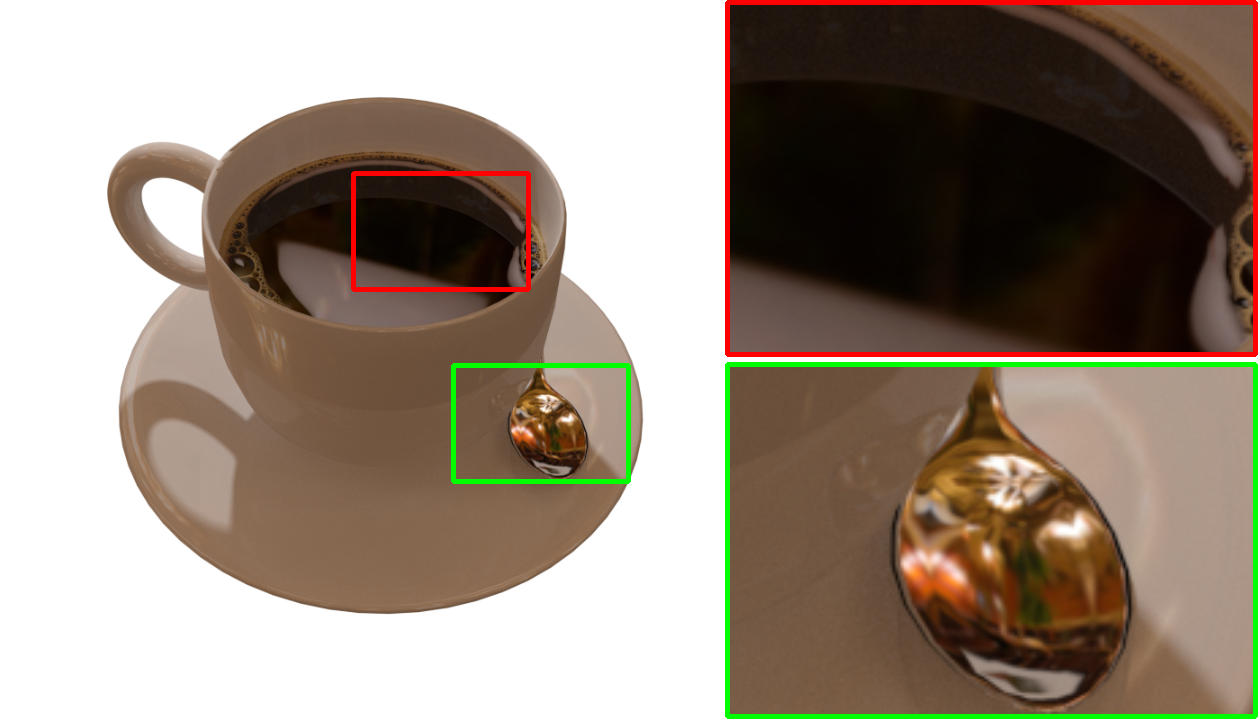} \hspace{-0.5cm} &
\includegraphics[width=0.093\linewidth]{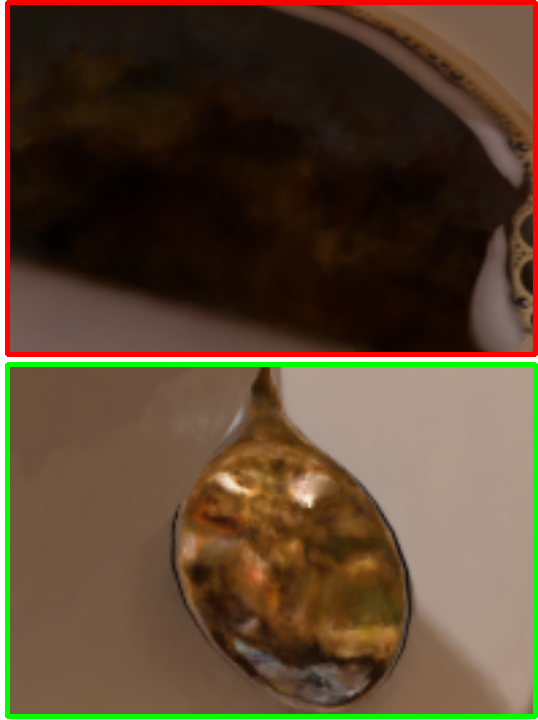} \hspace{-0.5cm} &
\includegraphics[width=0.093\linewidth]{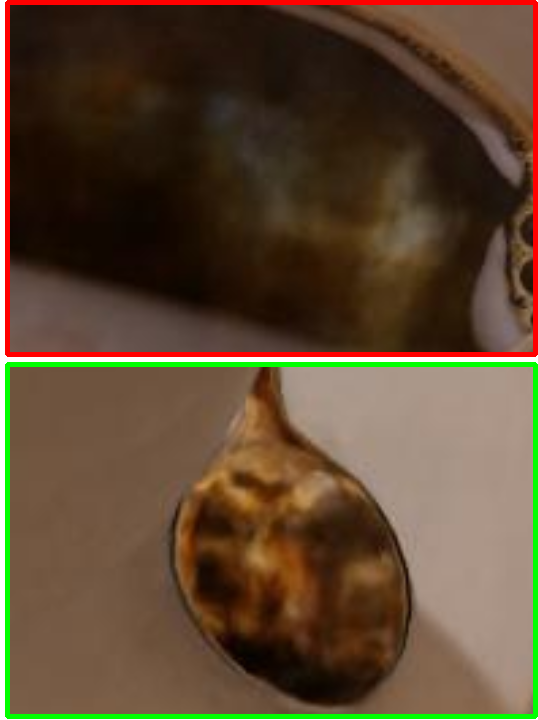} \hspace{-0.5cm} &
\includegraphics[width=0.093\linewidth]{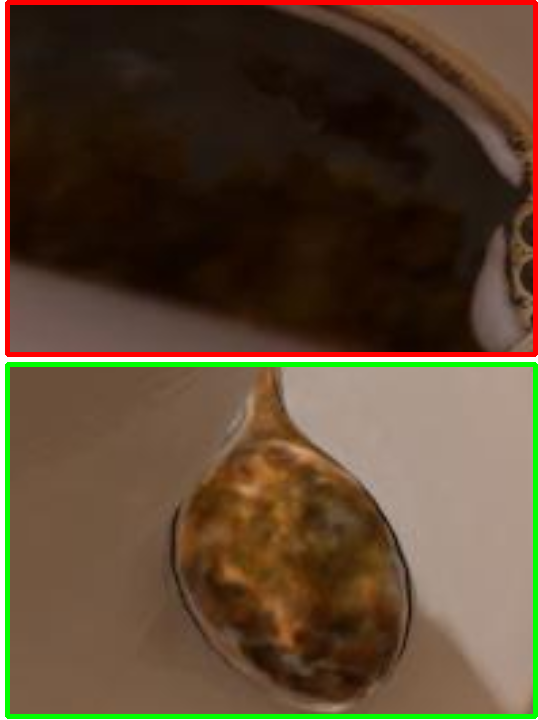} \hspace{-0.4cm} &

\includegraphics[width=0.243\linewidth]{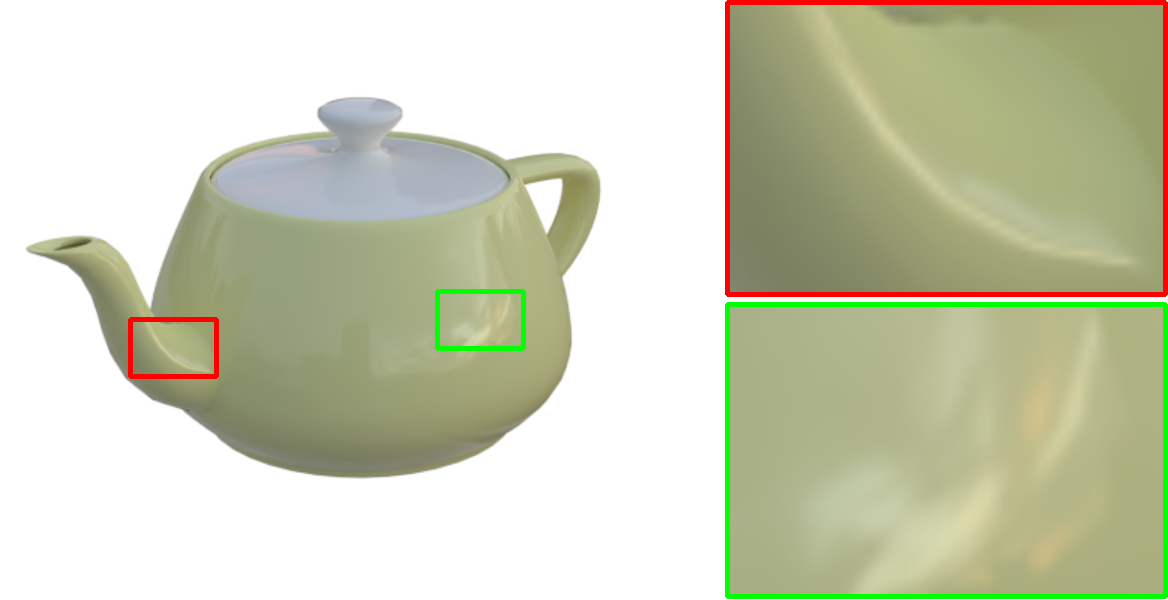} \hspace{-0.5cm} &
\includegraphics[width=0.093\linewidth]{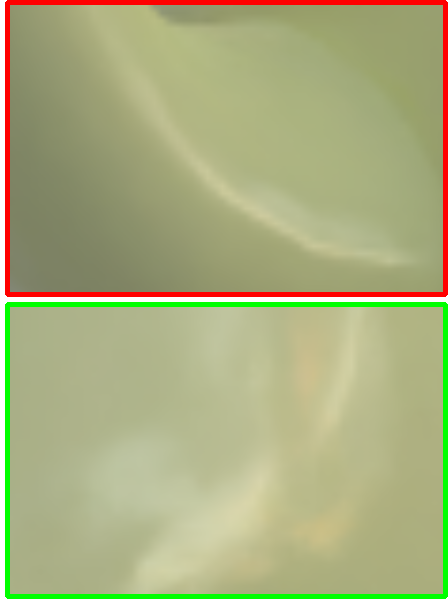} \hspace{-0.5cm} &
\includegraphics[width=0.093\linewidth]{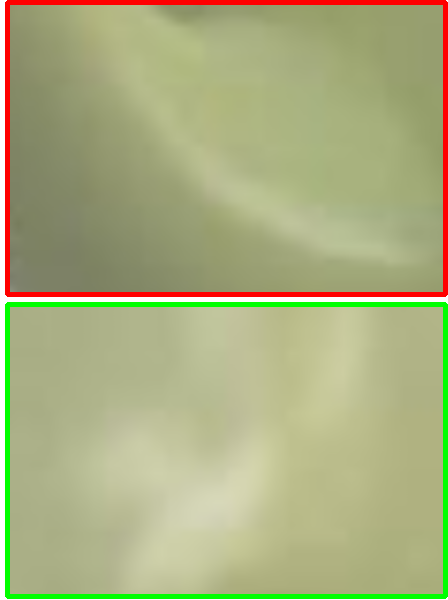} \hspace{-0.5cm} &
\includegraphics[width=0.093\linewidth]{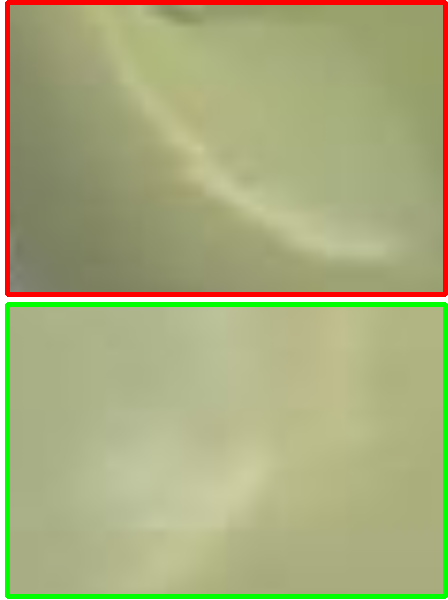} \hspace{-0.3cm} &
\\
\tiny \hspace{1.7cm} GT & 
\tiny Ours \hspace{-0.1cm} & 
\tiny Mip-NeRF \hspace{-0.3cm} & 
\tiny Ref-NeRF \hspace{-0.4cm} &
\tiny \hspace{2.5cm} GT & 
\tiny Ours \hspace{-0.1cm} & 
\tiny Mip-NeRF \hspace{-0.3cm} & 
\tiny Ref-NeRF \hspace{-0.4cm} &
\end{tabular}
\caption{Qualitative results on the Shiny Blender dataset. In the lack of explicitly modeling mirror reflections, CeRF produces better results for the reflections in the coffee and spoon. For the ceramic teapot, our method is capable of rendering high-gloss patterns with more precise edge contours.}
\label{fig:shiny}
\end{figure}

\paragraph{Hyperparameters setting}

The implementation of our CeRF is based on the Mip-NeRF~\cite{barron2021mipnerf} in PyTorch. 
To construct the networks, we use convolutional layers with a kernel size of 1 for all linear layers. In Convolutional Ray Feature Extractor, a convolutional layer with a kernel size of 3, a stride of 2 and padding of 1 is employed to perform downsampling in the ray direction. Linear interpolation is used in the upsampling layer and is followed by a layer of MLP to recover the channel size of the feature. The hidden state dimension of the GRU in $\rho_G$ is set to 64 with one recurrent layer. For $\alpha$, the rescale parameter $\theta_{\alpha}$ is set to 10. Moreover, the value of the raw geometry attribute coefficient $s_e$ at the epipolar is set to $1/N$, and the color $c_e$ value at the epipolar is set to the same white color as the background. These parameters are selected based on our experimentation and analysis of the results.

\paragraph{Training details}
We utilize the Adam optimizer with a logarithmically annealed learning rate that ranged from $2 \times 10^{-3}$ to $2 \times 10^{-5}$.
The coefficients used for computing the running averages of the gradient and its square are set to 0.8 and 0.888, respectively.
In our experiments, unless otherwise specified, we train the network using a batch size of 16384 for 500,000 iterations on 8 NIVIDIA RTX3080Ti GPUs. The weight $\lambda=0.1$ is assigned to the coarse stage, and it is assigned to 1 for the fine stage. The suggested weight for the Empty Space Regularization $\lambda_w$ is set to $0.01$. 

\subsection{Comparison with State-of-the-Art Methods}

In this section, we compare our CeRF with other neural field representations \footnote{Results from \url{https://huggingface.co/nrtf/nerf_factory}.}, including NeuRay~\cite{Liu2021NeuralRF}, Mip-NeRF~\cite{barron2021mipnerf}, Plenoxel~\cite{Yu2021PlenoxelsRF}, Ref-NeRF~\cite{Verbin2021RefNeRFSV}, as well as R2L~\cite{wang2022r2l}. The qualitative and quantitative results  on the Blender dataset are presented in Figure \ref{fig:blender} and Table \ref{tb:blender}, respectively. 
Compared to the baseline, our method outperforms in all metrics on the Blender dataset. Without requiring complex distillation training, CeRF achieves an $8.5\%$ increase in PSNR compared to R2L. Furthermore, CeRF also outperforms the SOTA, Ref-NeRF, in all three metrics.

Figure \ref{fig:shiny} and Table \ref{tb:shiny} demonstrate the qualitative and quantitative results of the Shiny Blender dataset. CeRF achieves the second rank after Ref-NeRF and outperforms all the other methods. However, our method still outperforms the current state-of-the-art method, Ref-NeRF, in scenes with subsurface reflections, such as coffee and teapot.
Furthermore, we choose not to model reflections because our goal was to create a model that is as simple and general as possible, and we opt for a pure implicit representation. Moreover, we believe that designing structures with separated materials, such as Ref-NeRF~\cite{Verbin2021RefNeRFSV} and I2-SDF~\cite{zhu20232}, can further enhance network performance.

\begin{table}[!t]
\centering
  \begin{minipage}[b]{0.33\linewidth}
    \caption{Comparison with SOTA on the Blender dataset}
    \label{tb:blender}
    \centering
    \resizebox{\linewidth}{!}{
    \begin{tabular}{c|ccc}
        \toprule
        & PSNR$\uparrow$ & SSIM$\uparrow$ & LPIPS$\downarrow$ \\ 
        \midrule  \midrule
        Plenoxel~\cite{Yu2021PlenoxelsRF} & 31.71 & 0.958 & 0.049  \\
        R2L~\cite{wang2022r2l} & 31.87 & \cellcolor[RGB]{244,181,180} 0.995 & \cellcolor[RGB]{244,181,180}0.034  \\ 
        NeuRay~\cite{Liu2021NeuralRF} & 32.35 & 0.960 & 0.048  \\
        Mip-NeRF~\cite{barron2021mipnerf} & \cellcolor[RGB]{255,255,187} 33.11 & 0.962 & 0.042  \\
        Ref-NeRF~\cite{Verbin2021RefNeRFSV} & \cellcolor[RGB]{249,218,183}33.99 & \cellcolor[RGB]{255,255,187} 0.966 & \cellcolor[RGB]{255,255,187}0.038 \\
        CeRF & \cellcolor[RGB]{244,181,180}34.57 & \cellcolor[RGB]{249,218,183} 0.969 & \cellcolor[RGB]{244,181,180}0.034 \\ 
    \bottomrule
    \end{tabular}
    }
  \end{minipage}
  \begin{minipage}[b]{0.34\linewidth}
    \caption{Comparison with SOTA on the Shiny Blender dataset}
    \label{tb:shiny}
    \centering
    \resizebox{\linewidth}{!}{
    \begin{tabular}{c|ccc}
        \toprule
        & PSNR$\uparrow$ & SSIM$\uparrow$ & LPIPS$\downarrow$ \\ 
        \midrule  \midrule
        Plenoxel~\cite{Yu2021PlenoxelsRF} & 28.81 & 0.8967 & 0.1557  \\
        DVGO~\cite{Sun2022ImprovedDV} & 29.87 & 0.9258 & 0.1444  \\
        PhySG~\cite{Zhang2021PhySGIR}  & 26.2133 & 0.9212 & 0.2077   \\
        Mip-NeRF~\cite{barron2021mipnerf} & \cellcolor[RGB]{255,255,187} 29.7600 & \cellcolor[RGB]{255,255,187} 0.9417 & \cellcolor[RGB]{255,255,187} 0.0920  \\
        Ref-NeRF~\cite{Verbin2021RefNeRFSV} & \cellcolor[RGB]{244,181,180} 35.9617 & \cellcolor[RGB]{244,181,180} 0.9670 & \cellcolor[RGB]{244,181,180} 0.0587  \\
        CeRF & \cellcolor[RGB]{249,218,183} 33.6015 & \cellcolor[RGB]{249,218,183} 0.9644 & \cellcolor[RGB]{249,218,183} 0.0597  \\ 
    \bottomrule
    \end{tabular}
    }
  \end{minipage}
  \begin{minipage}[b]{0.317\linewidth}
  \caption{Ablation study on the drum scene}
  \label{tb:ablation}
  \centering
  \resizebox{\linewidth}{!}{
    \begin{tabular}{c|ccc}
    \toprule
    & PSNR$\uparrow$ & SSIM$\uparrow$ & LPIPS$\downarrow$ \\ 
    \midrule  \midrule
    baseline & 26.4104 & 0.9420 & 0.0500  \\
    w/o $\rho_F$ & 26.7711 & 0.9453 & 0.0467  \\
    w/o $\rho_G$ & \cellcolor[RGB]{255,255,187}27.1833 & 0.9497 & 0.0427  \\ 
    w/o $\alpha$ & 23.6040 & 0.8996 & 0.1546  \\ 
    w/o $\beta$ & 27.1475 & \cellcolor[RGB]{255,255,187}0.9509 & \cellcolor[RGB]{255,255,187}0.0411  \\
    w/o $\mathcal{L}_e$ & \cellcolor[RGB]{249,218,183}27.3703 & \cellcolor[RGB]{244,181,180}0.9533 & \cellcolor[RGB]{244,181,180}0.0391 \\
    CeRF & \cellcolor[RGB]{244,181,180}27.3900 & \cellcolor[RGB]{249,218,183}0.9526 & \cellcolor[RGB]{244,181,180}0.0391  \\ 
    \bottomrule
    \end{tabular}
  }
  \end{minipage}
\end{table}

\begin{figure}[!t]
\centering
\begin{tabular}{ccccccccc}
\hspace{-0.6cm}
\includegraphics[width=0.358\linewidth]{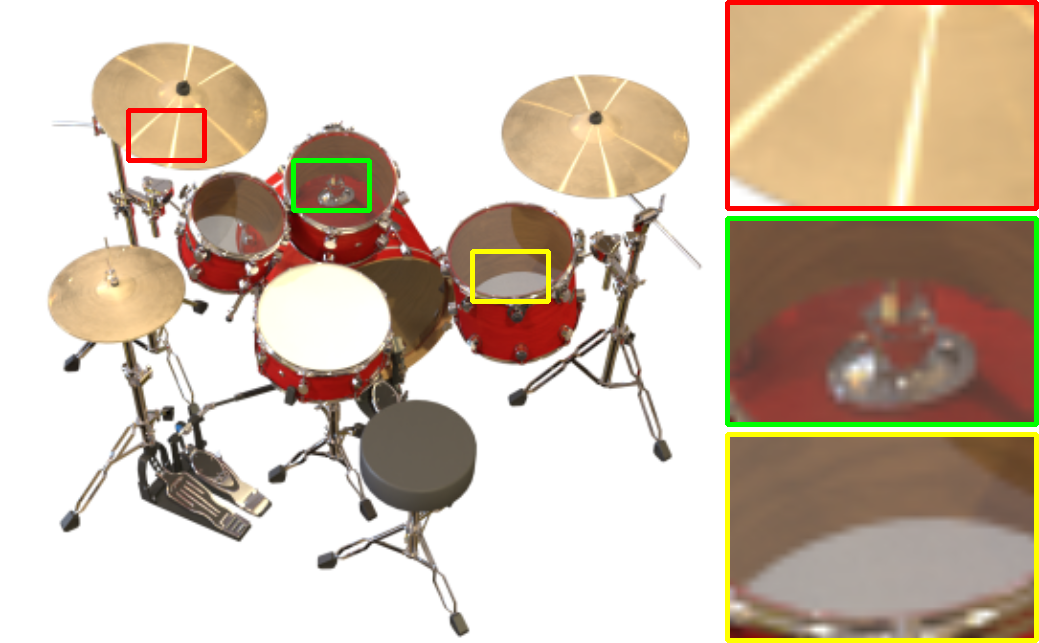} &
\hspace{-0.5cm}
\includegraphics[width=0.11\linewidth]{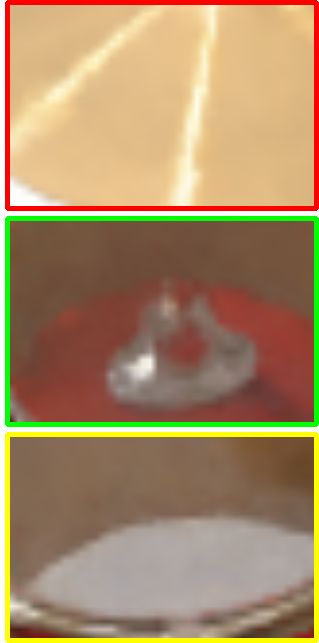} &
\hspace{-0.5cm}
\includegraphics[width=0.11\linewidth]{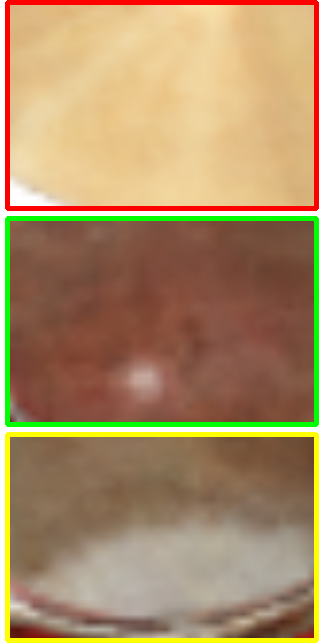} &
\hspace{-0.5cm}
\includegraphics[width=0.11\linewidth]{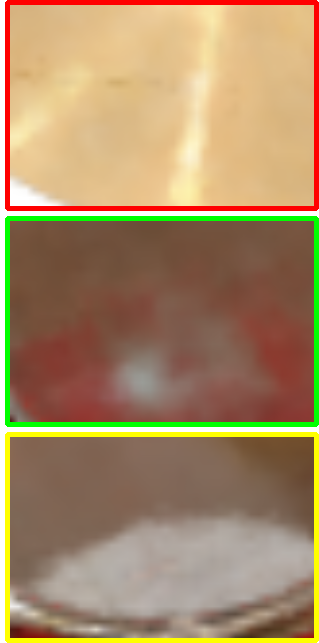} &
\hspace{-0.5cm}
\includegraphics[width=0.11\linewidth]{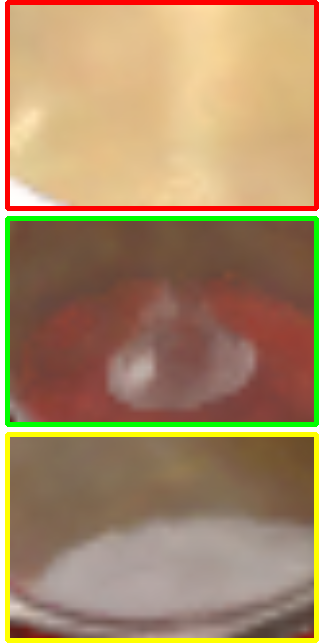} &
\hspace{-0.5cm}
\includegraphics[width=0.11\linewidth]{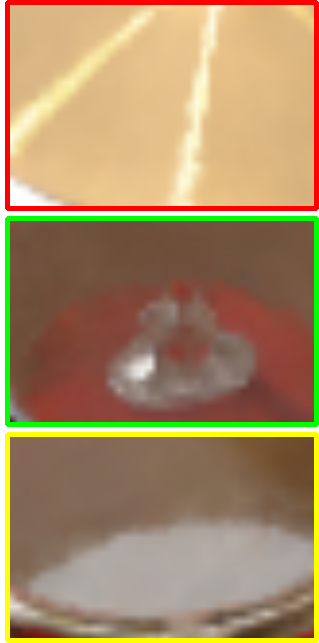} &
\hspace{-0.5cm}
\includegraphics[width=0.11\linewidth]{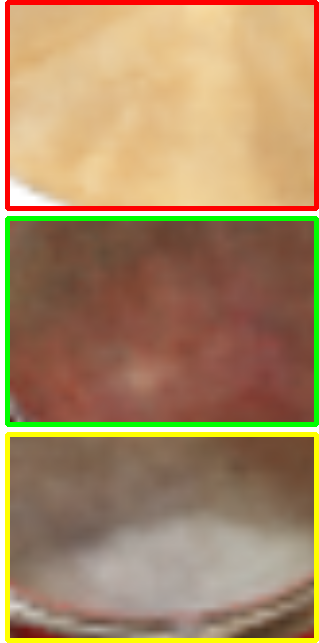} &
\\
\hspace{-1cm}
\hspace{3.8cm} \scriptsize GT & 
\hspace{-0.5cm} \scriptsize Ours & 
\hspace{-0.5cm} \scriptsize w/o $\rho_F$ & 
\hspace{-0.5cm} \scriptsize w/o $\rho_G$  &
\hspace{-0.5cm} \scriptsize w/o $\alpha$ &
\hspace{-0.5cm} \scriptsize w/o $\beta$ &
\hspace{-0.5cm} \scriptsize w/o $\rho_F \& \rho_G$ &
\\
\end{tabular}
\caption{Ablation studies on drums scene 
}
\label{fig:lego}
\end{figure}

\subsection{Ablation studies}

In order to evaluate the effectiveness of the individual components in our proposed CeRF model, we conducted ablation experiments on the drum scene using half resolution, a batch size of 14,744, and an iteration step size of 250,000. Figure \ref{fig:lego} and Table \ref{tb:ablation} present the qualitative and quantitative results of our ablation experiments, which demonstrate the effectiveness of our proposed modules from both a metric and visual perspective.

The absence of the \textbf{Convolutional Ray Feature Extractor $\rho_F$} results in the loss of directionally-correlated visual representations, making it difficult to extract the specular properties of the scene. Removing the \textbf{Radiance Attribute Network $\rho_G$} causes obstacles in capturing information about the sampled light point sequences and results in noise on the rendered object surface.
We found that removing the \textbf{Unique Surface Constraint} $\alpha$ gives the worst results, as the radiance of the highlight cannot be correctly reconstructed. Ablation experiments based on the \textbf{Epipolar Expectation} show that if $\beta$ is removed, the model will struggle to accurately compute the radiance of the empty space that rays pass through in the background of the dataset. This would have a negative impact on the accuracy and quality of the rendered output. In the absence of both the $\rho_F$ and $\rho_G$ modules, the mirror reflections in the drum scene become unacceptably blurred, as shown in Figure \ref{fig:lego}.
The \textbf{Empty Space Regularization} $L_e$ ensures the correct matching of empty rays on epipolar lines. Although the metric may slightly decrease, we demonstrate through depth visualization that regularization is important in fitting geometry at infinity, which is described in the supplementary materials.

\subsection{Parameters experiments}

In this subsection, we conduct a series of experiments to explore the hyperparameters configurations for $\rho_F$, $\rho_G$, $\alpha$, and $\beta$. The testing was performed on a half-resolution lego scene with an iteration step of 250,000. The notations in Table \ref{tb:abl_table_2}, such as W128U4K3D8, signify a setting with a width of 128, 4 layers of up-sampling and down-sampling, a kernel size of 3, and a total module depth of 8.

Our experimental results, as presented in Table \ref{tb:abl_table_2}, demonstrate that our parameter settings achieve a balance between resource consumption and effectiveness. Although the setting of a total module depth of 8 achieves better results, it requires significantly more computational resources.

Table \ref{tb:abl_table_3} displays the results of scaling $\theta_{\alpha}$ and $s_e$ in CeRF. We proposed a new mechanism to achieve learnable $s_e$. However, experiments showed that this method was not effective, and thus fixed $s_e$ was used instead. Further details can be found in the supplementary materials.


\begin{table}[!t]
  \vspace{-0.3cm}
  \begin{minipage}[b]{0.557\linewidth}
    \caption{Different $\mathbf{\rho_F}$ setting on the drum scene}
    \label{tb:abl_table_2}
    \centering
    \resizebox{\linewidth}{!}{
    \begin{tabular}{c|ccccc}
    \toprule
    & PSNR$\uparrow$ & SSIM$\uparrow$ & LPIPS$\downarrow$ & \makecell[c]{GPU Mem \\(GB)$\downarrow$} & \makecell[c]{Training \\ Speed (s/epoch)$\downarrow$} \\ 
    \midrule  \midrule
    W128U4K3D8 & 26.2041 & 0.9387 & 0.0580 & 8.649 & 190  \\
    W256U4K5D8 & 25.4161 & 0.9289 & 0.0739 & 11.109 & 300  \\
    W256U2K3D8 & 27.1174 & 0.9497 & 0.0424 & 11.867 & 350  \\
    W256U6K3D8 & \cellcolor[RGB]{255,255,187} 27.3028 & \cellcolor[RGB]{255,255,187} 0.9521 & \cellcolor[RGB]{255,255,187} 0.0403  & 10.577 & 330  \\
    W256U4K3D6 & 27.1732 & 0.9507 & 0.0417 & 9.351 & 310  \\ 
    W256U4K3D10 & \cellcolor[RGB]{244,181,180} 27.4342 & \cellcolor[RGB]{244,181,180} 0.9538 & \cellcolor[RGB]{244,181,180} 0.0379 & 11.849 & 380  \\ 
    W256U4K3D8(ours) & \cellcolor[RGB]{249,218,183} 27.3900 & \cellcolor[RGB]{249,218,183} 0.9526 & \cellcolor[RGB]{249,218,183} 0.0391 & 11.039 & 340  \\
    \bottomrule
    \end{tabular}
    }
  \end{minipage}
  \begin{minipage}[b]{0.428\linewidth}
    \caption{ $\rho_G$ variance on the lego scene} 
    \label{tb:abl_table_3}
    \centering
    \resizebox{\linewidth}{!}{
    \begin{tabular}{c|ccc}
    \toprule
    & PSNR$\uparrow$ & SSIM$\uparrow$ & LPIPS$\downarrow$ \\ 
    \midrule  \midrule
    $\theta_{\alpha}=1 $ & 23.0863 & 0.8916 & 0.1452 \\
    $\theta_{\alpha}=20$ & 27.2061 & \cellcolor[RGB]{244,181,180} 0.9526 & \cellcolor[RGB]{249,218,183} 0.0394 \\
    learnable $\theta_{\alpha}$ & \cellcolor[RGB]{249,218,183} 27.3559 & \cellcolor[RGB]{244,181,180} 0.9526 & \cellcolor[RGB]{249,218,183} 0.0394 \\
    $s_e=2/N$ & \cellcolor[RGB]{255,255,187} 27.2547 & 0.9518 & 0.0402 \\
    learnable $s_e$ & 14.9650 & 0.8358 & 0.1665 \\
    $\theta_{\alpha}=10, s_e=1/N$ & \cellcolor[RGB]{244,181,180} 27.3900 & \cellcolor[RGB]{244,181,180} 0.9526 & \cellcolor[RGB]{244,181,180} 0.0391 \\
    \bottomrule
    \end{tabular}
    }
  \end{minipage}
  \vspace{-0.3cm}
\end{table}

\section{Conclusion and discussion}

In conclusion, we proposed CeRF as a novel approach for new view synthesis based on the derivatives of the radiance model. Our proposed model, with a CNN-based feature extractor and GRU-based neural rendering, has reduced the learning difficulties for implicit representation and recovered high-quality rendering results even under complex geometry.
Moreover, CeRF can be easily integrated with other deep learning frameworks, making it a promising alternative to NeRF for a variety of high-level tasks. Extensive experiments showed that our proposed CeRF yields promising results with a simple network structure.

\textbf{Discussion.} While CeRF extends linear layers to convolution layers, CNNs are still a simple and preliminary type of network. There are still many ways to optimize convolutional operations, as well as utilizing superior structure like Transformers to improve current models.

\textbf{Limitations.} The convolution operation increases the computational complexity, and the RNN introduces more parameters. Therefore, the calculation speed and cost are higher than NeRF. There is no modeling of physically-based materials, nor is there any supervision or constraints placed on the geometry, such as depths and normals, or extended to semantic information. Further future work is needed to demonstrate the generality of this framework.

\textbf{Negative Social Impact.} The training of the program requires a long time to run on a GPU, and the power consumption is not environmentally friendly.


{\small
\bibliographystyle{ieee_fullname}
\bibliography{egbib}
}

\newpage
\appendix

\setcounter{equation}{0}
\setcounter{figure}{0}
\setcounter{table}{0}

\nolinenumbers

\vspace{10mm}
\section*{\centering  \LARGE Supplementary Material for \\ CeRF: Convolutional Neural Radiance Fields for New View Synthesis with Derivatives of Ray Modeling }
\vspace{10mm}

\linenumbers
\setcounter{linenumber}{1}
\pagenumbering{arabic} 

\section{Empty Space Regularization}

NeRF utilizes classical volume rendering to render the 3D radiance field into a 2D image, as

\begin{equation}
    C(\mathbf{r})=\sum_{i=1}^N T_i\left(1-\exp \left(-\sigma_i \delta_i\right)\right) \mathbf{c}_i, \text { where } T_i=\exp \left(-\sum_{j=1}^{i-1} \sigma_j \delta_j\right),
\end{equation}

where $\sigma_i$ is the volume density of the ith point along the ray, $\delta_i$ is the distance between adjacent samples.
The accumulated transmittance $_i$ introduced in the formula constrains the sum of the volume density over the ray to be no greater than 1 and allows NeRF to handle the occlusion relationship between objects to some extent. In cases where rays pass through empty space, NeRF can learn that the volume density at all sampled points is 0, resulting in a total sum of 0 for the entire ray.

CeRF proposes a Neural Radiance Ray Rendering technique that can effectively handle complex layers between objects. Unlike the rendering method used in NeRF, the neural renderer does not explicitly model the rendering function, allowing it to fit more complex optical phenomena such as subsurface scattering. When dealing with rays in empty space, the neural renderer does not impose any constraints on the location of selected points along these rays. However, by implementing Unique Surface Constrain, which is implemented with  $softmax$, we force the sum of radiance coefficients for all sampled points on the rays in empty space to be 1 instead of learning an incorrect result where this sum is 0. In other words, to satisfy this constraint imposed by softmax, random points are chosen along these rays by the Neural Radiance Ray Renderer.

\begin{figure}[!t]
\centering
\begin{tabular}{ccccc}
\rotatebox{90}{\quad\quad\quad CeRF} &
\includegraphics[width=0.22\linewidth]{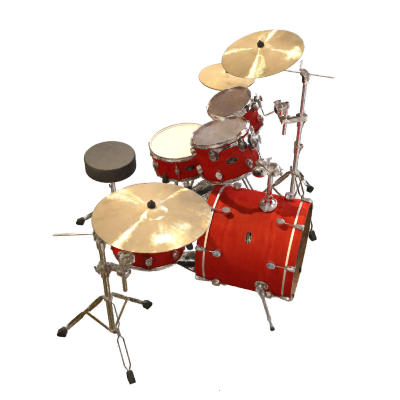} &
\hspace{-0.4cm}
\includegraphics[width=0.22\linewidth]{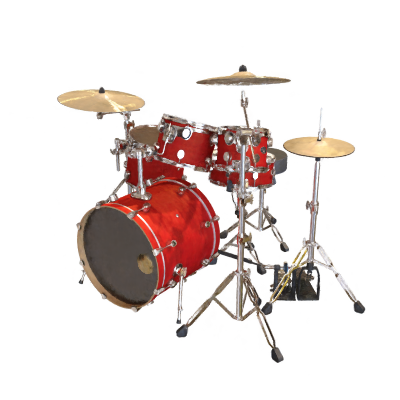} &
\hspace{-0.4cm}
\includegraphics[width=0.22\linewidth]{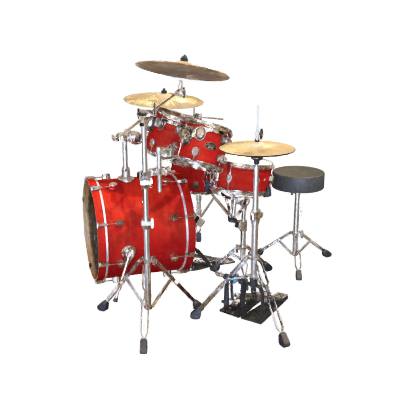} &
\hspace{-0.4cm}
\includegraphics[width=0.22\linewidth]{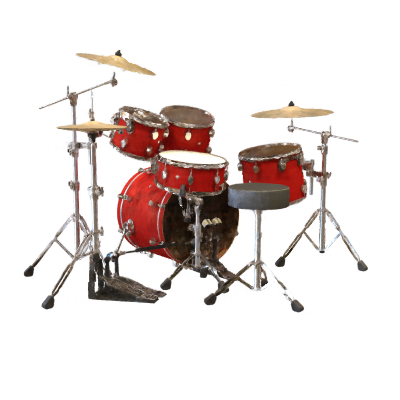}
\hspace{-0.4cm}
\\

\rotatebox{90}{\quad\quad\quad CeRF} &
\includegraphics[width=0.22\linewidth]{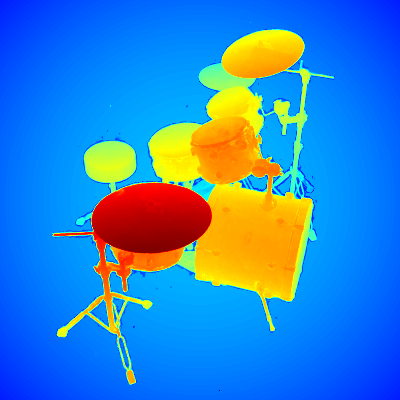} &
\hspace{-0.4cm}
\includegraphics[width=0.22\linewidth]{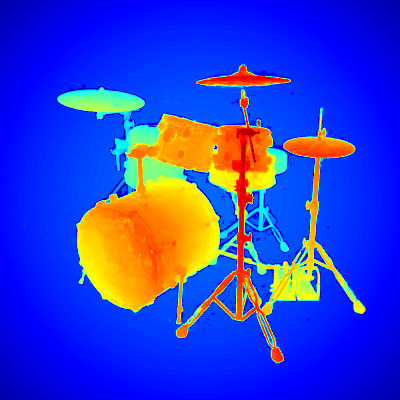} &
\hspace{-0.4cm}
\includegraphics[width=0.22\linewidth]{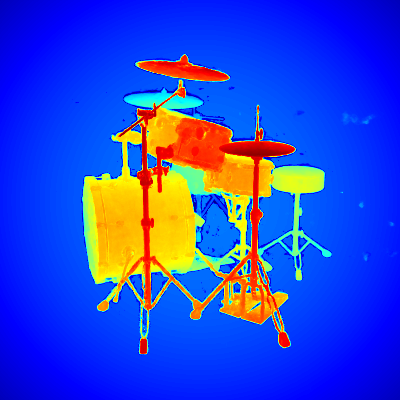} &
\hspace{-0.4cm}
\includegraphics[width=0.22\linewidth]{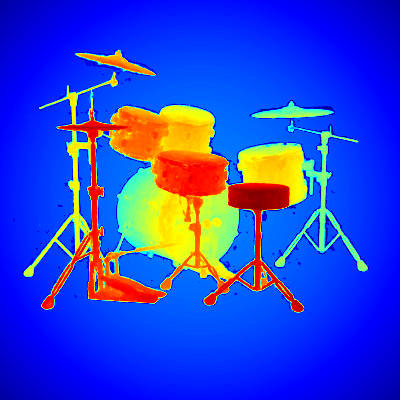}
\hspace{-0.4cm}
\\

\rotatebox{90}{\quad\quad CeRF w/o $\mathcal{L}_e$} &
\includegraphics[width=0.22\linewidth]{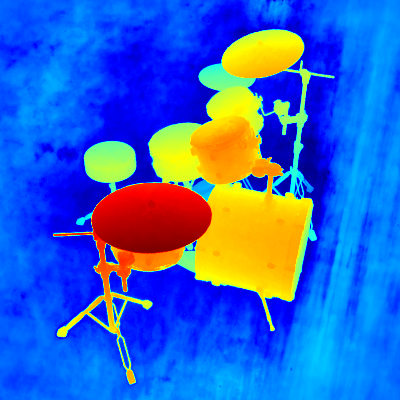} &
\hspace{-0.4cm}
\includegraphics[width=0.22\linewidth]{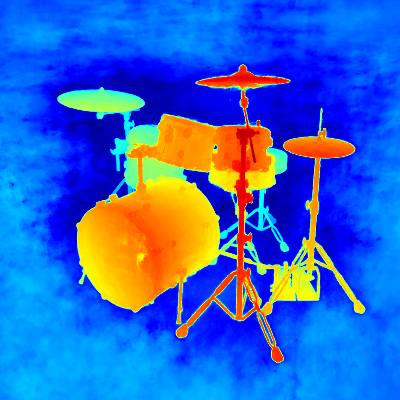} &
\hspace{-0.4cm}
\includegraphics[width=0.22\linewidth]{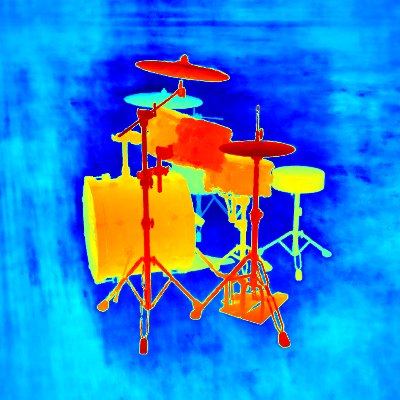} &
\hspace{-0.4cm}
\includegraphics[width=0.22\linewidth]{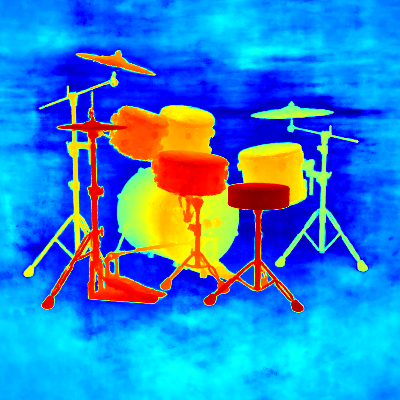} 
\hspace{-0.4cm}
\\
\\
\end{tabular}
\caption{Ablation of Empty Space Regularization. The first line is rendered images. The second line is depth maps generated by full version of CeRF and the third line is depth maps generated by CeRF without Empty Space Regularization.
}
\label{fig:empty_loss_ablation}
\end{figure}

As depicted in Figure \ref{fig:empty_loss_ablation}, the bluer area indicates a larger depth value and vice versa. We set the epipolar point at a depth of 120, where the maximum depth of objects in the scene is 6. It is evident that CeRF's results are displayed as white in empty spaces whereas missing Loss with Empty Space Regularization results appear in varying shades of gray at their respective positions. This demonstrates that rays directed towards empty spaces select points located farthest away - i.e., the epipolar point. The epipolar mechanism and Loss with Empty Space Regularization functioned as expected.

\section{CeRF with learnable $s_e$}

\begin{figure*}[!t]
  \centering
     \includegraphics[clip, trim = 2cm 4.cm 2.5cm 2cm, width=0.98\linewidth]{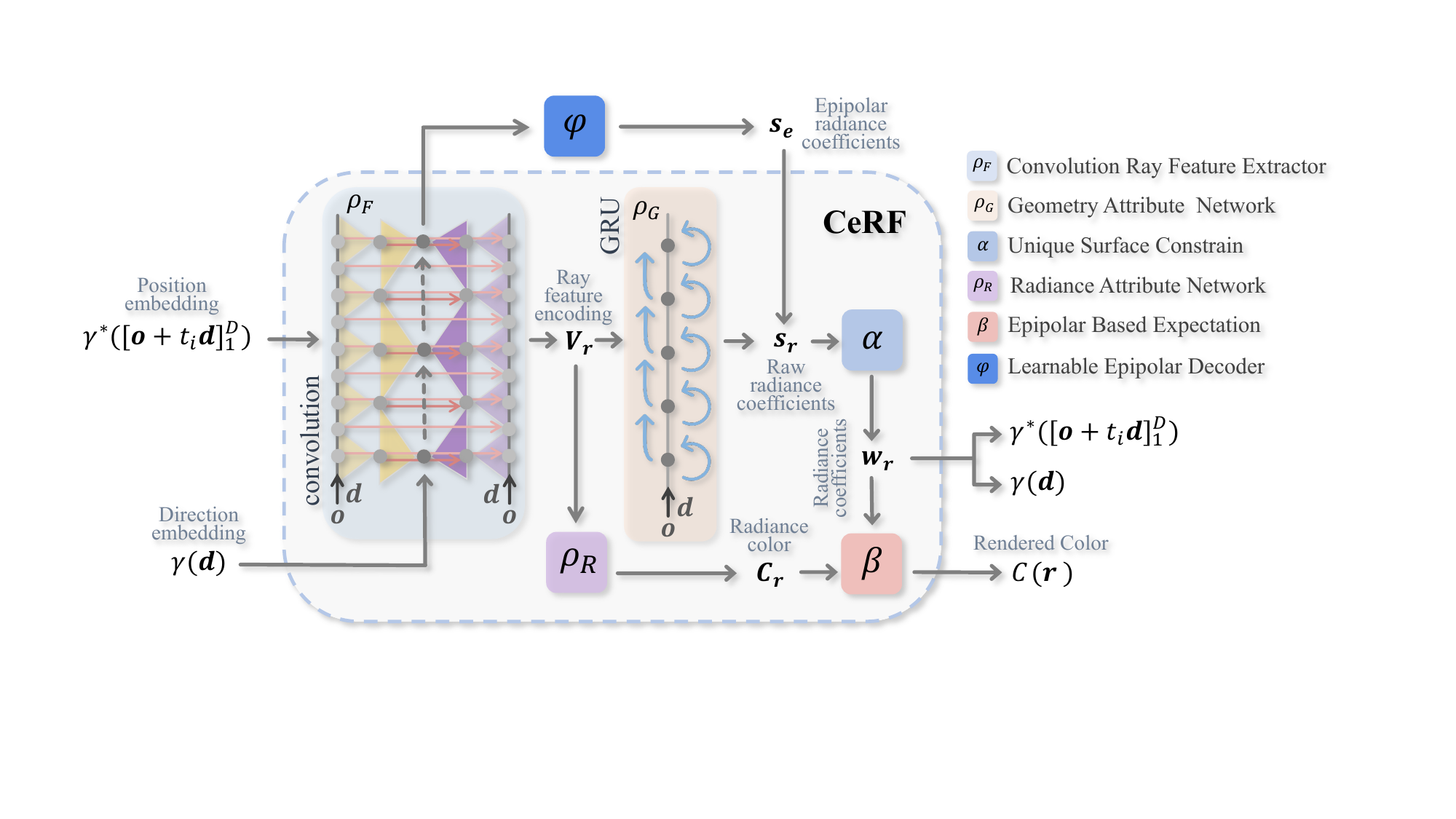}
     \caption{Illustration of CeRF with learnable $s_e$}
   \label{fig:overview-learn-se}
\end{figure*}

For learnable $s_e$, as shown in the Figure~\ref{fig:overview-learn-se}, we try a new mechanism to compute the corresponding $s_e$ value for each ray.  Specifically, we obtain the bottleneck layer features obtained in Convolutional Ray Feature Extractor for each ray and input that feature into a Learnable Epipolar Decoder $varphi$. The structure of Learnable Epipolar Decoder consists of two MLP layers with feature channels of 256 and 32 respectively, along with an additional sigmoid. 

Our main text explains that our U-shaped CNN architecture treats each ray's space as a complete scene representation for feature extraction. Therefore, we believe that these compact bottleneck layer features contain simple information such as whether the ray passes through an object. However, through the experiment, we discover that it is challenging to extract information about a single epipolar point from the bottleneck layer features. This is because these features encode complex spatial information for an entire ray. Despite compressing feature channels to $1/4$ of the original ray features (which has N 256-dimensional features), the bottleneck layer features still have high dimensionality and are difficult to reduce using a small MLP. As such, we opt for a fixed-value $s_e$ instead of the learnable $s_e$ mechanism. Our experiment shows that this approach resulted in significant performance improvements with minimal additional computational overhead.

\section{Geometry information}
\begin{figure}[!t]
\centering
\begin{tabular}{cc}

\includegraphics[width=0.4\linewidth]{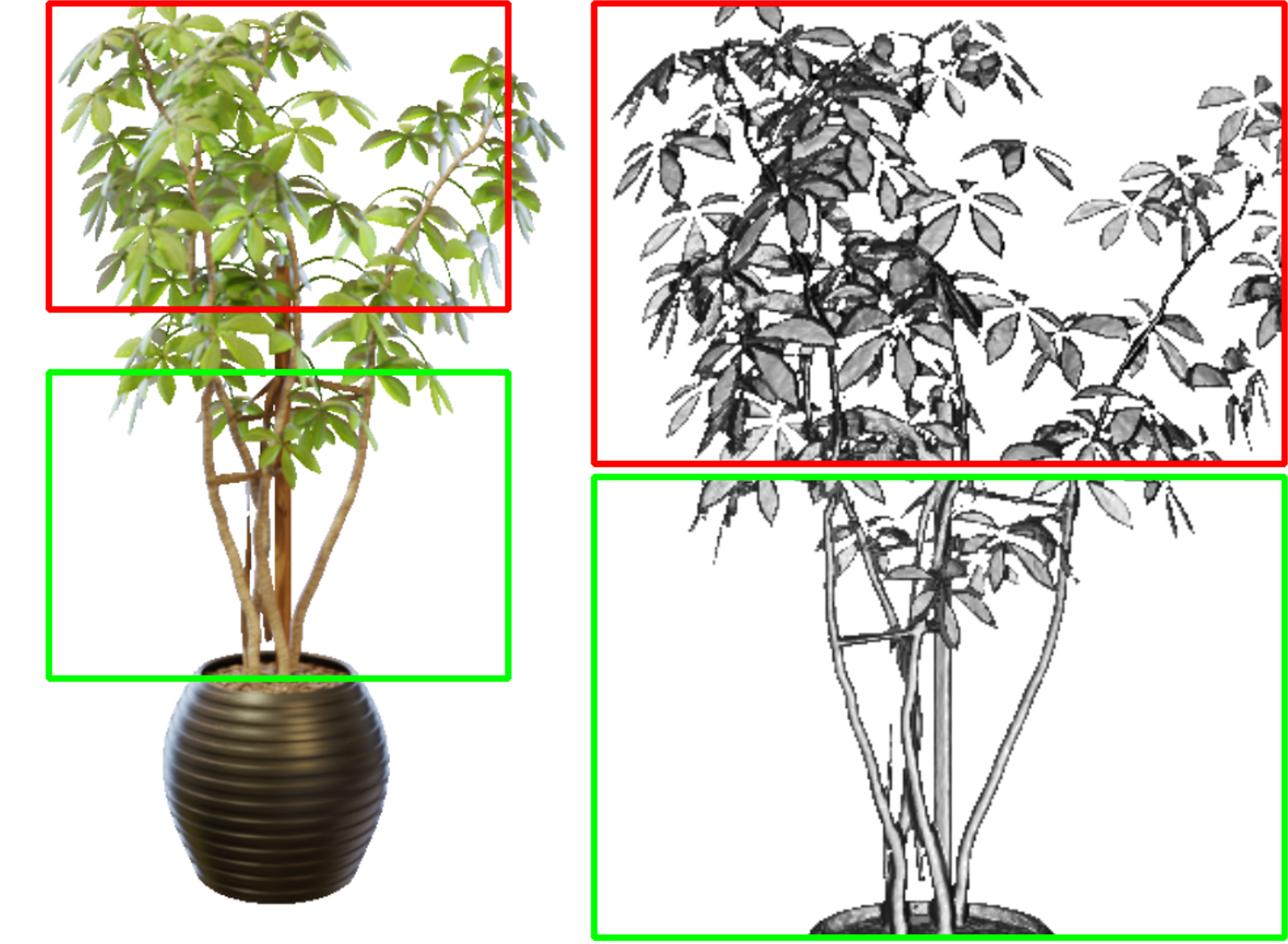} &
\hspace{0.67cm}
\includegraphics[width=0.5\linewidth]{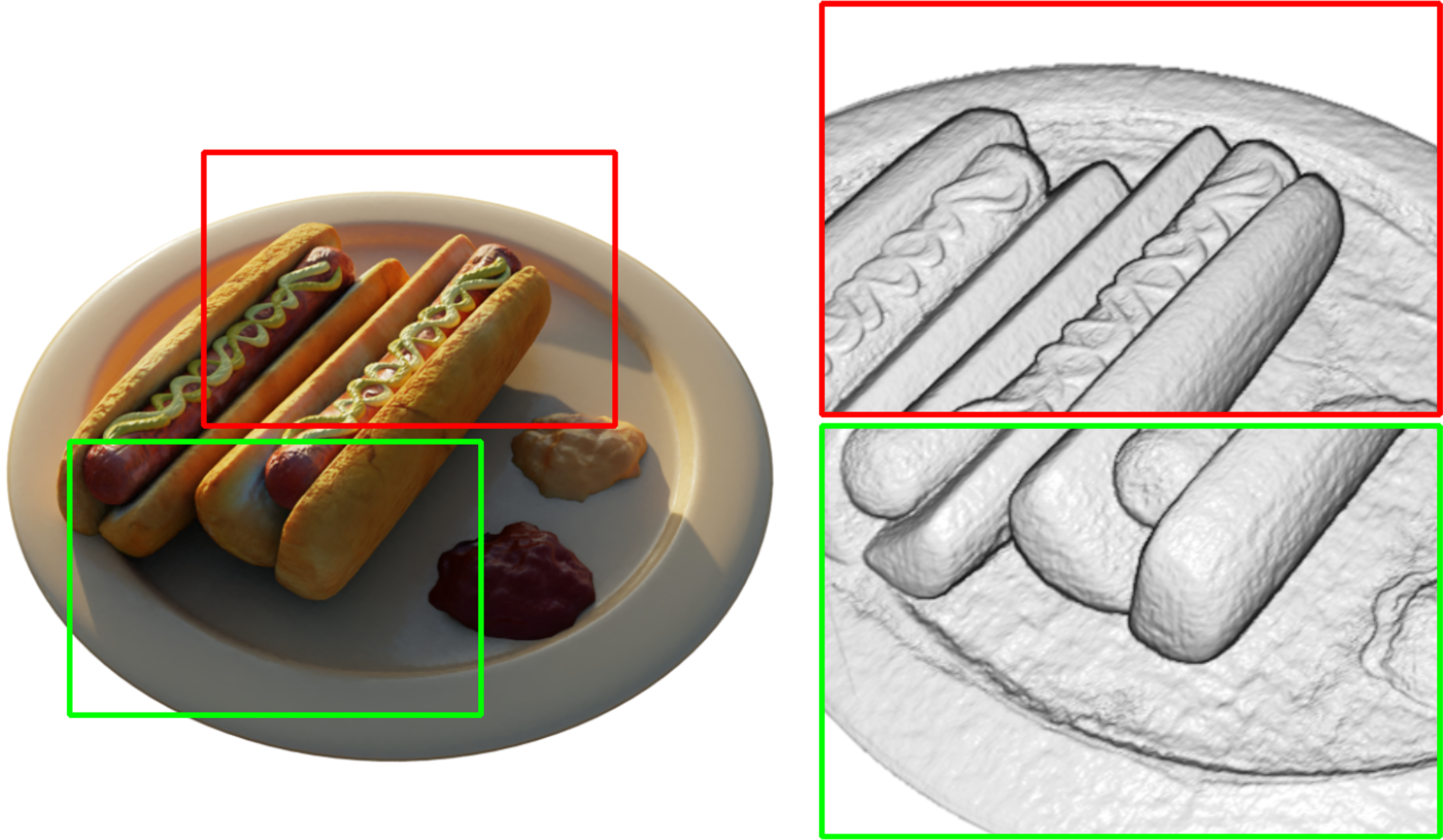}
\vspace{0.5cm}
\\

\hspace{-0.85cm}
\includegraphics[width=0.455\linewidth]{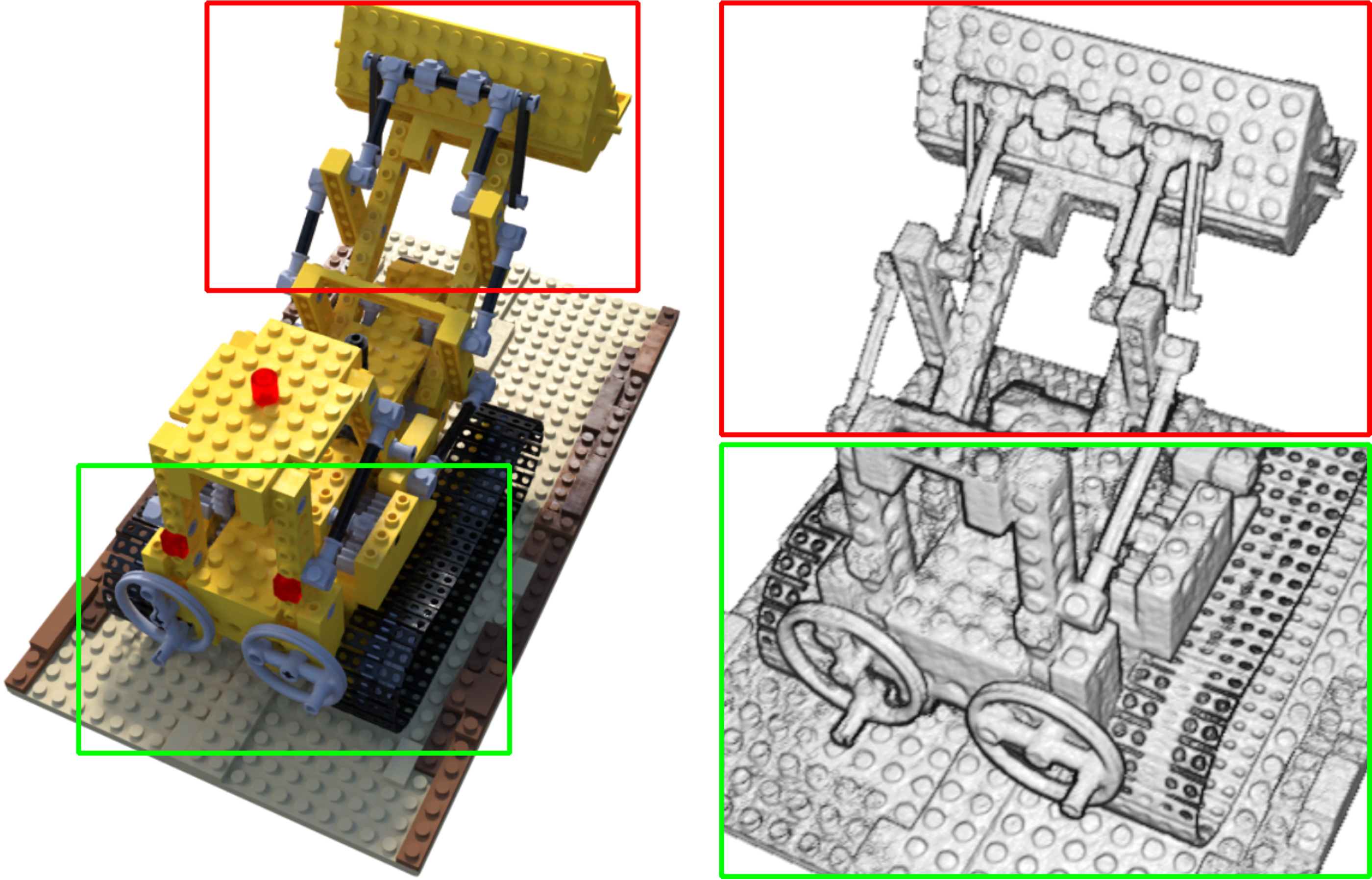} &
\hspace{-0.2cm}
\includegraphics[width=0.56\linewidth]{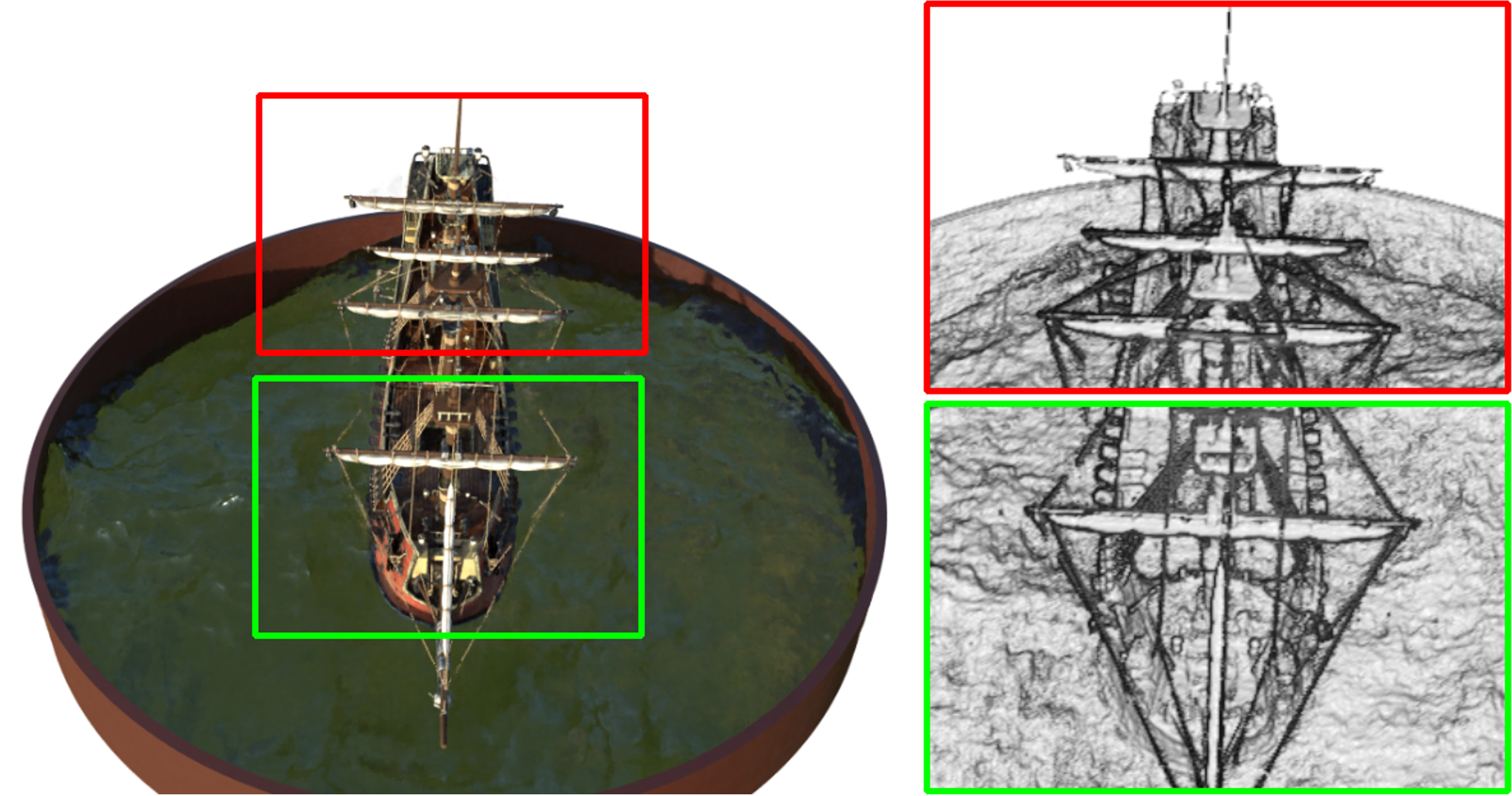}
\\
\end{tabular}
\caption{Rendered images and extracted meshes from CeRF. Unlike light field, our method fuses the features along the ray direction but still retains the information of each individual point, so that the mesh can be easily obtained, which retains the advantage of radiance field.
}
\label{fig:mesh_vis}
\end{figure}

\begin{figure*}[t]
  \centering
  \begin{tabular}{cccccc}
    \hspace{-0.8cm}
    \includegraphics[width=0.165\linewidth]{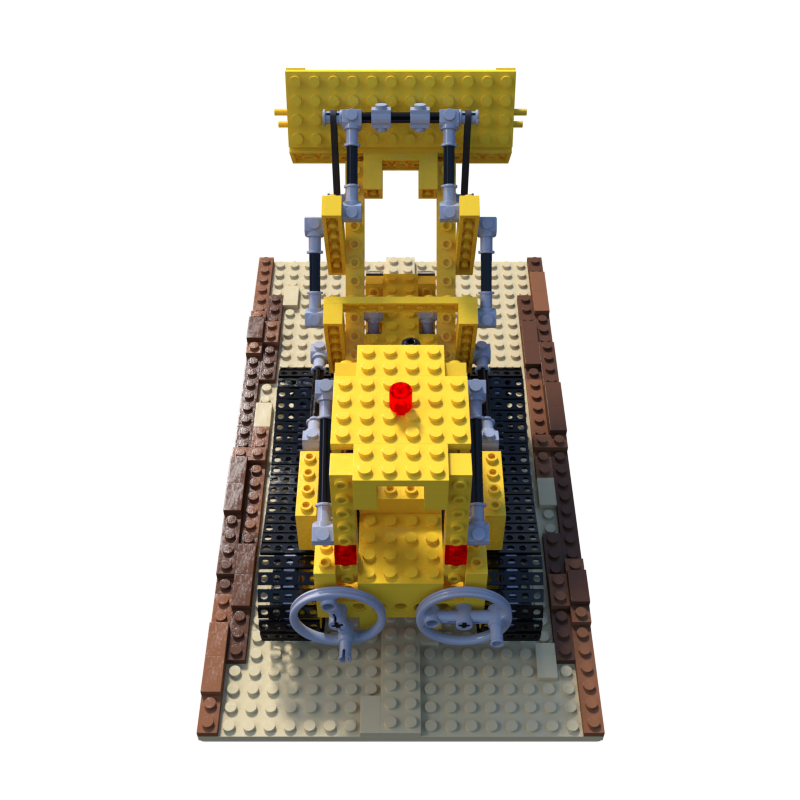} \hspace{-0.6cm} &
    \includegraphics[width=0.165\linewidth]{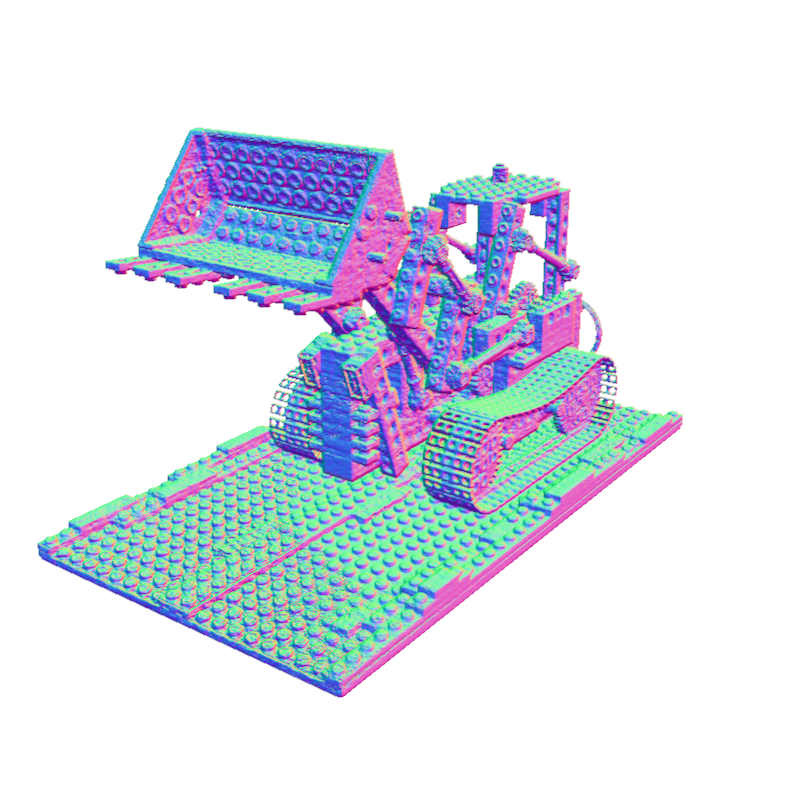} \hspace{-0.6cm} &
    \includegraphics[width=0.165\linewidth]{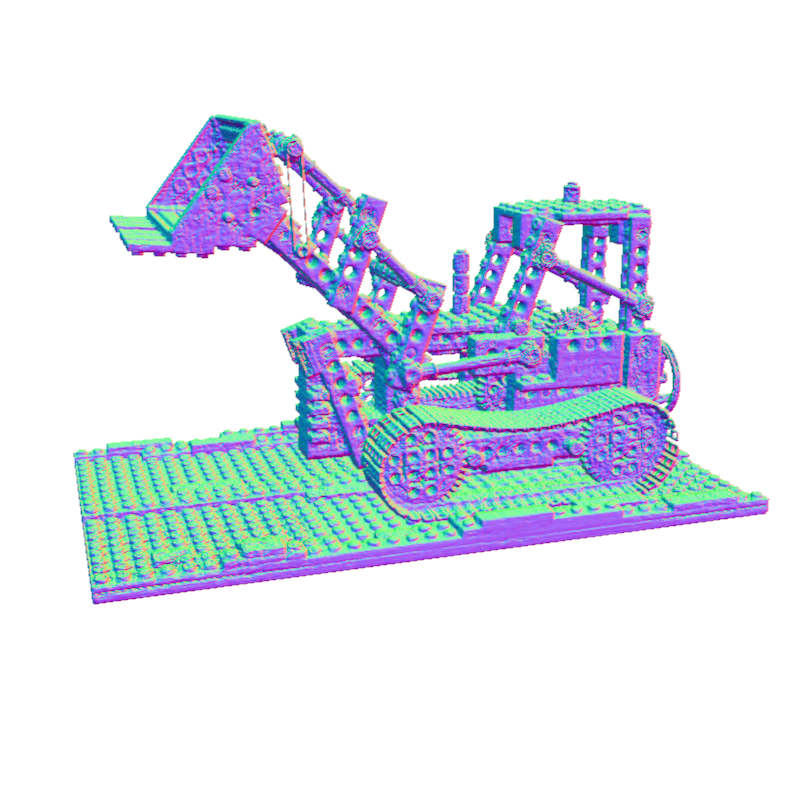} \hspace{-0.6cm} &
    \includegraphics[width=0.165\linewidth]{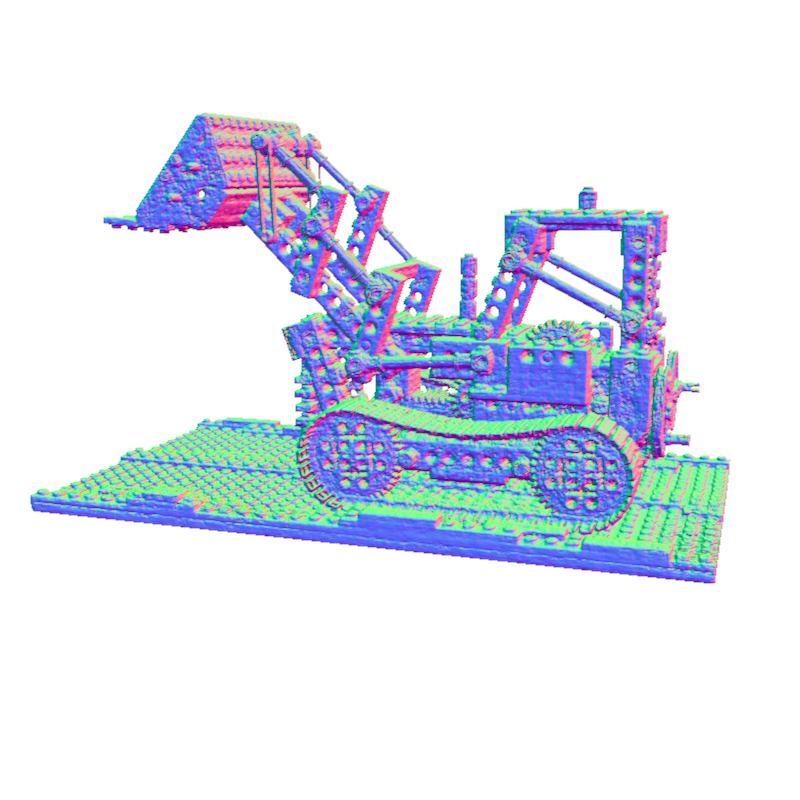} \hspace{-0.6cm} &
    \includegraphics[width=0.165\linewidth]{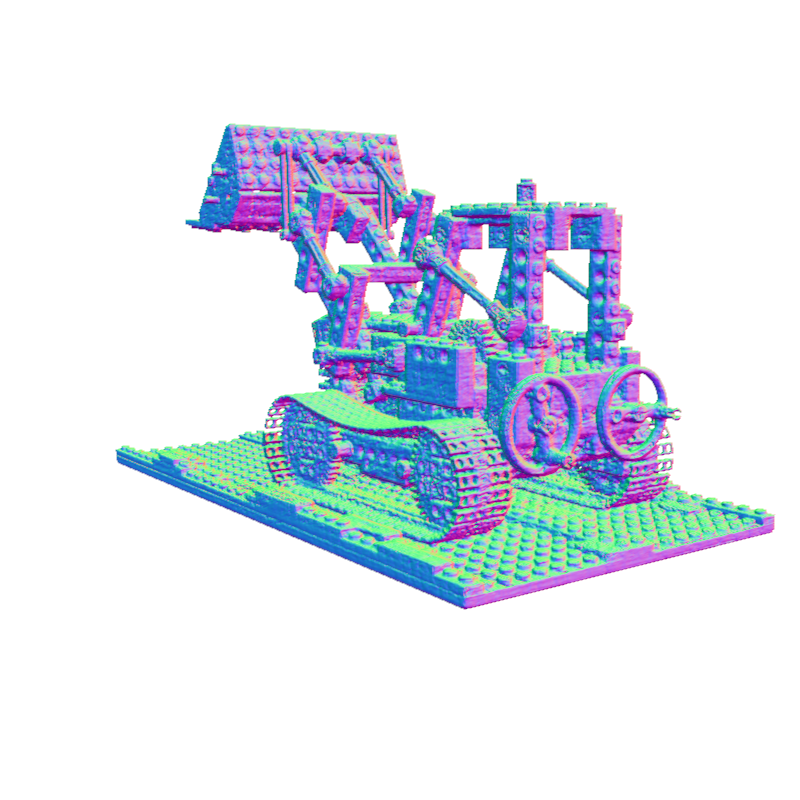} \hspace{-0.6cm} &
    \includegraphics[width=0.165\linewidth]{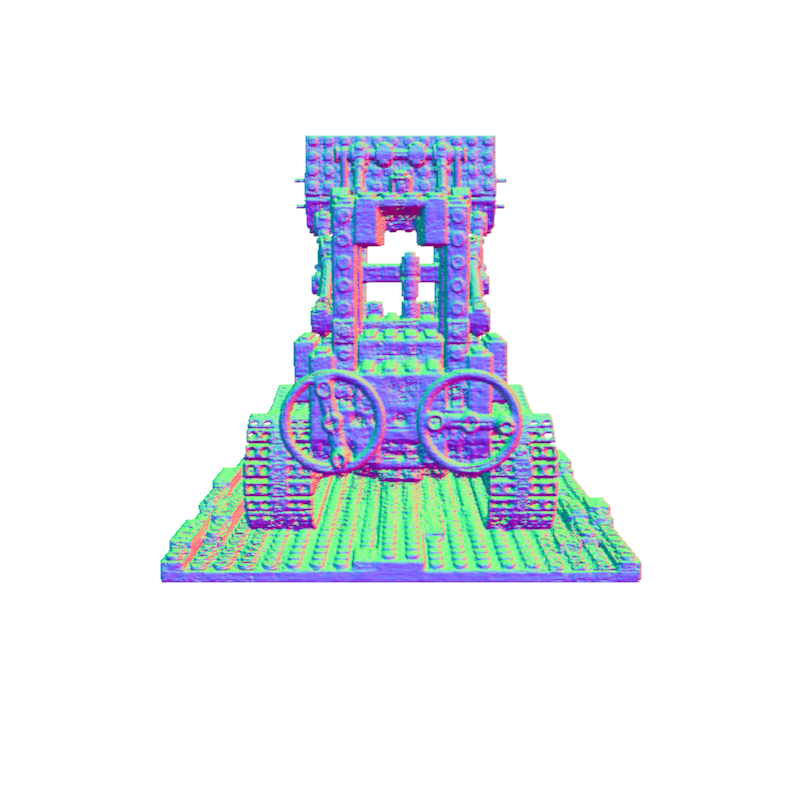} \hspace{-0.6cm} 

    \\ 
    \hspace{-0.8cm}
    \includegraphics[width=0.165\linewidth]{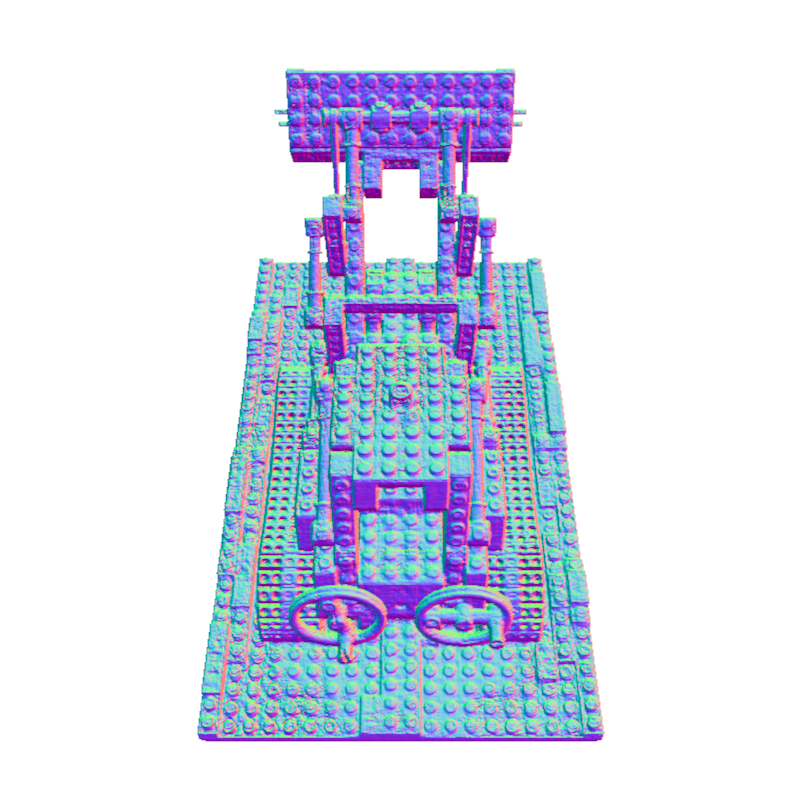} \hspace{-0.6cm} &
    \includegraphics[width=0.165\linewidth]{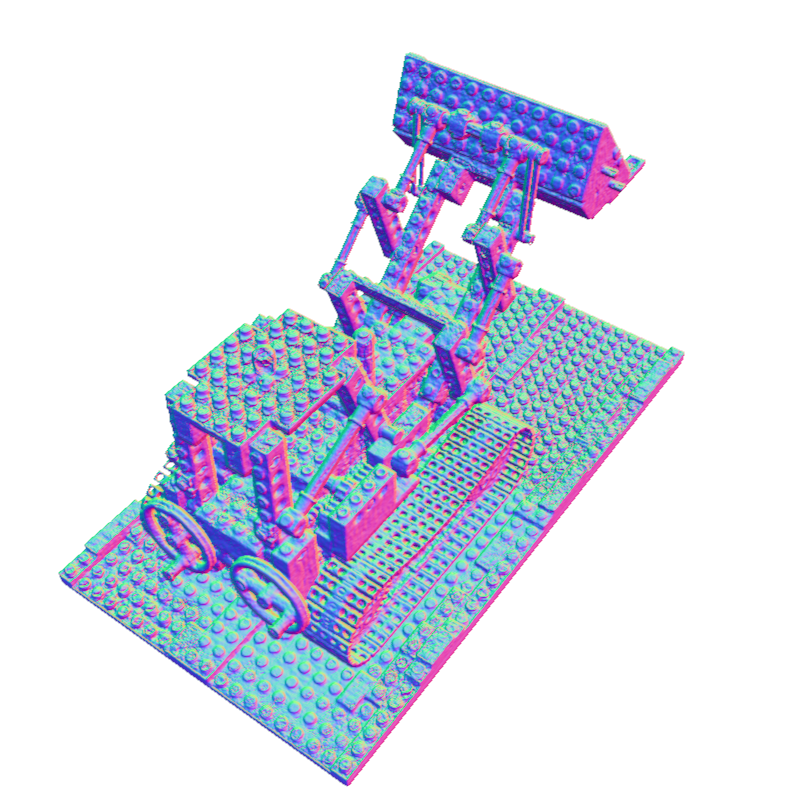} \hspace{-0.6cm} &
    \includegraphics[width=0.165\linewidth]{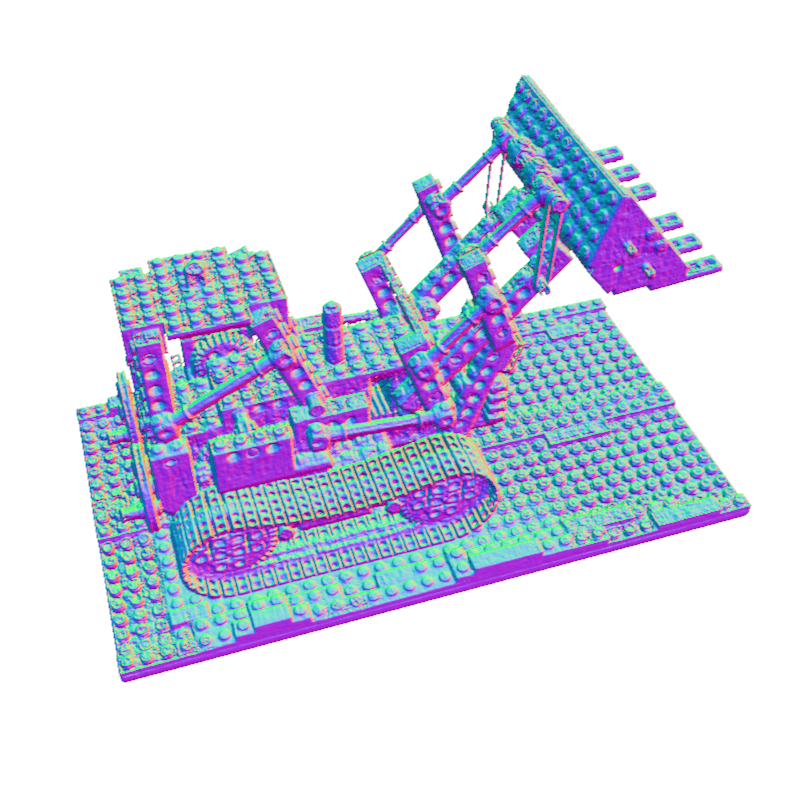} \hspace{-0.6cm} &
    \includegraphics[width=0.165\linewidth]{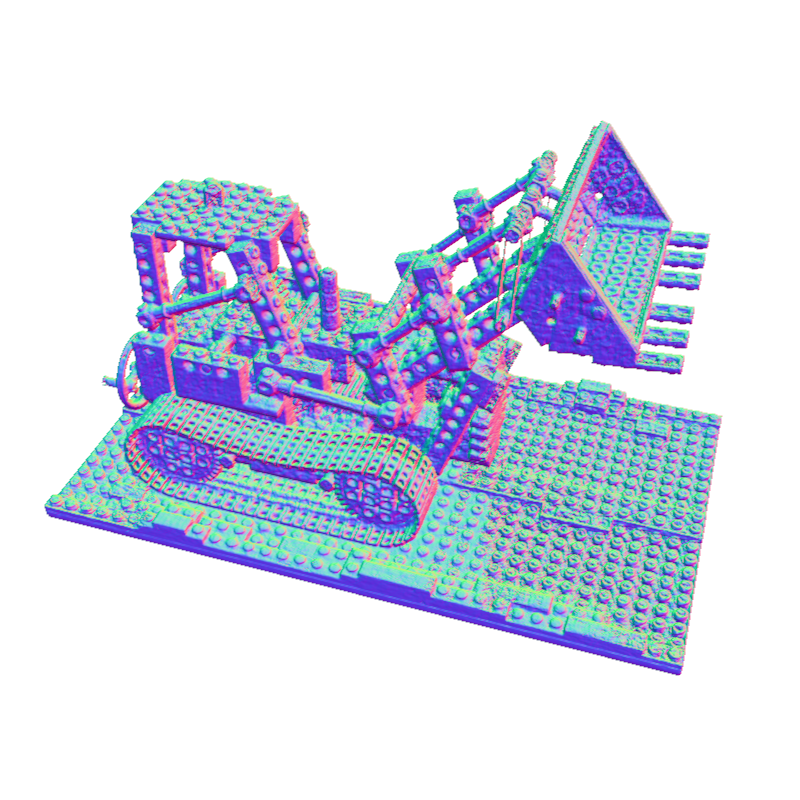} \hspace{-0.6cm} &
    \includegraphics[width=0.165\linewidth]{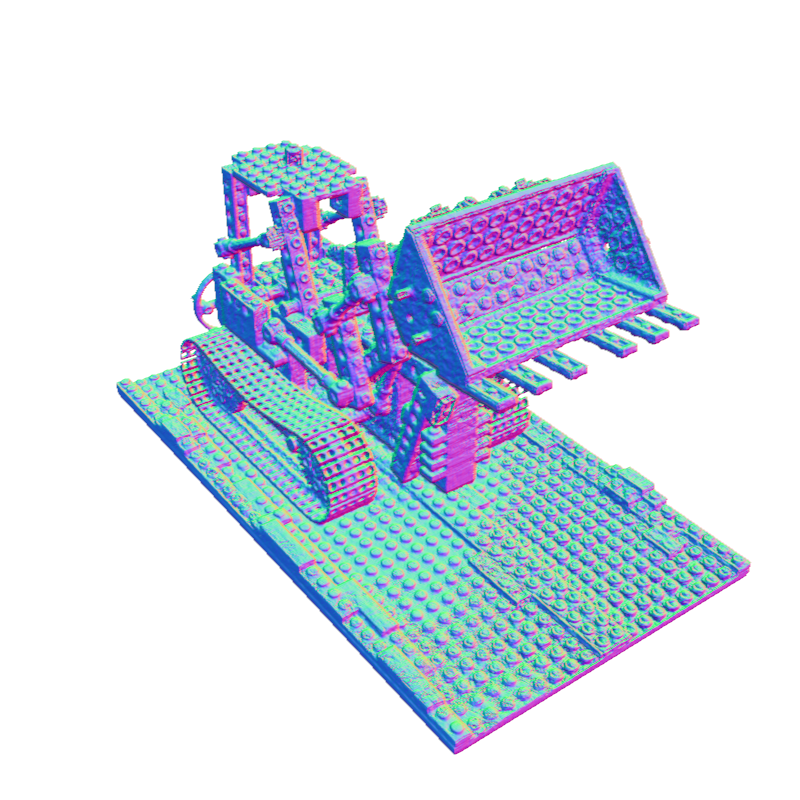} \hspace{-0.6cm} &
    \includegraphics[width=0.165\linewidth]{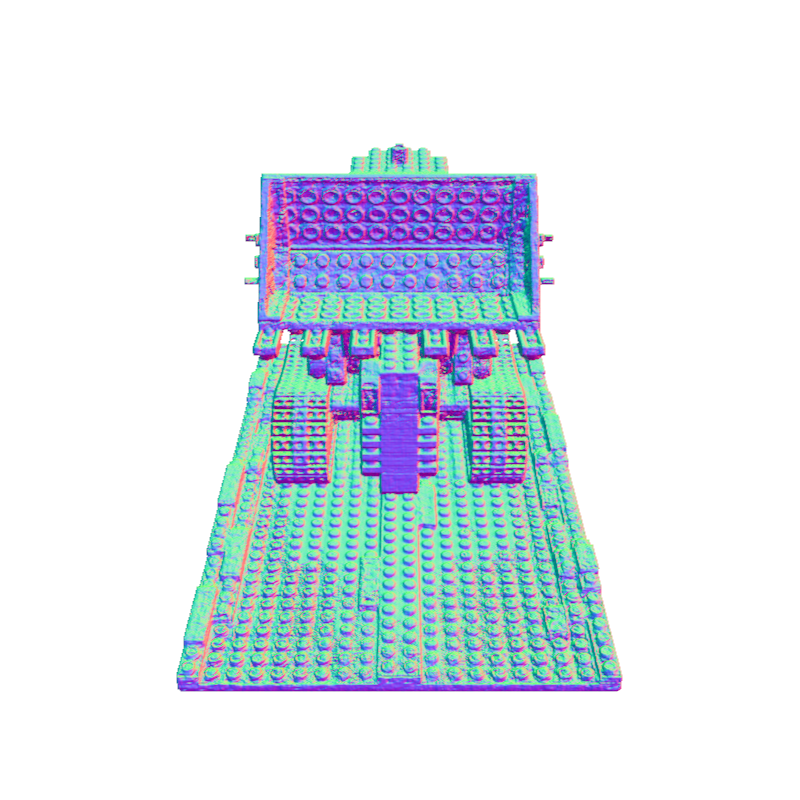} \hspace{-0.6cm} 
    \\ 
    \hspace{-0.8cm}
    \includegraphics[width=0.165\linewidth]{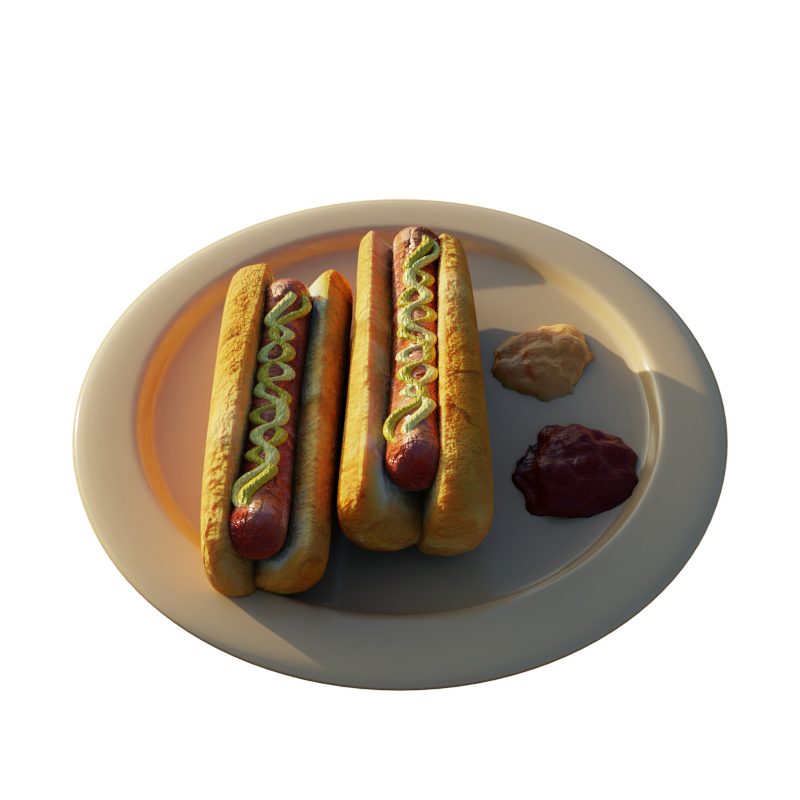} \hspace{-0.6cm} &
    \includegraphics[width=0.165\linewidth]{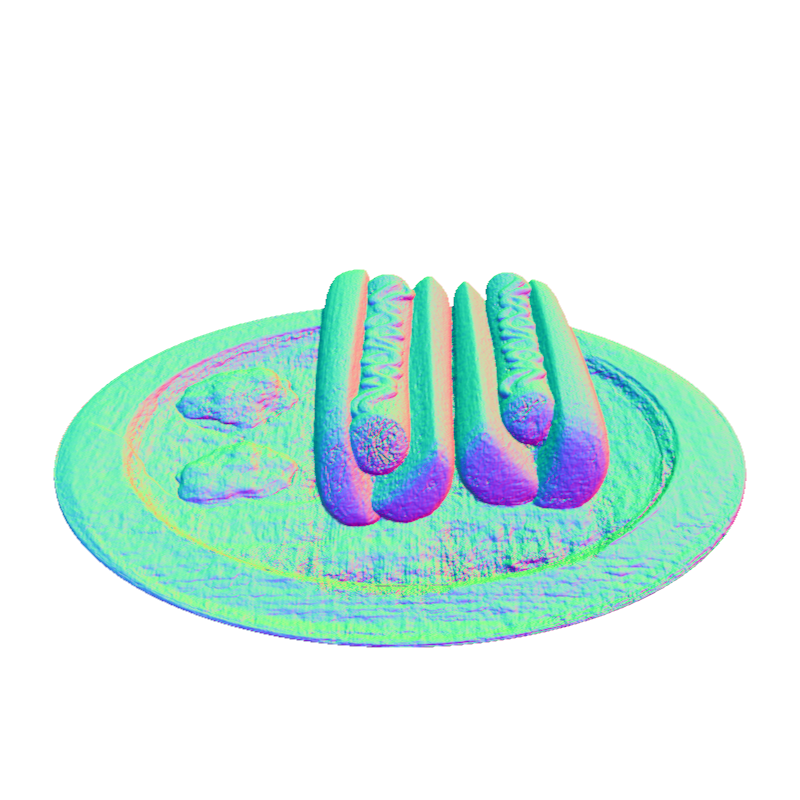} \hspace{-0.6cm} &
    \includegraphics[width=0.165\linewidth]{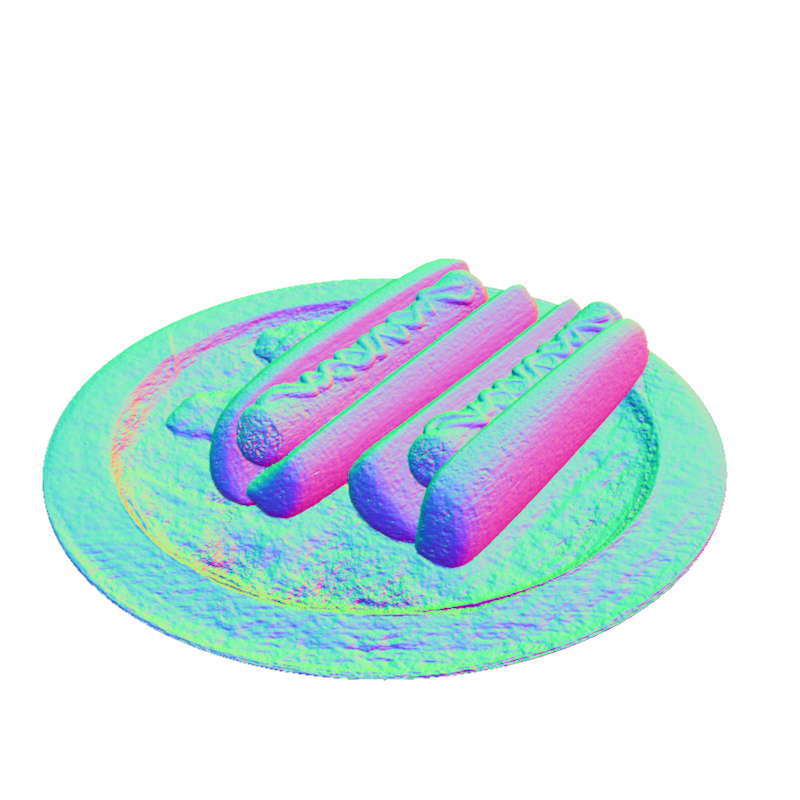} \hspace{-0.6cm} &
    \includegraphics[width=0.165\linewidth]{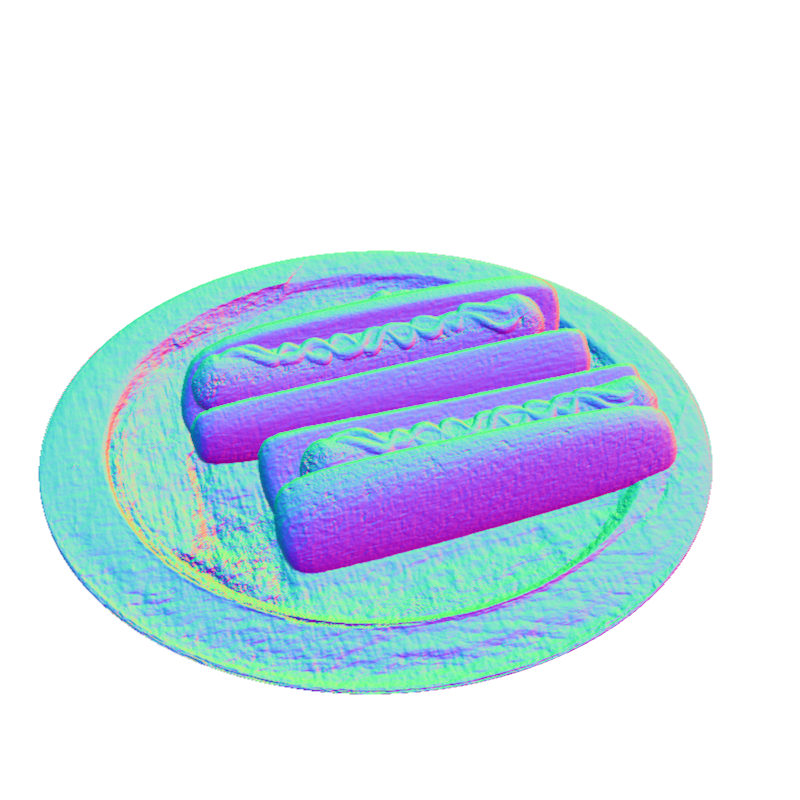} \hspace{-0.6cm} &
    \includegraphics[width=0.165\linewidth]{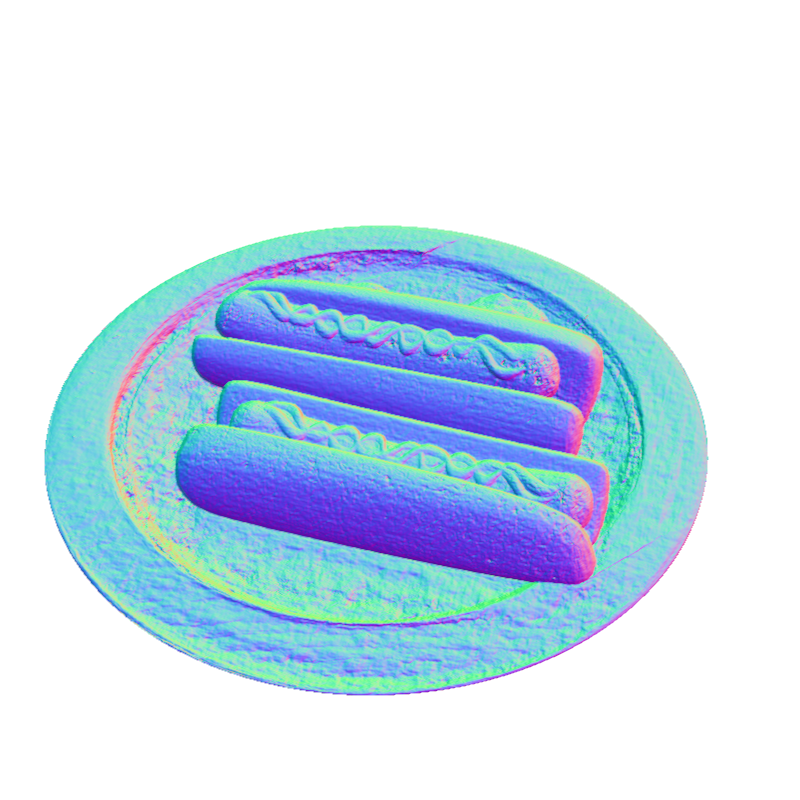} \hspace{-0.6cm} &
    \includegraphics[width=0.165\linewidth]{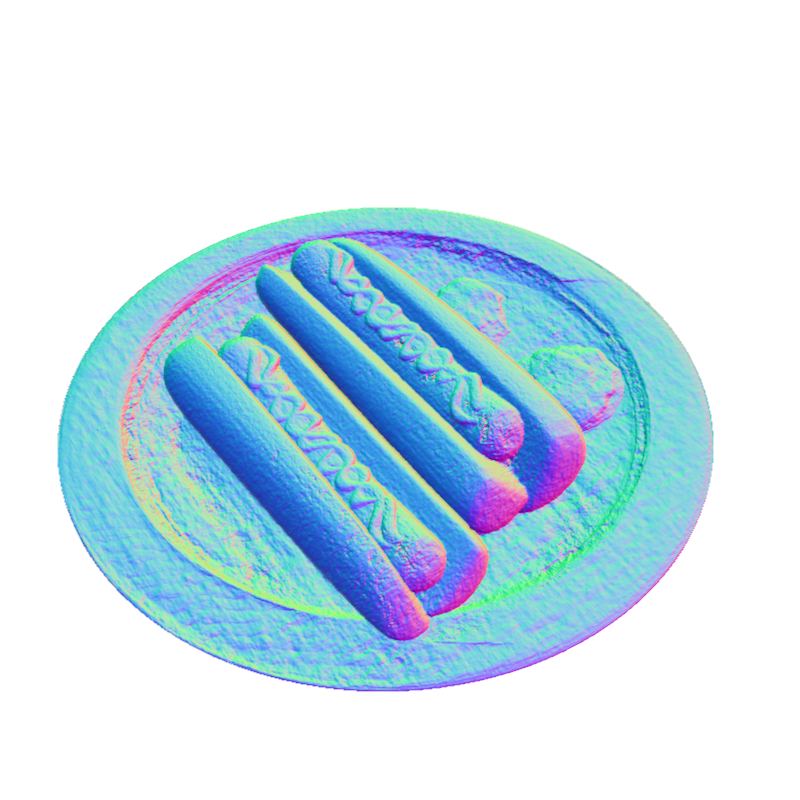} \hspace{-0.6cm} 
    \\ 
    \hspace{-0.8cm}
    \includegraphics[width=0.165\linewidth]{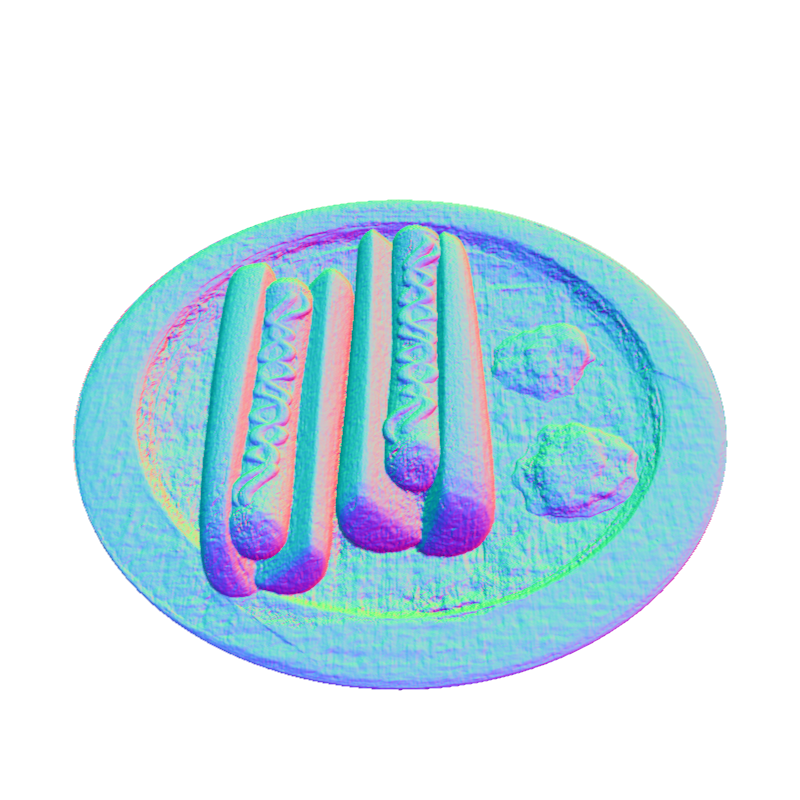} \hspace{-0.6cm} &
    \includegraphics[width=0.165\linewidth]{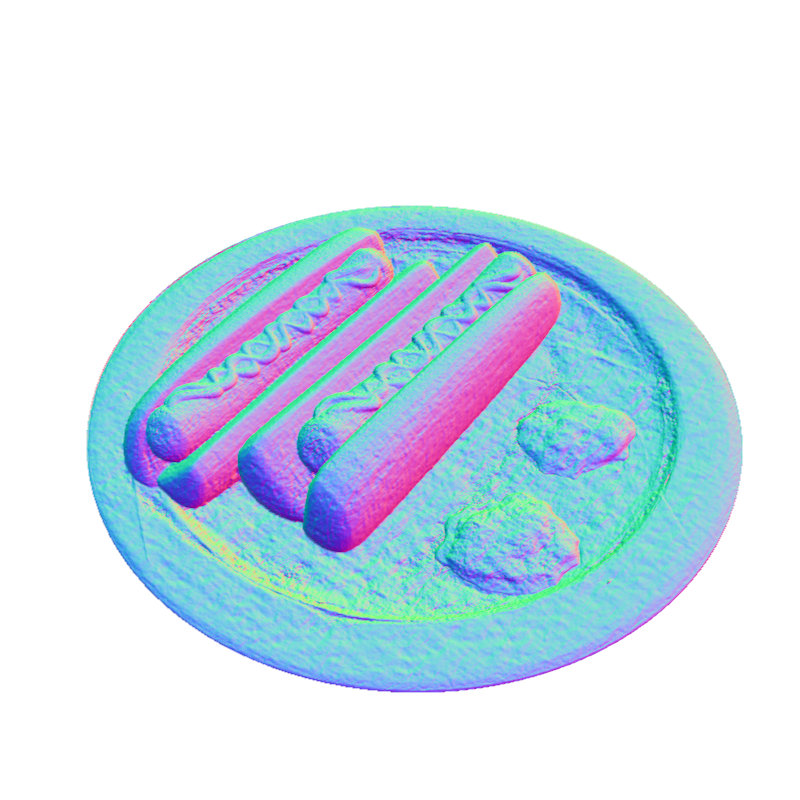} \hspace{-0.6cm} &
    \includegraphics[width=0.165\linewidth]{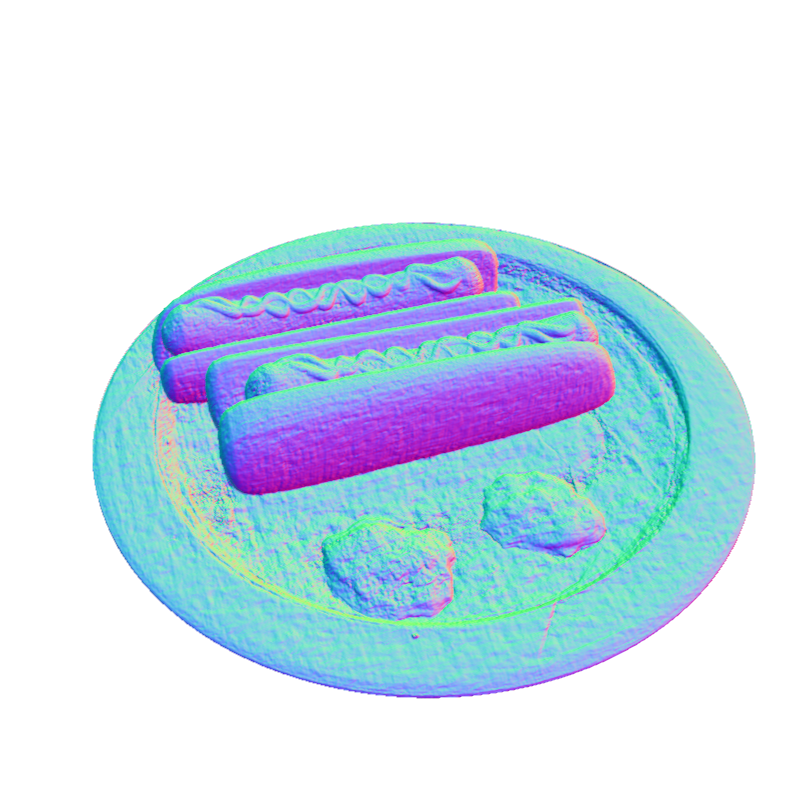} \hspace{-0.6cm} &
    \includegraphics[width=0.165\linewidth]{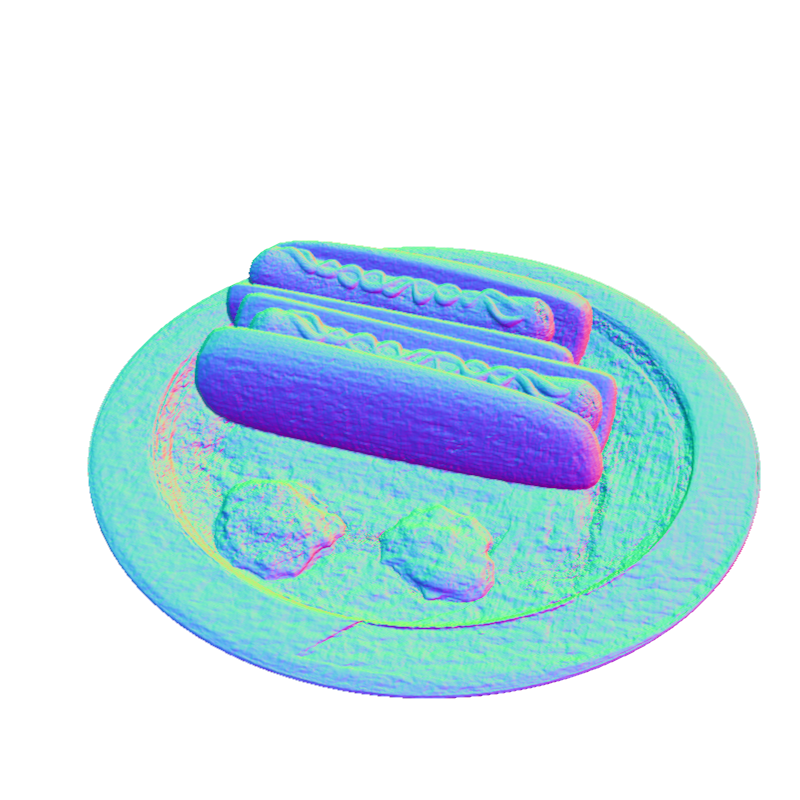} \hspace{-0.6cm} &
    \includegraphics[width=0.165\linewidth]{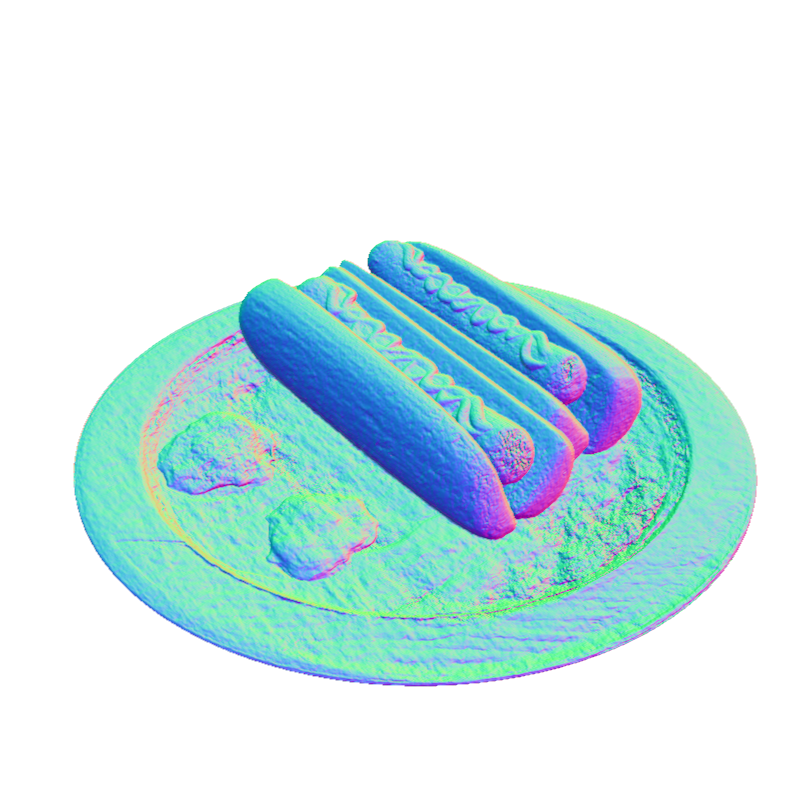} \hspace{-0.6cm} &
    \includegraphics[width=0.165\linewidth]{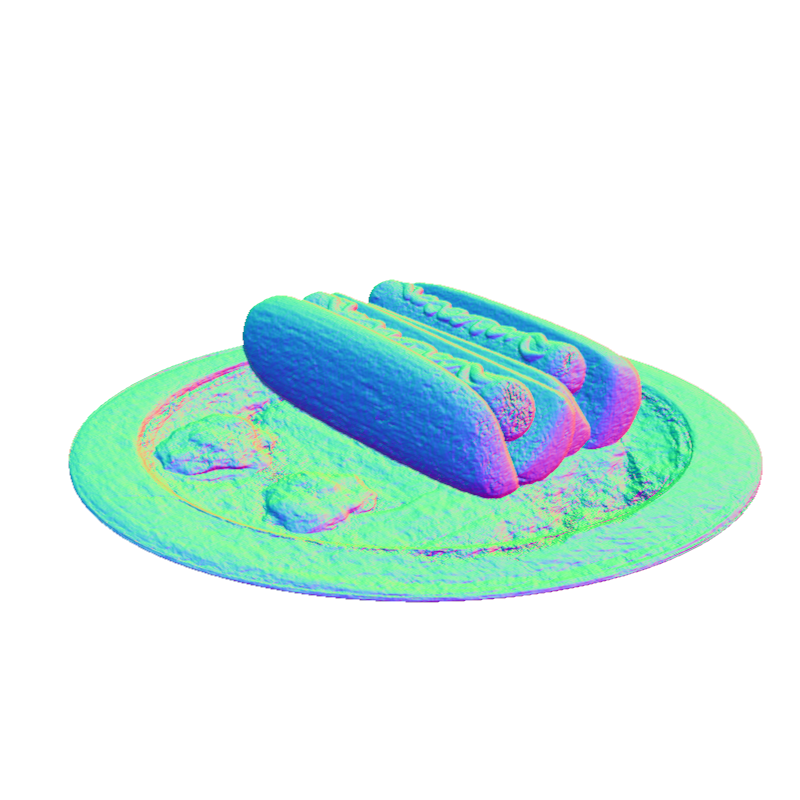} \hspace{-0.6cm} 
    \\ 
    \hspace{-0.8cm}
    \includegraphics[width=0.165\linewidth]{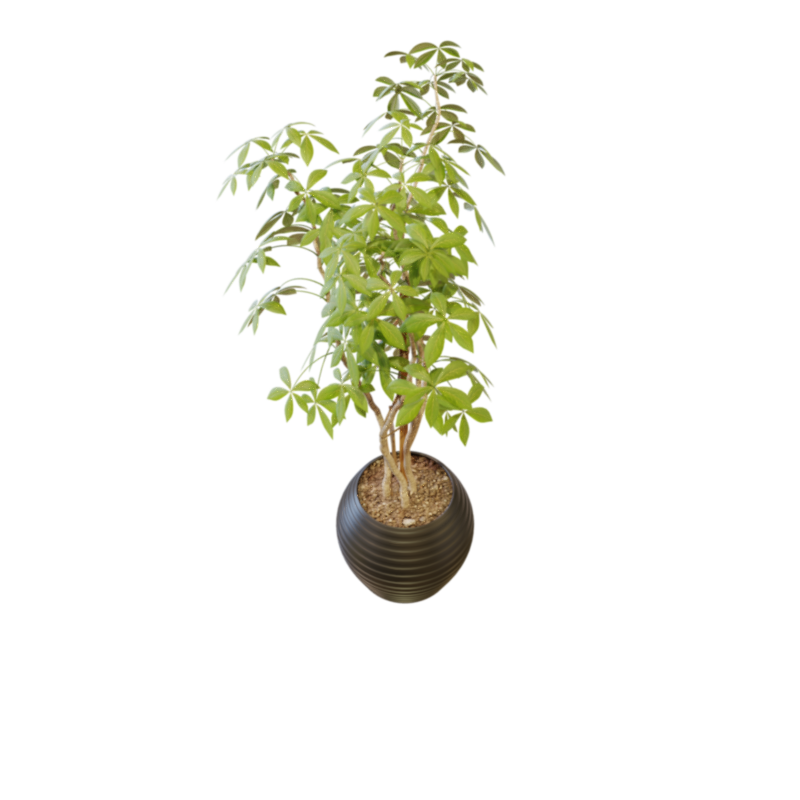} \hspace{-0.6cm} &
    \includegraphics[width=0.165\linewidth]{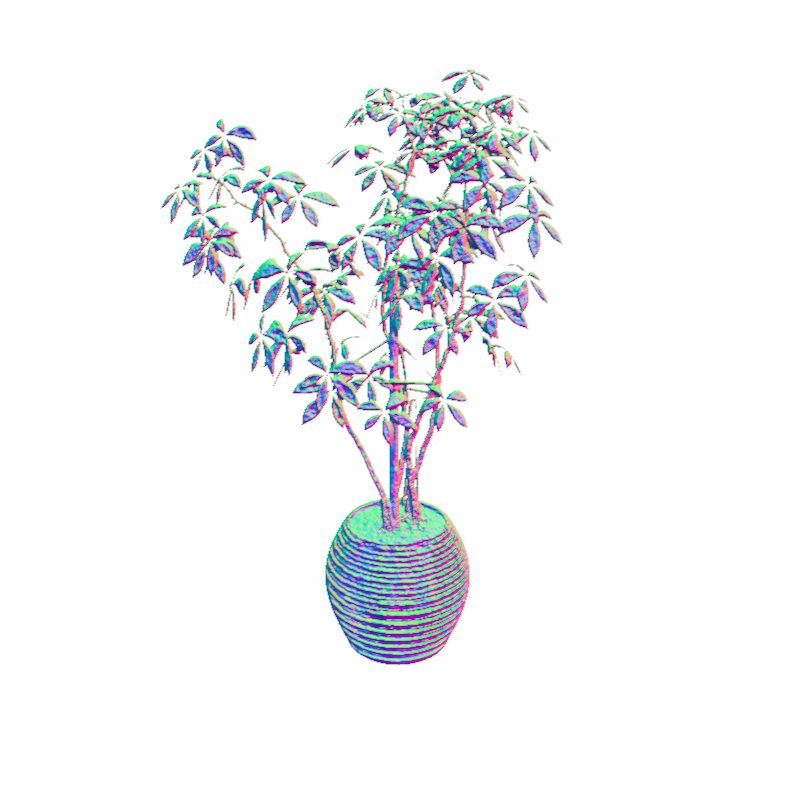} \hspace{-0.6cm} &
    \includegraphics[width=0.165\linewidth]{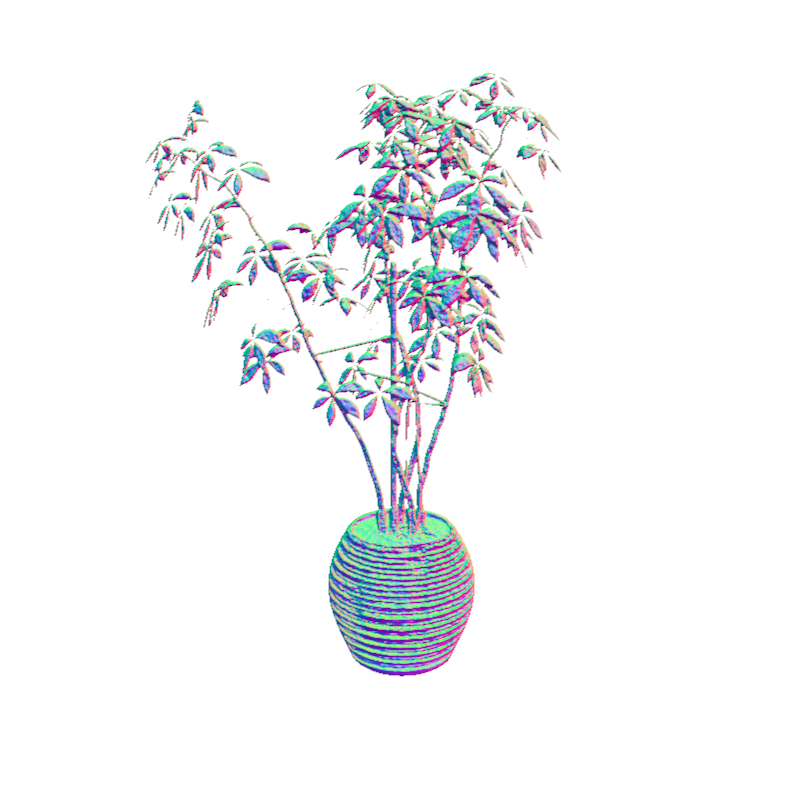} \hspace{-0.6cm} &
    \includegraphics[width=0.165\linewidth]{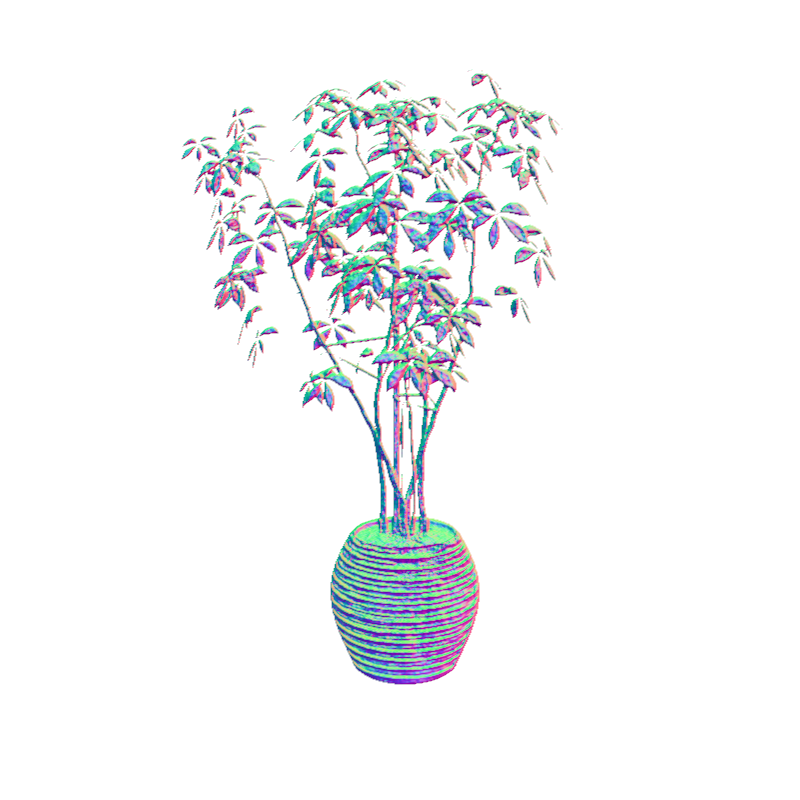} \hspace{-0.6cm} &
    \includegraphics[width=0.165\linewidth]{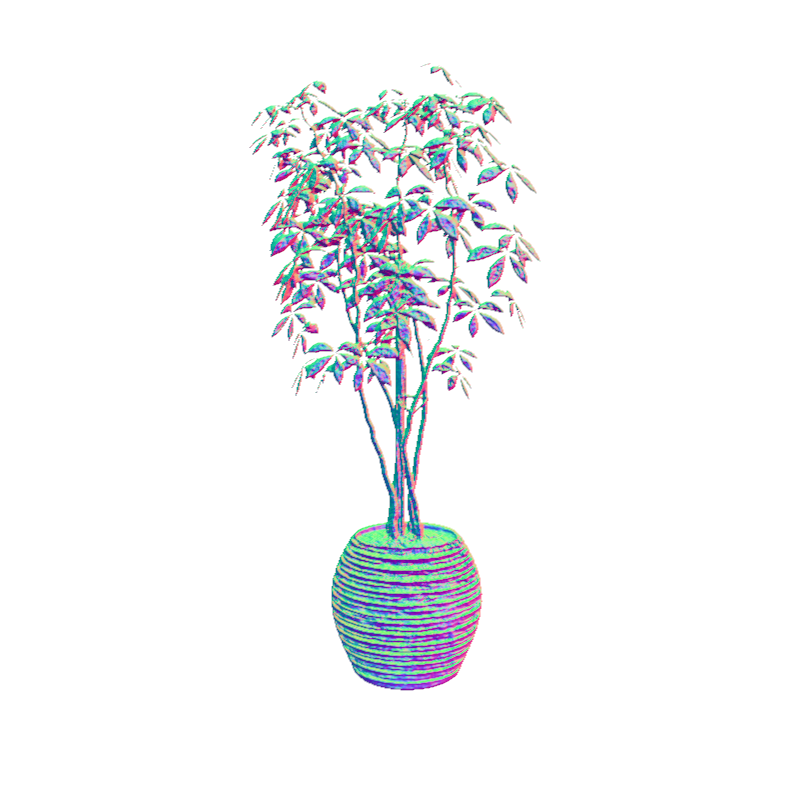} \hspace{-0.6cm} &
    \includegraphics[width=0.165\linewidth]{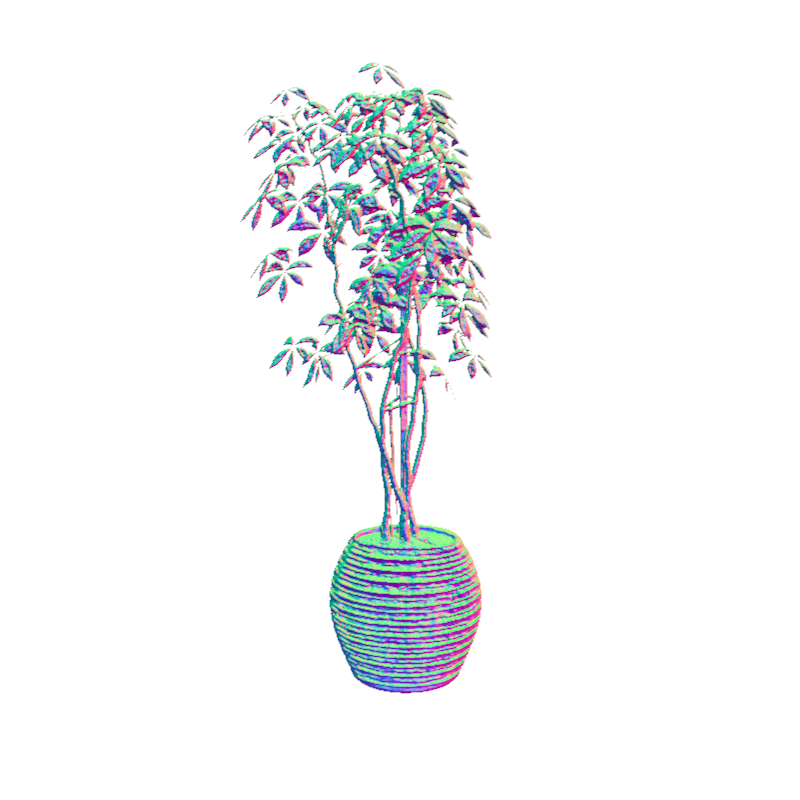} \hspace{-0.6cm} 
    \\ 
    \hspace{-0.8cm}
    \includegraphics[width=0.165\linewidth]{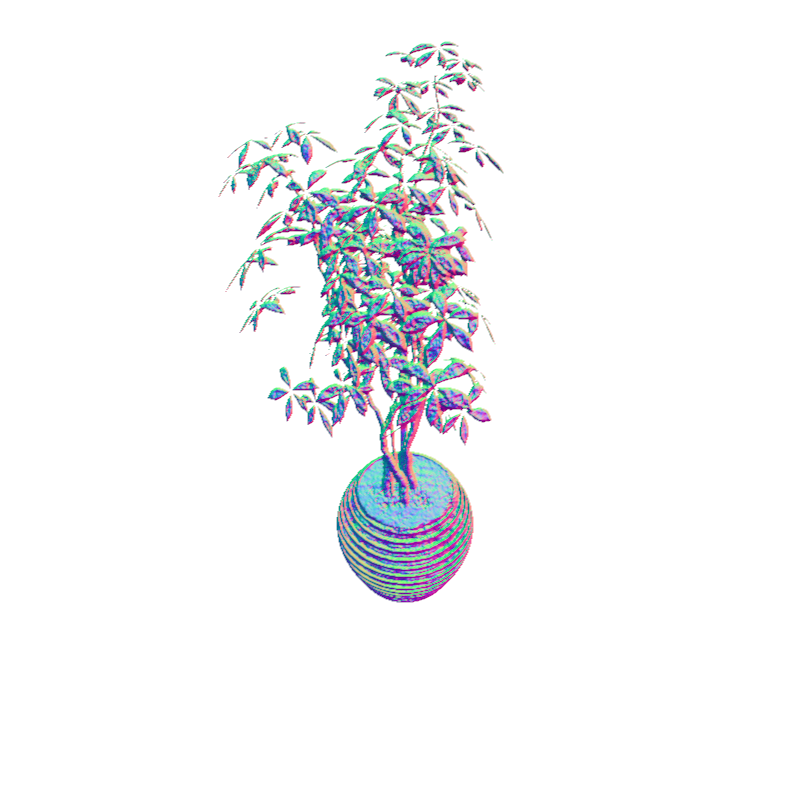} \hspace{-0.6cm} &
    \includegraphics[width=0.165\linewidth]{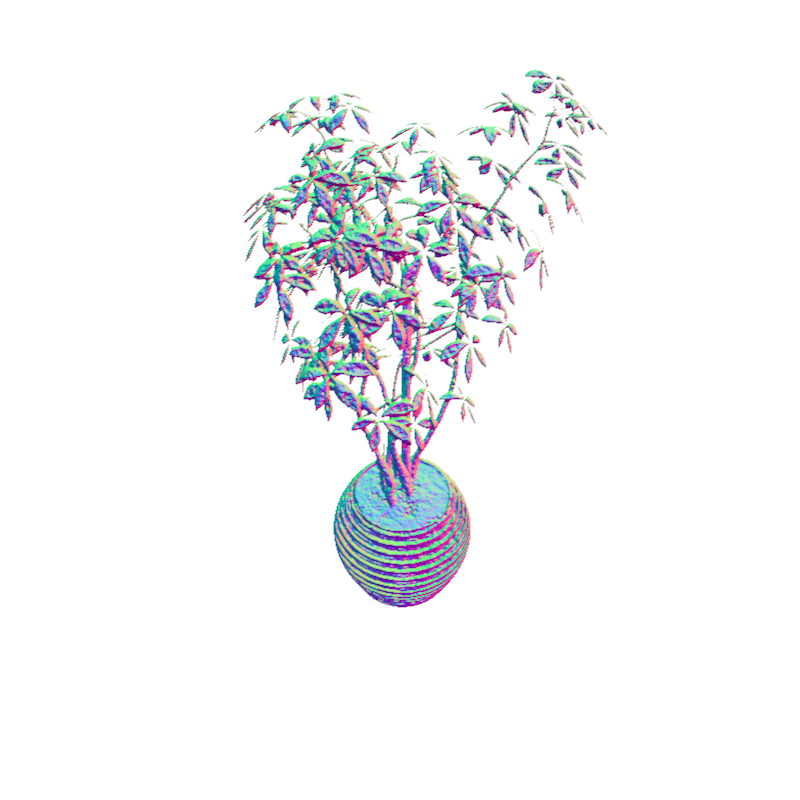} \hspace{-0.6cm} &
    \includegraphics[width=0.165\linewidth]{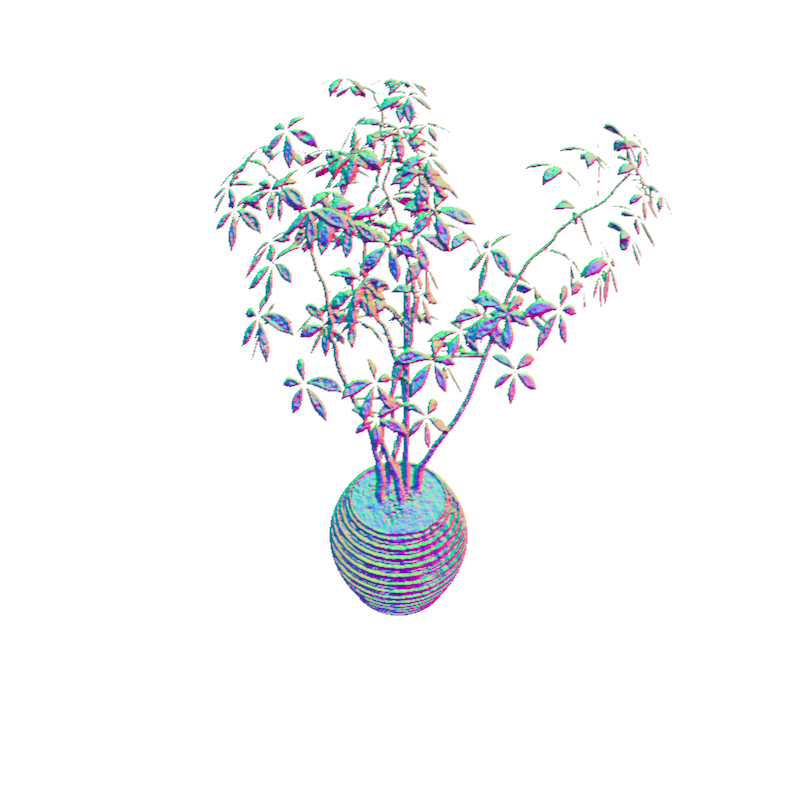} \hspace{-0.6cm} &
    \includegraphics[width=0.165\linewidth]{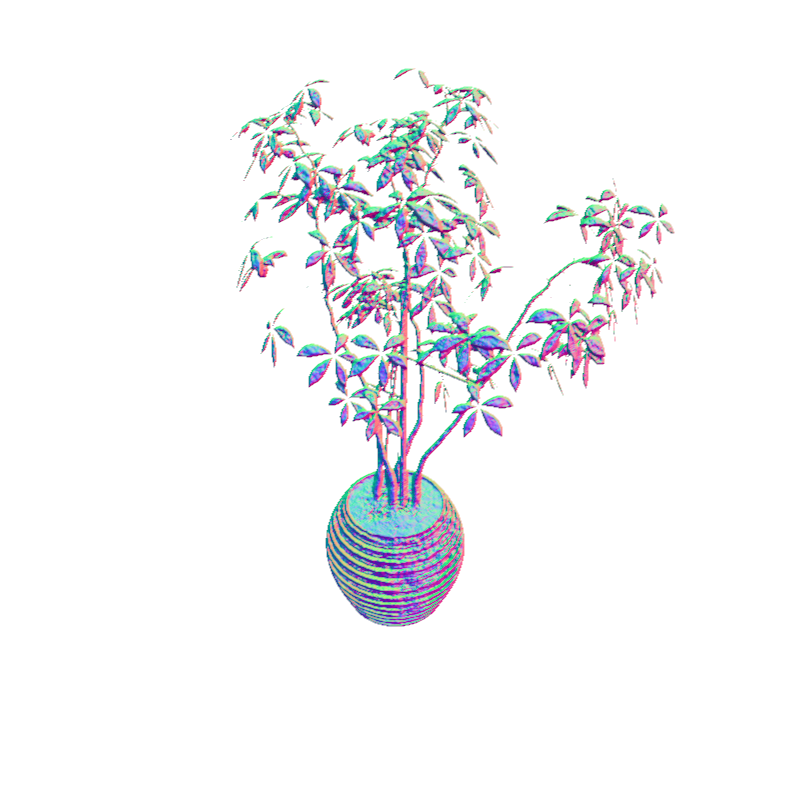} \hspace{-0.6cm} &
    \includegraphics[width=0.165\linewidth]{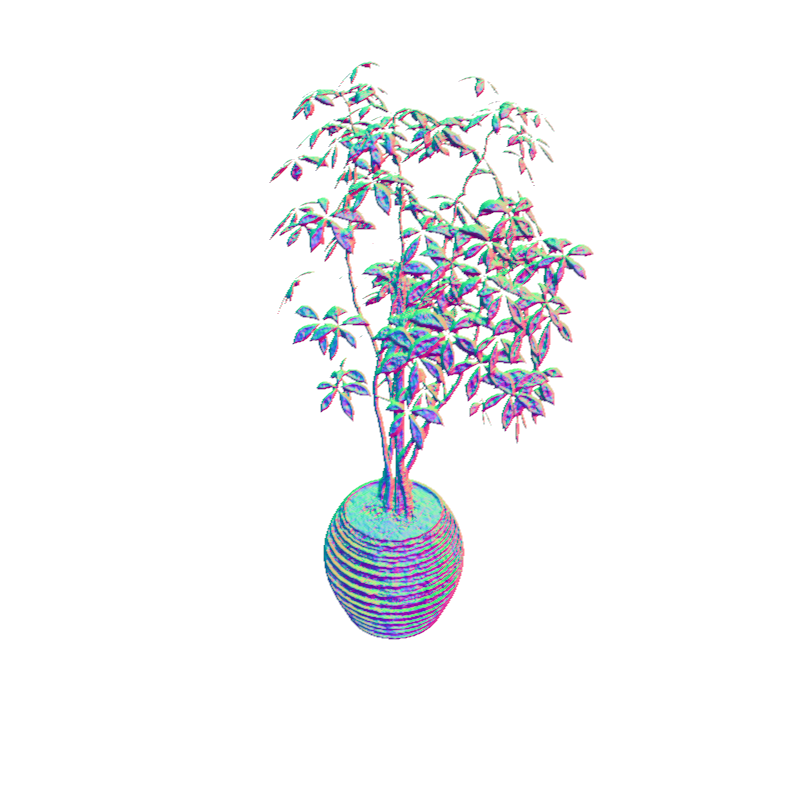} \hspace{-0.6cm} &
    \includegraphics[width=0.165\linewidth]{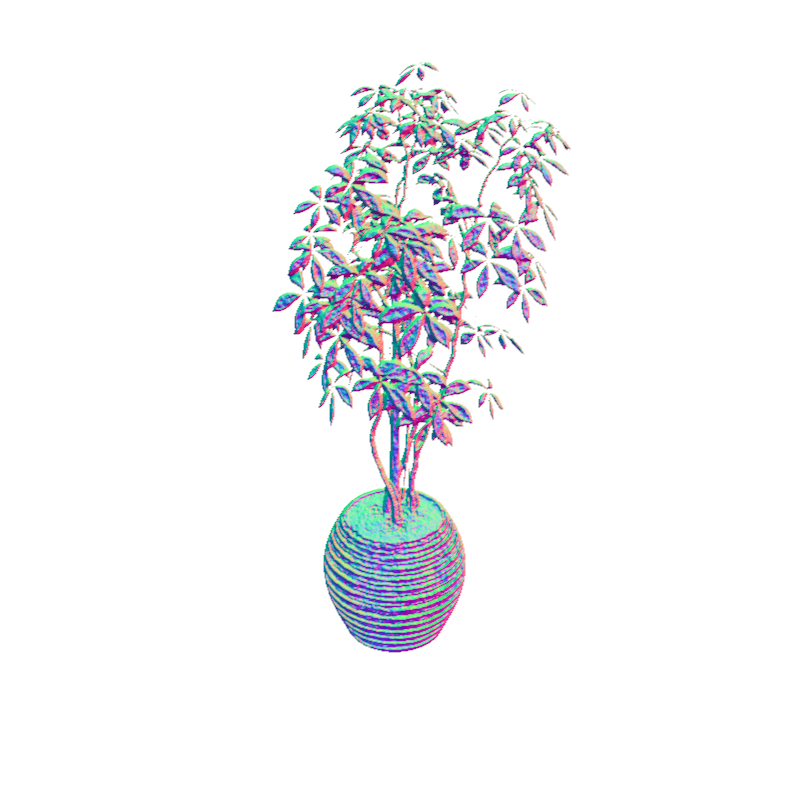} \hspace{-0.6cm} 
    \\ 
    \hspace{-0.8cm}
    \includegraphics[width=0.165\linewidth]{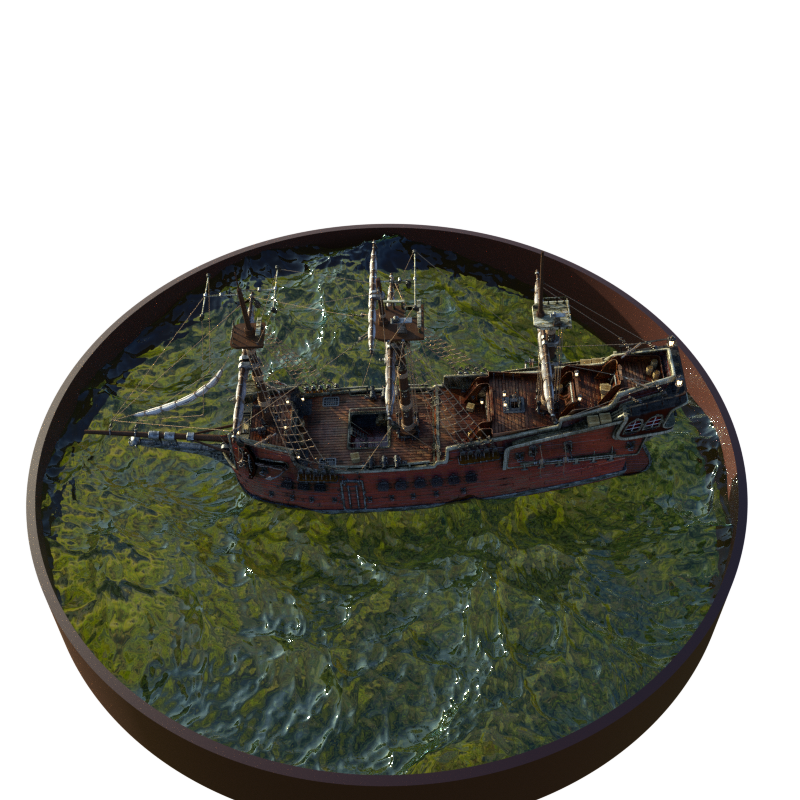} \hspace{-0.6cm} &
    \includegraphics[width=0.165\linewidth]{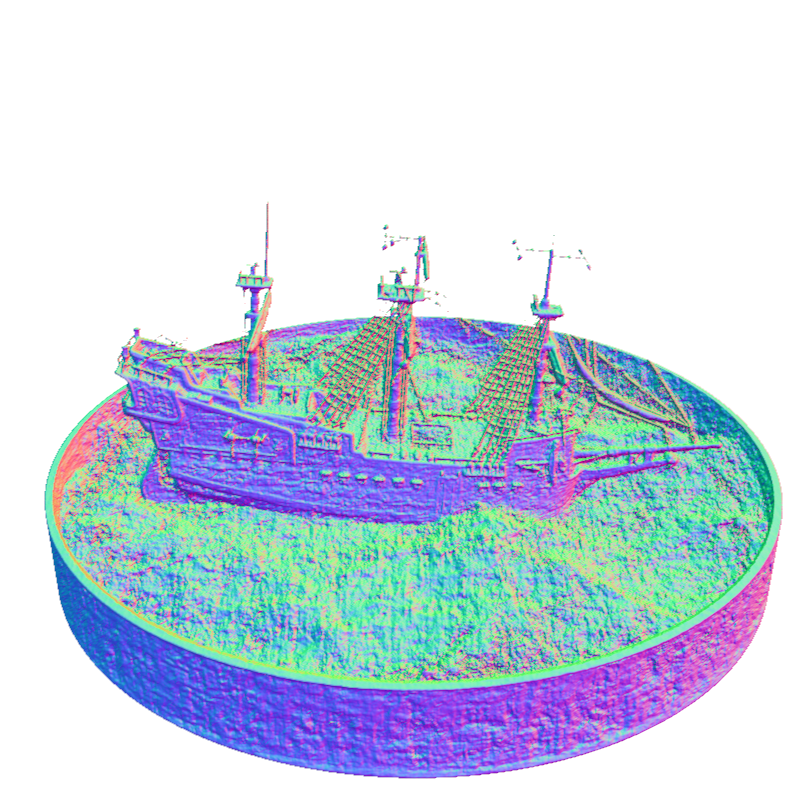} \hspace{-0.6cm} &
    \includegraphics[width=0.165\linewidth]{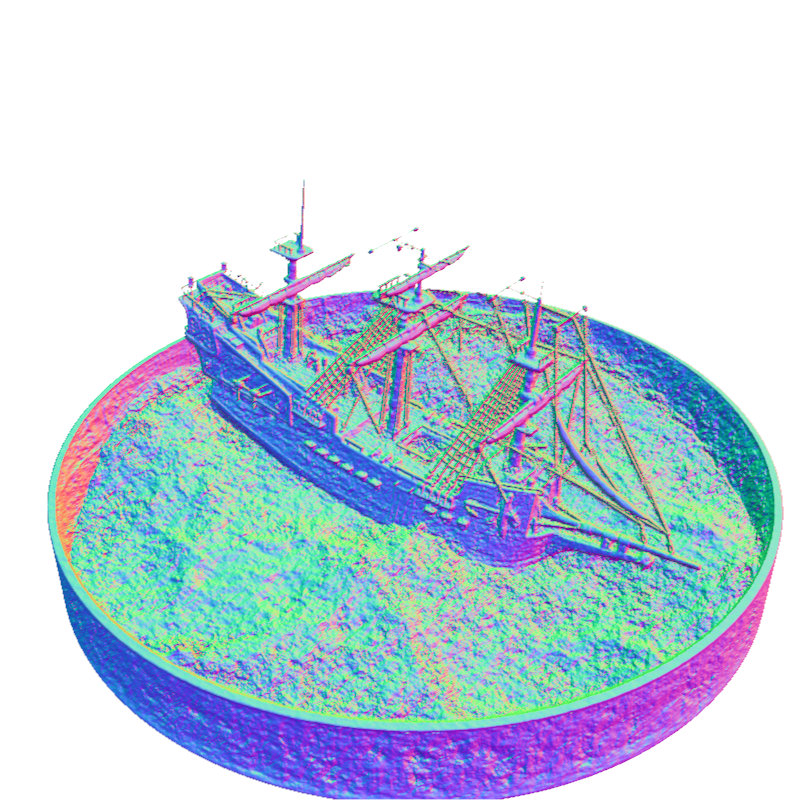} \hspace{-0.6cm} &
    \includegraphics[width=0.165\linewidth]{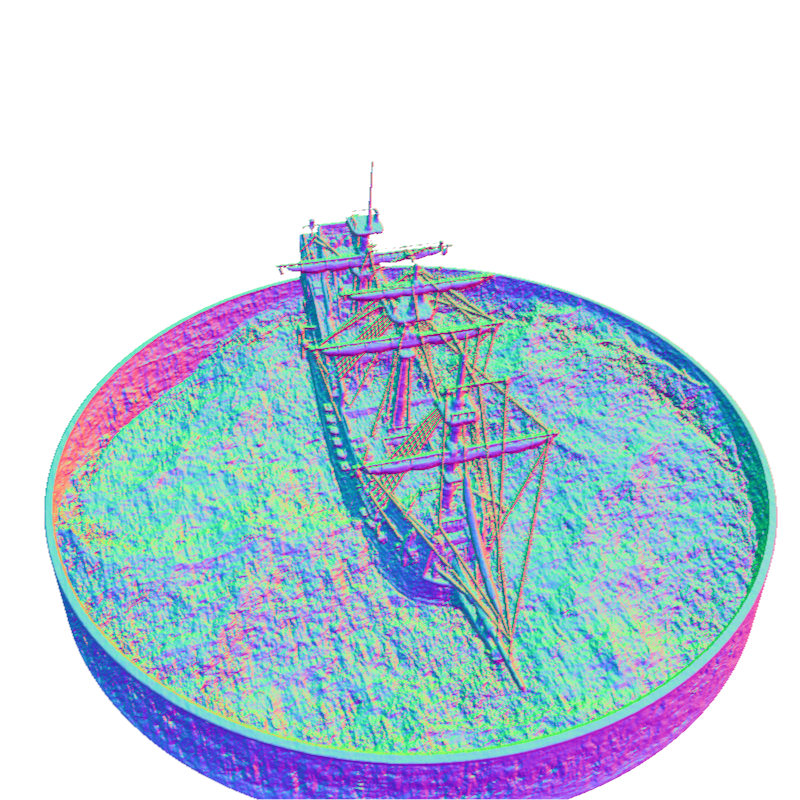} \hspace{-0.6cm} &
    \includegraphics[width=0.165\linewidth]{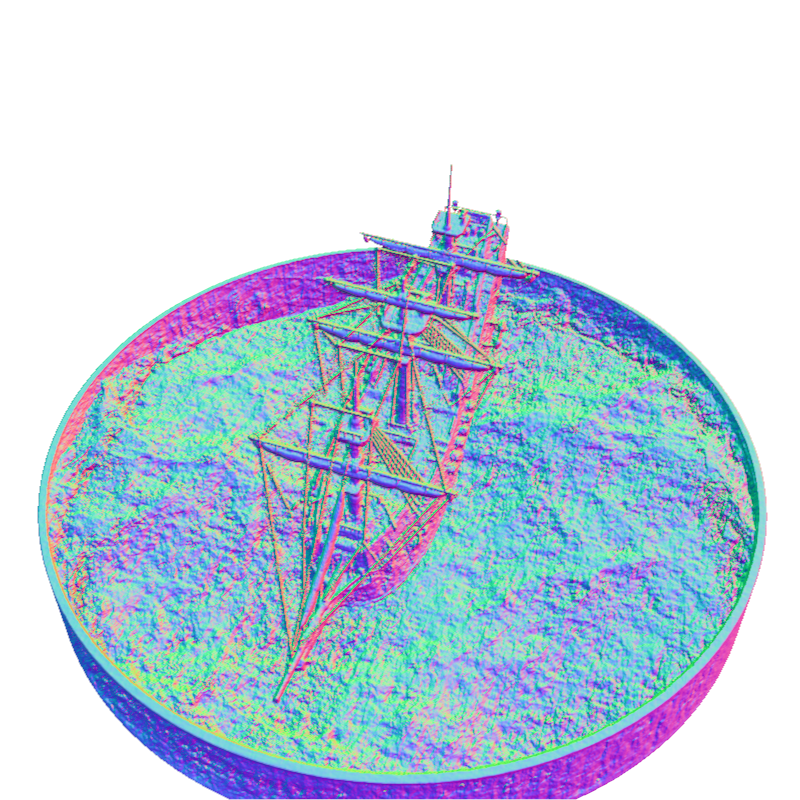} \hspace{-0.6cm} &
    \includegraphics[width=0.165\linewidth]{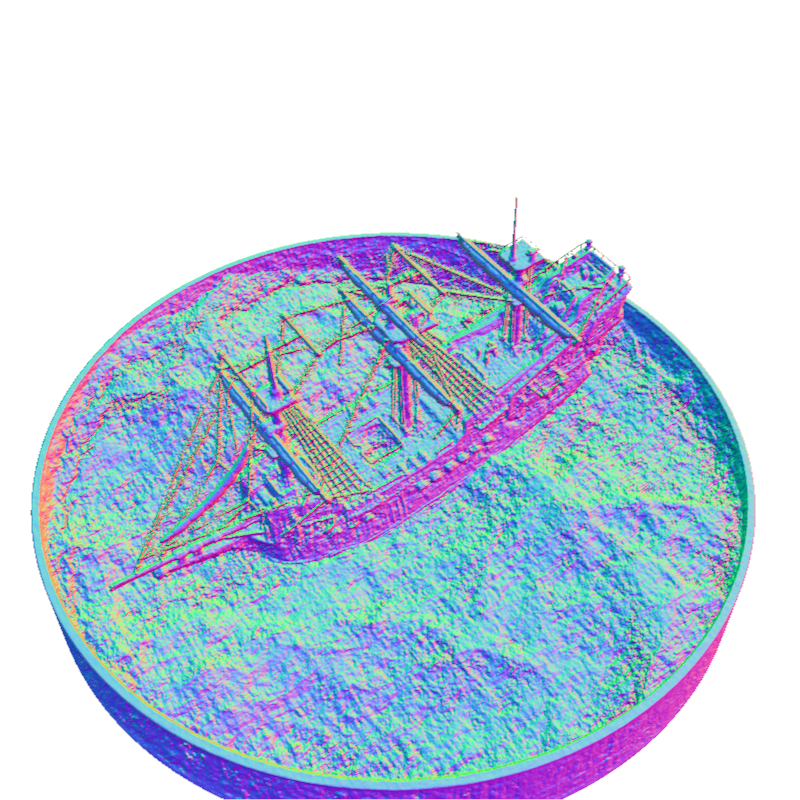} \hspace{-0.6cm} 
    \\ 
    \hspace{-0.8cm}
    \includegraphics[width=0.165\linewidth]{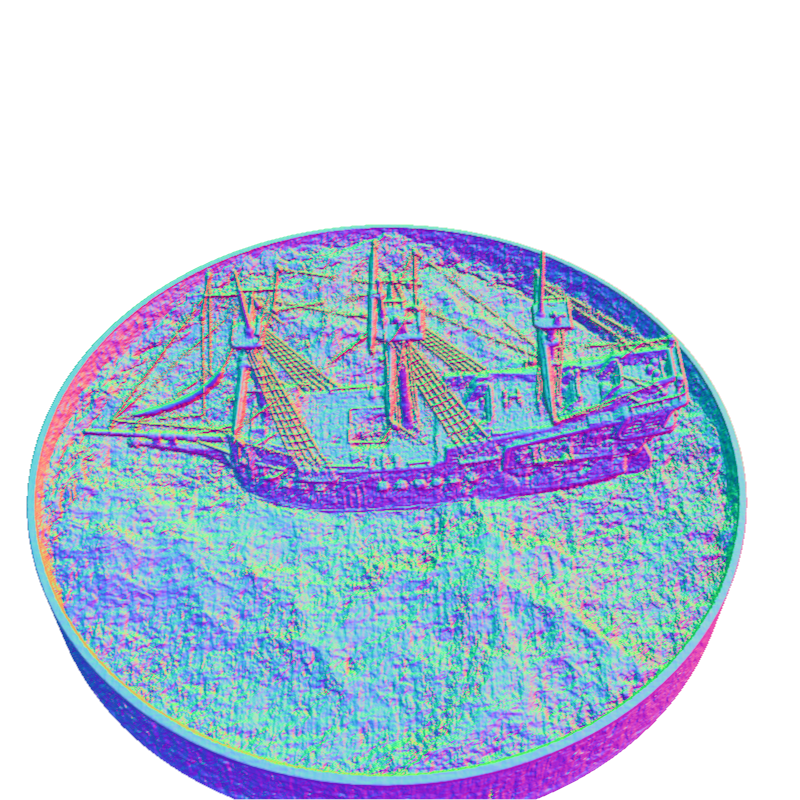} \hspace{-0.6cm} &
    \includegraphics[width=0.165\linewidth]{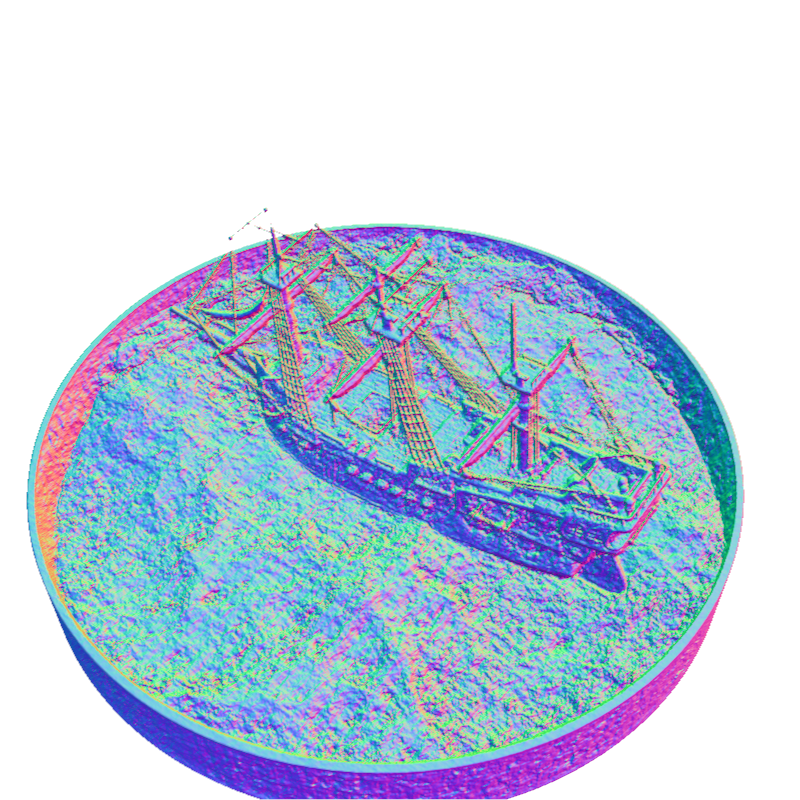} \hspace{-0.6cm} &
    \includegraphics[width=0.165\linewidth]{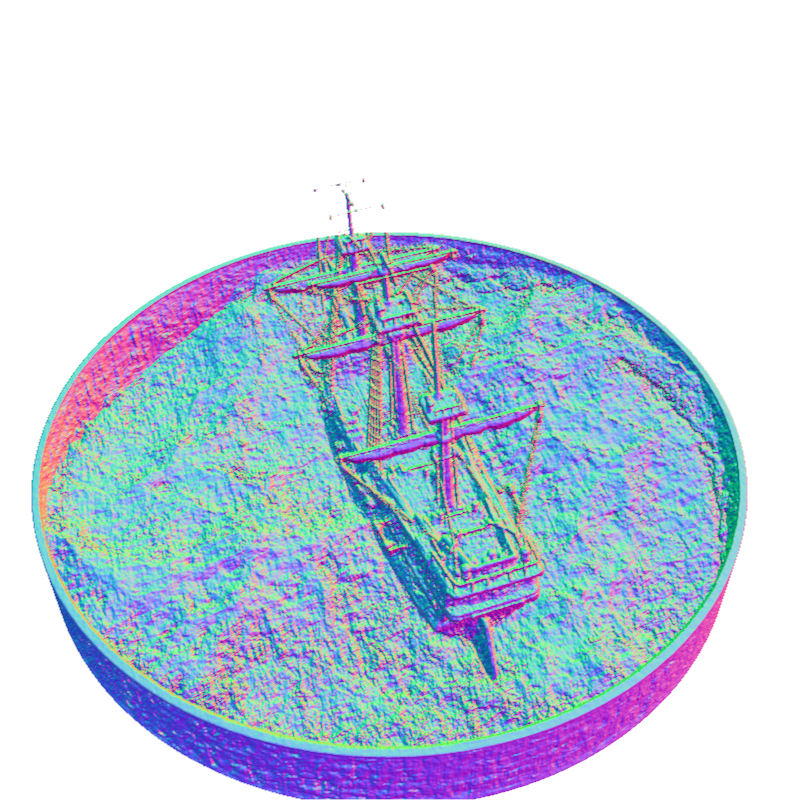} \hspace{-0.6cm} &
    \includegraphics[width=0.165\linewidth]{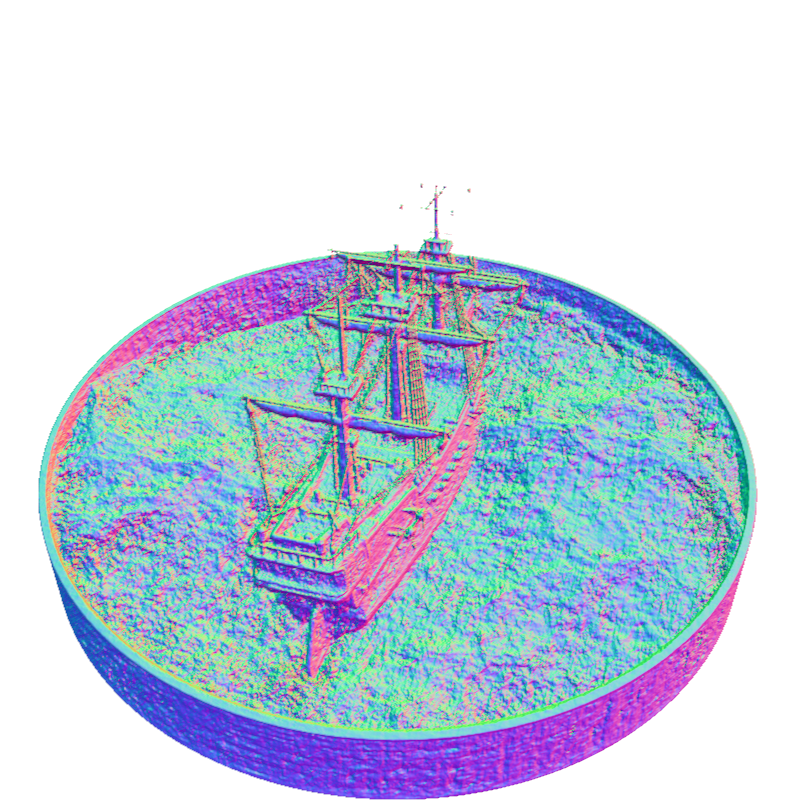} \hspace{-0.6cm} &
    \includegraphics[width=0.165\linewidth]{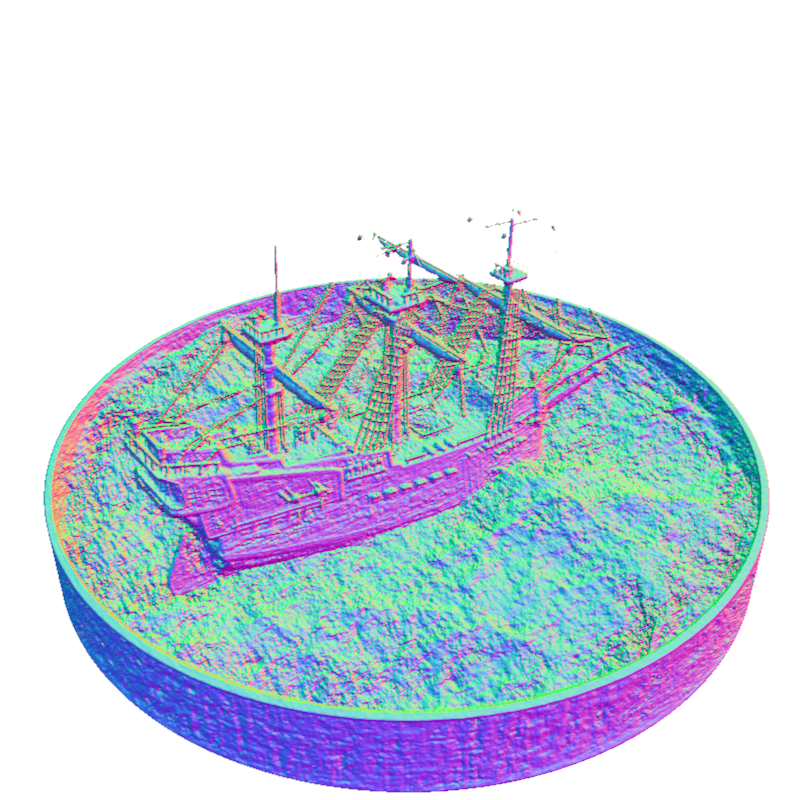} \hspace{-0.6cm} &
    \includegraphics[width=0.165\linewidth]{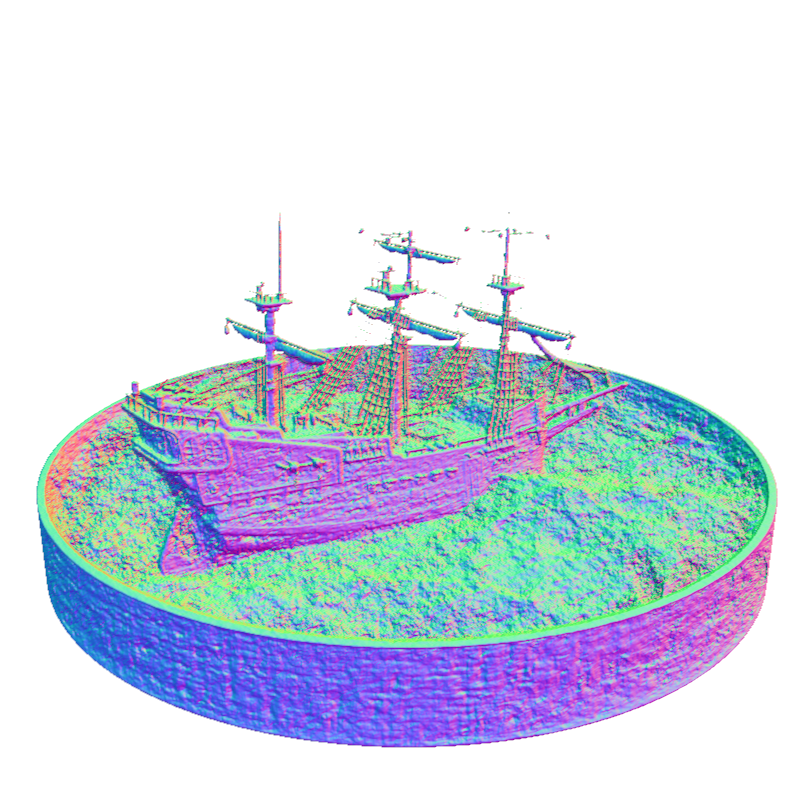} \hspace{-0.6cm} 
    \\ 
    \end{tabular}
    \caption{Visualization of extracted mesh results. The surface reconstructed by our method is colored with normal vectors. The color image on the left is the ground truth.}
    \label{fig:mesh_normal}
\end{figure*}

Our experiments demonstrate that CeRF is able to learn the geometric information of the objects in the scene, despite the fact that the features extracted in 3D space are incorporated along the ray. We obtain the distance map by normalizing the radiance coefficients and then multiplying them with the distances of the sampled points. As shown in Figure~\ref{fig:mesh_vis} and Figure~\ref{fig:mesh_normal}, the mesh is extracted from the model trained from CeRF, and it can be seen that the rich details on the objects are well reconstructed on the mesh, such as the protrusions on the thin horizontal bar in the Lego scene, and hollow parts of the track. CeRF can learn the geometric information well and extract the geometric information easily, which retains the advantage of radiance field.

\section{Additional results}

We additionally provide a per-scene comparison between CeRF and other methods, where Table \ref{tb:blender_per_scene_psnr}, \ref{tb:blender_per_scene_ssim} and \ref{tb:blender_per_scene_lpips} show the metrics on the Blender dataset, and Table \ref{tb:shiny_per_scene_psnr}, \ref{tb:shiny_per_scene_ssim} and \ref{tb:shiny_per_scene_lpips} reflect the quantitative results on the Shiny Blender Dataset. We also included comparison videos with additional and other methods in our supplementary material, as well as ablative videos of our module.

\newpage

\begin{table}[!t]
    \caption{Per-scene test set PSNRs on the Blender dataset}
    \label{tb:blender_per_scene_psnr}
    \centering
    \begin{tabular}{c|cccccccc}
    \toprule
    & chair & drums & ficus & hotdog & lego & materials & mic & ship \\ 
    \midrule  \midrule
    Plenoxel & 33.98 & 25.35 & \cellcolor[RGB]{255,255,187} 31.83 & 36.43 & 34.10 & 29.14 & 33.26 & 29.62  \\
    R2L & \cellcolor[RGB]{244,181,180} 36.71 & \cellcolor[RGB]{249,218,183} 26.03 & 28.63 & \cellcolor[RGB]{249,218,183} 38.07 & 32.53 & 30.20 & 32.80 & 29.98  \\
    Mip-NeRF & 35.12 & 25.36 & 33.19 & 37.34 & \cellcolor[RGB]{255,255,187} 35.92 & \cellcolor[RGB]{255,255,187} 30.64 & \cellcolor[RGB]{244,181,180} 36.76 & \cellcolor[RGB]{249,218,183} 30.52 \\
    Ref-NeRF & \cellcolor[RGB]{255,255,187} 35.83 & \cellcolor[RGB]{255,255,187} 25.79 & \cellcolor[RGB]{249,218,183} 33.91 & \cellcolor[RGB]{255,255,187} 37.72 & \cellcolor[RGB]{249,218,183} 36.25 & \cellcolor[RGB]{244,181,180} 35.41 & \cellcolor[RGB]{244,181,180} 36.76 & \cellcolor[RGB]{255,255,187} 30.28 \\ 
    CeRF &\cellcolor[RGB]{249,218,183} 35.99 & \cellcolor[RGB]{244,181,180} 26.41 & \cellcolor[RGB]{244,181,180} 37.32 &  \cellcolor[RGB]{244,181,180}38.24 & \cellcolor[RGB]{244,181,180} 37.79 & \cellcolor[RGB]{249,218,183} 33.22 & \cellcolor[RGB]{255,255,187} 36.35 & \cellcolor[RGB]{244,181,180} 31.23 \\ 
    \bottomrule
    \end{tabular}
\end{table}

\begin{table}[!t]
    \caption{Per-scene test set SSIMs on the Blender dataset}
    \label{tb:blender_per_scene_ssim}
    \centering
    \begin{tabular}{c|cccccccc}
    \toprule
    & chair & drums & ficus & hotdog & lego & materials & mic & ship \\ 
    \midrule  \midrule
    Plenoxel & 0.977 & 0.933 & 0.976 & 0.980 & 0.975 & 0.949 & 0.985 & 0.890  \\
    R2L & \cellcolor[RGB]{244,181,180} 0.999 & \cellcolor[RGB]{244,181,180} 0.988 & \cellcolor[RGB]{244,181,180} 0.996 & \cellcolor[RGB]{244,181,180} 0.999 & \cellcolor[RGB]{244,181,180} 0.994 & \cellcolor[RGB]{244,181,180} 0.992 & \cellcolor[RGB]{244,181,180} 0.997 & \cellcolor[RGB]{244,181,180} 0.986  \\
    Mip-NeRF & 0.981 & 0.933 & 0.980 & 0.982 & 0.980 & 0.959 & \cellcolor[RGB]{249,218,183} 0.992 & 0.885 \\
    Ref-NeRF & \cellcolor[RGB]{249,218,183} 0.984 & \cellcolor[RGB]{255,255,187} 0.937 & \cellcolor[RGB]{255,255,187} 0.983 & \cellcolor[RGB]{255,255,187} 0.984 & \cellcolor[RGB]{255,255,187} 0.981 & \cellcolor[RGB]{249,218,183} 0.983 & \cellcolor[RGB]{249,218,183} 0.992 & \cellcolor[RGB]{255,255,187} 0.880 \\ 
    CeRF & \cellcolor[RGB]{249,218,183} 0.984 & \cellcolor[RGB]{249,218,183} 0.945 & \cellcolor[RGB]{249,218,183} 0.990 & \cellcolor[RGB]{249,218,183} 0.985 & \cellcolor[RGB]{249,218,183} 0.985 & \cellcolor[RGB]{255,255,187} 0.975 & \cellcolor[RGB]{255,255,187} 0.991 & \cellcolor[RGB]{249,218,183} 0.897 \\ 
    \bottomrule
    \end{tabular}
\end{table}

\begin{table}[!t]
    \caption{Per-scene test set LPIPS on the Blender dataset}
    \label{tb:blender_per_scene_lpips}
    \centering
    \begin{tabular}{c|cccccccc}
    \toprule
    & chair & drums & ficus & hotdog & lego & materials & mic & ship \\ 
    \midrule  \midrule
    Plenoxel & 0.031 & 0.067 & 0.026 & 0.037 & 0.028 & 0.057 & 0.015 & \cellcolor[RGB]{249,218,183} 0.134  \\
    Mip-NeRF & \cellcolor[RGB]{255,255,187} 0.020 & \cellcolor[RGB]{255,255,187} 0.064 & \cellcolor[RGB]{255,255,187} 0.021 & \cellcolor[RGB]{255,255,187} 0.026 & \cellcolor[RGB]{249,218,183} 0.018 & \cellcolor[RGB]{255,255,187} 0.040 & \cellcolor[RGB]{249,218,183} 0.008 & \cellcolor[RGB]{255,255,187} 0.135 \\
    Ref-NeRF & \cellcolor[RGB]{244,181,180} 0.017 & \cellcolor[RGB]{249,218,183} 0.059 & \cellcolor[RGB]{249,218,183} 0.019 & \cellcolor[RGB]{249,218,183} 0.022 & \cellcolor[RGB]{249,218,183} 0.018 & \cellcolor[RGB]{244,181,180} 0.022 & \cellcolor[RGB]{244,181,180} 0.007 & 0.139 \\ 
    CeRF & \cellcolor[RGB]{244,181,180} 0.017 & \cellcolor[RGB]{244,181,180} 0.054 & \cellcolor[RGB]{244,181,180} 0.012 & \cellcolor[RGB]{244,181,180} 0.021 & \cellcolor[RGB]{244,181,180} 0.014 & \cellcolor[RGB]{249,218,183} 0.028 & \cellcolor[RGB]{249,218,183} 0.008 & \cellcolor[RGB]{244,181,180} 0.122 \\ 
    \bottomrule
    \end{tabular}
\end{table}

\begin{table}[!t]
    \caption{Per-scene test set PSNRs on the Shiny Blender dataset}
    \label{tb:shiny_per_scene_psnr}
    \centering
    \begin{tabular}{c|cccccc}
    \toprule
    & teapot & toaster & coffee & helmet & car & ball \\ 
    \midrule  \midrule
    Plenoxel & 44.25& 19.51 & \cellcolor[RGB]{255,255,187} 31.55 & 26.94 & 26.11 & 24.52  \\
    DVGO & 44.79 & 22.18 & 31.48 & \cellcolor[RGB]{255,255,187} 27.75 & \cellcolor[RGB]{255,255,187} 26.90 & 26.13  \\
    PhySG & 35.83 & 18.59 & 23.71 & 27.51 & 24.40 & \cellcolor[RGB]{255,255,187} 27.24 \\
    Mip-NeRF & \cellcolor[RGB]{255,255,187} 46.00 & \cellcolor[RGB]{255,255,187} 22.37 & 30.36 & 27.39 & 26.50 & 25.94 \\
    Ref-NeRF & \cellcolor[RGB]{244,181,180} 47.90 & \cellcolor[RGB]{249,218,183} 25.70 & \cellcolor[RGB]{244,181,180} 34.21 & \cellcolor[RGB]{244,181,180} 29.68 & \cellcolor[RGB]{244,181,180} 30.82 & \cellcolor[RGB]{244,181,180} 47.46 \\ 
    CeRF & \cellcolor[RGB]{249,218,183} 47.17 & \cellcolor[RGB]{244,181,180} 26.82 & \cellcolor[RGB]{249,218,183} 32.44 & \cellcolor[RGB]{249,218,183} 29.53 & \cellcolor[RGB]{249,218,183} 27.99 & \cellcolor[RGB]{249,218,183} 37.66 \\ 
    \bottomrule
    \end{tabular}
\end{table}

\begin{table}[!t]
    \caption{Per-scene test set SSIMs on the Shiny Blender dataset}
    \label{tb:shiny_per_scene_ssim}
    \centering
    \begin{tabular}{c|cccccc}
    \toprule
    & teapot & toaster & coffee & helmet & car & ball \\ 
    \midrule  \midrule
    Plenoxel & 0.996 & 0.772 & 0.963 & 0.913 & 0.905 & 0.832 \\
    DVGO & 0.996 & 0.848 & 0.962 & 0.932 & 0.916 & 0.901 \\
    PhySG & 0.990 & 0.805 & 0.922 & \cellcolor[RGB]{255,255,187} 0.953 & 0.910 & \cellcolor[RGB]{255,255,187} 0.947 \\
    Mip-NeRF & \cellcolor[RGB]{249,218,183} 0.997 & \cellcolor[RGB]{255,255,187} 0.891 & \cellcolor[RGB]{255,255,187} 0.966 & 0.939 & \cellcolor[RGB]{255,255,187} 0.922 & 0.935 \\
    Ref-NeRF & \cellcolor[RGB]{244,181,180} 0.998 & \cellcolor[RGB]{249,218,183} 0.922 & \cellcolor[RGB]{244,181,180} 0.974 & \cellcolor[RGB]{244,181,180} 0.958 & \cellcolor[RGB]{244,181,180} 0.955 & \cellcolor[RGB]{244,181,180} 0.995 \\ 
    CeRF & \cellcolor[RGB]{249,218,183} 0.997 & \cellcolor[RGB]{244,181,180} 0.937 & \cellcolor[RGB]{249,218,183} 0.972 & \cellcolor[RGB]{244,181,180} 0.958 & \cellcolor[RGB]{249,218,183} 0.936 & \cellcolor[RGB]{249,218,183} 0.987 \\ 
    \bottomrule
    \end{tabular}
\end{table}

\begin{table}[!t]
    \caption{Per-scene test set LPIPS on the Shiny Blender dataset}
    \label{tb:shiny_per_scene_lpips}
    \centering
    \begin{tabular}{c|cccccc}
    \toprule
    & teapot & toaster & coffee & helmet & car & ball \\ 
    \midrule  \midrule
    Plenoxel & 0.016 & 0.244 & 0.144 & 0.170 & 0.086 & 0.274\\
    DVGO & 0.019 & 0.220 & 0.141 & 0.154 & 0.081 & 0.253  \\
    PhySG & 0.022 & 0.194 & 0.15 & \cellcolor[RGB]{255,255,187} 0.089 & 0.091 & 0.700 \\
    Mip-NeRF & \cellcolor[RGB]{255,255,187} 0.008 & \cellcolor[RGB]{255,255,187} 0.123 & \cellcolor[RGB]{255,255,187} 0.086 & 0.108 & \cellcolor[RGB]{255,255,187} 0.059 & \cellcolor[RGB]{255,255,187} 0.168 \\
    Ref-NeRF & \cellcolor[RGB]{244,181,180} 0.004 & \cellcolor[RGB]{249,218,183} 0.095 & \cellcolor[RGB]{244,181,180} 0.078 & \cellcolor[RGB]{249,218,183} 0.075 & \cellcolor[RGB]{244,181,180} 0.041 & \cellcolor[RGB]{244,181,180} 0.059 \\ 
    CeRF & \cellcolor[RGB]{249,218,183} 0.005 & \cellcolor[RGB]{244,181,180} 0.077 & \cellcolor[RGB]{249,218,183} 0.081 & \cellcolor[RGB]{244,181,180} 0.069 & \cellcolor[RGB]{249,218,183} 0.052 & \cellcolor[RGB]{249,218,183} 0.073 \\ 
    \bottomrule
    \end{tabular}
\end{table}

\end{document}